%% file: ms.tex
\RequirePackage{snapshot}
\documentclass[final]{cvpr}

\usepackage{times}
\usepackage{epsfig}
\usepackage{graphicx}
\usepackage{amsmath}
\usepackage{amssymb}
\usepackage{booktabs}
\usepackage{graphbox} %
\usepackage{capt-of} %
\usepackage{mathtools} %
\usepackage{multirow}
\usepackage{makecell}
\usepackage{placeins} %
\usepackage{adjustbox}
\usepackage{nicefrac}
\usepackage[pagebackref=true,breaklinks=true,colorlinks,bookmarks=false]{hyperref}
\usepackage{breqn}

\def\cvprPaperID{4270} %
\def\confYear{CVPR 2021}

\newcommand{\RR}{\mathbb{R}}
\newcommand{\codebook}{\mathcal{Z}}
\newcommand{\discriminator}{D}
\newcommand{\decoder}{G}
\newcommand{\encoder}{E}
\newcommand{\quantize}{\mathbf{q}}
\newcommand{\quantizedcode}{z_{\mathbf{q}}}
\DeclareMathOperator*{\argmin}{arg\,min}

\newcommand{\blfootnote}[1]{%
  \begingroup
  \renewcommand\thefootnote{}\footnote{#1}%
  \addtocounter{footnote}{-1}%
  \endgroup
  }

\input{cvpr_figures}

\input{cvpr_tables}

\begin{document}
\title{Taming Transformers for High-Resolution Image Synthesis}

\author{Patrick Esser\thanks{Both authors contributed equally to this
work.} \qquad Robin Rombach\footnotemark[1] \qquad Bj\"orn Ommer\\
Heidelberg Collaboratory for Image Processing, IWR, Heidelberg University, Germany\\
\small{*Both authors contributed equally to this work}
}

\twocolumn[{%
\vspace{-2em}
\maketitle%
\alaskatop%
}]

\begin{abstract}
\noindent Designed to learn long-range interactions on sequential data, transformers
  continue to show state-of-the-art results on a wide variety of tasks.  In
  contrast to CNNs, they contain no inductive bias that prioritizes local
  interactions. This makes them expressive, but also computationally infeasible
  for long sequences, such as high-resolution images. We demonstrate how
  combining the effectiveness of the inductive bias of CNNs with the
  expressivity of transformers enables
  them to model and thereby synthesize high-resolution images.
  We show how to (i) use CNNs to learn a context-rich vocabulary of
  image constituents, and in turn (ii) utilize transformers to efficiently
  model their composition within high-resolution images.
  Our approach is readily applied to conditional synthesis tasks, where both
  non-spatial information, such as object classes, and spatial information,
  such as segmentations, can control the generated image.
  In particular, we present the first results on semantically-guided synthesis
  of megapixel images with transformers and obtain the state of the art among 
  autoregressive models on class-conditional ImageNet. 
  Code and pretrained models can be found at \url{https://git.io/JnyvK}.
\end{abstract}
\vspace{-1em} %
\enlargethispage{1\baselineskip}
\section{Introduction}
\noindent Transformers are on the rise---they are now the de-facto standard architecture
for language tasks
\cite{vaswani2017attention,Radford2018ImprovingLU,Radford2019LanguageMA,brown2020language}
and are increasingly adapted in other areas such as audio
\cite{child2019generating} and vision \cite{chen2020generative,
anonymous2021an}. In contrast to the predominant vision architecture,
convolutional neural networks (CNNs), the transformer architecture contains no
built-in inductive prior on the locality of interactions and is therefore free
to learn complex relationships among its inputs.
\enlargethispage{1\baselineskip}
However, this generality also implies that it \emph{has to} learn all
relationships, whereas CNNs have been designed to exploit
prior knowledge about strong local correlations within images.
Thus, the increased expressivity of transformers comes with quadratically
increasing computational costs, because all pairwise interactions are taken
into account. The resulting energy and time requirements of state-of-the-art
transformer models thus pose fundamental problems for scaling them to
high-resolution images with millions of pixels.
Observations that transformers tend to learn convolutional structures
\cite{anonymous2021an} thus beg the question: Do we have to re-learn everything
we know about the local structure and regularity of images from scratch each time we train a vision model, or can we efficiently encode inductive image biases while still retaining the flexibility of
transformers? We hypothesize that low-level image structure is well
described by a local connectivity, i.e.\ a convolutional architecture,
whereas this structural assumption ceases to be effective on higher semantic levels.
Moreover, CNNs not only exhibit a strong locality bias, but also a bias towards
spatial invariance through the use of shared weights across all positions. This
makes them ineffective if a more holistic understanding of the input is
required.

Our key insight to obtain an effective and expressive model is that, \emph{taken
together, convolutional and transformer architectures can model
the compositional nature of our visual world \cite{ommer2007learning}}: We use a
convolutional approach to efficiently learn a codebook of context-rich visual parts
and, subsequently, learn a model of their global compositions.
The long-range interactions within these compositions require
an expressive transformer architecture to model distributions over their consituent
visual parts. Furthermore, we utilize an adversarial approach to ensure that the dictionary of local parts captures perceptually important local structure to alleviate the need for modeling low-level
statistics with the transformer architecture. 
Allowing transformers to concentrate on their unique strength---modeling
long-range relations---enables them
to generate high-resolution images as in
Fig.~\ref{fig:frontpage}, a
feat which previously has been out of reach. Our formulation% directly
gives control over the generated images by means of conditioning information
regarding desired object classes or spatial layouts. Finally, experiments
demonstrate that our approach retains the advantages of transformers by
outperforming previous codebook-based state-of-the-art approaches based on convolutional
architectures.

\section{Related Work}
\newcommand{\seqlength}{n}
\paragraph{The Transformer Family}
The defining characteristic of the transformer architecture
\cite{vaswani2017attention} is that it models interactions between its inputs
solely through attention
\cite{bahdanau2016neural,kim2017structured,parikh2016decomposable} which
enables them to faithfully handle interactions between inputs regardless of
their relative position to one another.
Originally applied to language tasks, inputs to the transformer were given by tokens,
but other signals, such as those obtained from audio \cite{li2019neural} or images
\cite{chen2020generative},
can be used. Each layer of the transformer then consists of an attention
mechanism, which allows for interaction between inputs at different positions,
followed by a position-wise fully connected network, which is applied to all positions
independently. More specifically, the (self-)attention mechanism can be described by mapping an intermediate representation with three position-wise linear layers into three representations, query $Q \in \RR^{N\times d_k}$, key $K \in \RR^{N\times d_k}$
and value $V \in \RR^{N\times d_v}$, to compute the output as
\newcommand{\Attn}{\operatorname{Attn}}
\newcommand{\softmax}{\operatorname{softmax}}
\begin{equation}
  \Attn(Q,K,V) = \softmax\Big(\frac{QK^t}{\sqrt{d_k}}\Big)V \in
  \RR^{N\times d_v}.
\end{equation}

When performing autoregressive maximum-likelihood learning, non-causal entries
of $QK^t$, i.e. all entries below its diagonal, are set to $-\infty$ and the
final output of the transformer is given after a linear, point-wise
transformation to predict logits of the next sequence element.
Since the attention mechanism relies on the computation of inner products
between all pairs of elements in the sequence, its computational complexity
increases quadratically with the sequence length.  While the ability to
consider interactions between \emph{all} elements is the reason transformers
efficiently learn long-range interactions, it is also the reason transformers
quickly become infeasible, especially on images, where the sequence length
itself scales quadratically with the resolution.
Different approaches have been proposed to reduce the computational
requirements to make transformers feasible for longer sequences.
\cite{parmar2018image} and \cite{weissenborn2020scaling} restrict the receptive
fields of the attention modules, which reduces the expressivity and, especially
for high-resolution images, introduces assumptions on the
independence of pixels.
\cite{child2019generating} and \cite{ho2019axial} retain the full receptive
field but can reduce costs for a sequence of length $\seqlength$ only from $n^2$ to $n \sqrt{n}$, which makes
resolutions beyond $64$ pixels still prohibitively expensive.
\modelfigure
\paragraph{Convolutional Approaches}
The two-dimensional structure of images suggests that local
interactions are particularly important. CNNs exploit this structure by
restricting interactions between input variables to a local neighborhood
defined by the kernel size of the convolutional kernel. Applying a kernel thus
results in costs that scale linearly with the overall sequence length (the
number of pixels in the case of images) and quadratically in the kernel size,
which, in modern CNN architectures, is often fixed to a small constant such as
$3\times3$. This inductive bias towards local interactions thus leads to
efficient computations, but the wide range of specialized layers which are
introduced into CNNs to handle different synthesis tasks
\cite{park2019semantic,zhang2020cross,Siarohin_2019_NeurIPS,zhu2019sean,zhou2017view}
suggest that this bias is often too restrictive.

Convolutional architectures have been used for 
autoregressive modeling of images \cite{oord2016pixel,oord2016conditional,chen2018pixelsnail} but, for
low-resolution images,
previous works \cite{parmar2018image,child2019generating,ho2019axial}
demonstrated that transformers consistently outperform their convolutional
counterparts.
Our approach allows us to efficiently model high-resolution images with
transformers while retaining their advantages over state-of-the-art
convolutional approaches.

\paragraph{Two-Stage Approaches}
Closest to ours are two-stage approaches which first learn an encoding of data
and afterwards learn, in a second stage, a probabilistic model of this
encoding. \cite{dai2019diagnosing} demonstrated both theoretical and empirical
evidence on the advantages of first learning a data representation with a
Variational Autoencoder (VAE) \cite{vae,vae2}, and then again learning its
distribution with a VAE. \cite{esser2020invertible,xiao2019generative} demonstrate similar gains when using an
unconditional normalizing flow for the second stage, and \cite{rombach2020making,rombach2020network}
when using a conditional normalizing flow. To improve training efficiency of
Generative Adversarial Networks (GANs), \cite{liu2019acceleration} learns a
GAN \cite{gan} on representations of an
autoencoder and \cite{han2020notsobiggan} on low-resolution wavelet coefficients
which are then decoded to images with a learned generator.

\cite{oord2018neural} presents the Vector Quantised Variational Autoencoder (VQVAE), an
approach to learn discrete representations of images,
and models their distribution autoregressively with a convolutional
architecture. \cite{razavi2019generating} extends this approach to use a hierarchy of
learned representations. However, these methods still rely on
convolutional density estimation, which makes it difficult to capture
long-range interactions in high-resolution images.
\cite{chen2020generative} models images autoregressively with transformers in order to
evaluate the suitability of generative pretraining to learn image representations for downstream tasks. Since input resolutions of $32\times 32$ pixels
are still quite computationally expensive \cite{chen2020generative}, a VQVAE is used to
encode images up to a resolution of $192\times 192$. In an effort to keep the
learned discrete representation as spatially invariant as possible with respect
to the pixels, a shallow VQVAE with small receptive field is employed.
In contrast, we demonstrate that a powerful first stage, which captures as much
context as possible in the learned representation, is critical to enable
efficient high-resolution image synthesis with transformers.

\section{Approach}
\label{sec:approach} 
Our goal is to exploit the highly promising learning capabilities of
transformer models \cite{vaswani2017attention} and introduce them to high-resolution image
synthesis up to the megapixel range.
Previous work \cite{parmar2018image, chen2020generative} which applied transformers to
image generation demonstrated promising results for images up to a size of $64
\times 64$ pixels but, due to the quadratically increasing cost in sequence
length, cannot simply be scaled to higher resolutions. 

High-resolution image synthesis requires a model
that understands the global composition of images, enabling it to
generate locally realistic as well as globally consistent patterns.
Therefore, instead of representing an image with pixels, we represent it
as a composition of perceptually rich image constituents from a codebook.  By
learning an effective code, as described in Sec.~\ref{sec:vqgan}, we can
significantly reduce the description length of compositions, which allows us to
efficiently model their global interrelations within images with a transformer
architecture as described in
Sec.~\ref{sec:trafo}. This approach, summarized in Fig.~\ref{fig:model}, is
able to generate realistic and consistent high resolution
images both in an unconditional and a conditional setting.

\newcommand{\codebookdim}{n_z}
\subsection{Learning an Effective Codebook of Image Constituents for Use in Transformers}
\label{sec:vqgan}
\noindent To utilize the highly expressive transformer architecture for image synthesis, we need to 
express the constituents of an image in the form of a \emph{sequence}.
Instead of building on individual pixels, complexity necessitates
an approach that uses a discrete codebook of learned representations,
such that any image $x \in \RR^{H \times W \times 3}$ can be represented by a
spatial collection of codebook entries $\quantizedcode \in \RR^{h \times w
\times \codebookdim}$, where $\codebookdim$ is the dimensionality of codes.
An equivalent representation is a sequence of $h\cdot w$ indices which specify
the respective entries in the learned codebook.
To effectively learn such a discrete spatial codebook, we propose to directly
incorporate the inductive biases of 
CNNs and incorporate ideas from neural discrete representation learning \cite{oord2018neural}.
First, we learn a convolutional model
consisting of an encoder $\encoder$ and a decoder $\decoder$, such that taken
together, they learn to represent images with codes from a
learned, discrete codebook $\codebook = \{z_k\}_{k=1}^K \subset
\RR^\codebookdim$ (see Fig.~\ref{fig:model} for an overview).
More precisely, we approximate a given image $x$ by $\hat{x}=G(\quantizedcode)$. We
obtain $\quantizedcode$ using the encoding $ \hat{z} = \encoder(x) \in \RR^{h\times w \times
\codebookdim}$ and a subsequent element-wise quantization $\quantize(\cdot)$ of each spatial code
$\hat{z}_{ij} \in \RR^\codebookdim$ onto its closest codebook entry $z_k$:
\begin{equation}
  \quantizedcode = \quantize(\hat{z}) \coloneqq
  \left(\argmin_{z_k \in \codebook} \Vert \hat{z}_{ij} - z_k \Vert\right)
  \in \RR^{h\times w \times \codebookdim}.
\end{equation} 
The reconstruction $\hat{x} \approx x$ is then given by 
\begin{equation}
  \hat{x} = \decoder(\quantizedcode ) = \decoder\left(
    \quantize(\encoder(x)) \right).
\label{eq:vqrec}
\end{equation}
Backpropagation through the non-differentiable quantization operation in
Eq.~\eqref{eq:vqrec} is achieved by a straight-through gradient estimator,
which simply copies the gradients from the decoder to the encoder
\cite{bengio2013estimating}, such that the model and codebook can be trained end-to-end
via the loss function
\begin{align}
\mathcal{L}_{\text{VQ}}(\encoder, \decoder, \codebook) = \Vert x - \hat{x} \Vert^2 
  &+ \Vert \text{sg}[\encoder(x)] - \quantizedcode \Vert_2^2 \nonumber \\ &+ \Vert \text{sg}[\quantizedcode] - \encoder(x) \Vert_2^2.
\label{eq:origvqloss}
\end{align}
Here, $\mathcal{L}_{\text{rec}}=\Vert x - \hat{x} \Vert^2$ is a reconstruction loss,
$\text{sg}[\cdot]$ denotes the stop-gradient operation, and
$\Vert \text{sg}[\quantizedcode] - \encoder(x) \Vert_2^2$ is the so-called
``commitment loss'' \cite{oord2018neural}.

\paragraph{Learning a Perceptually Rich Codebook}
Using transformers to represent images as a distribution over latent image constituents requires us to push the limits of compression and learn a rich codebook. 
To do so, we propose \emph{VQGAN}, a variant of the original VQVAE,
and use a discriminator and perceptual loss
\cite{larsen2015autoencoding,johnson2016perceptual,lamb2016discriminative,dosovitskiy2016generating,mentzer2020highfidelity} to keep good perceptual quality at
increased compression rate. Note that this is in contrast to previous works which applied pixel-based
\cite{oord2016conditional,razavi2019generating} and transformer-based
autoregressive models \cite{chen2020generative} on top of only a shallow quantization model.
More specifically, we replace the $L_2$ loss
used in \cite{oord2018neural}
for $\mathcal{L}_{\text{rec}}$
by a perceptual loss and introduce an adversarial
training procedure with a patch-based discriminator $\discriminator$
\cite{isola2017image} that aims to differentiate between real and reconstructed images:
\begin{align}
\mathcal{L}_{\text{GAN}}(\{\encoder, \decoder, \codebook \}, \discriminator) =
  \left[ \log \discriminator(x) + \log (1 - \discriminator(\hat{x}))\right]
\label{eq:ourvqloss}
\end{align}
The complete objective for finding the optimal compression model $\mathcal{Q}^* = \{\encoder^*, \decoder^*, \codebook^* \}$ then reads
\begin{align}
\mathcal{Q}^* = \argmin_{\encoder, \decoder, \codebook} \max_{\discriminator}
  \mathbb{E}_{x\sim p(x)} \Big[
    &\mathcal{L}_{\text{VQ}}(\encoder, \decoder, \codebook) \nonumber \\
    + \lambda &\mathcal{L}_{\text{GAN}}(\{\encoder, \decoder, \codebook \}, \discriminator)
    \Big],
\end{align}
where we compute the adaptive weight $\lambda$ according to
\begin{equation}
\lambda = \frac{\nabla_{\decoder_L}[ \mathcal{L}_{\text{rec}} ]}{\nabla_{\decoder_L} [\mathcal{L}_{\text{GAN}}] + \delta}
\end{equation}
where $\mathcal{L}_{\text{rec}}$ is the perceptual reconstruction loss
\cite{zhang2018perceptual}, $\nabla_{\decoder_L}[\cdot]$ denotes the gradient of its input
w.r.t.\ the last layer $L$ of the decoder, and $\delta = 10^{-6}$ is used for numerical stability. To aggregate context from everywhere, we apply a single attention layer on the lowest resolution. This training procedure significantly reduces the sequence length when unrolling the latent code and thereby enables the application of powerful transformer models.

\subsection{Learning the Composition of Images with Transformers}
\label{sec:trafo}
\paragraph{Latent Transformers} With $\encoder$ and $\decoder$ available, we can now represent images in terms of the codebook-indices of
their encodings. More precisely, the quantized encoding of an image $x$ is given by
$\quantizedcode = \quantize(\encoder(x)) \in \RR^{h \times w \times
\codebookdim}$ and is equivalent to
a sequence $s \in \{0, \dots, \vert \codebook \vert - 1 \}^{h \times w}$ of indices from the codebook,
which is obtained
by replacing each code by its index in the codebook $\codebook$:
\begin{equation}
  s_{ij} = k \text{ such that } \left( \quantizedcode \right)_{ij} = z_k.
\label{eq:indexmap}
\end{equation}
By mapping indices of a sequence $s$ back to their corresponding codebook entries,
$\quantizedcode = \left( z_{s_{ij}} \right)$ is readily recovered and decoded to an image $\hat{x} =
\decoder(\quantizedcode)$.

Thus, after choosing some ordering of the indices in $s$, image-generation can be formulated as
autoregressive next-index prediction: Given indices
$s_{<i}$, the transformer learns to predict the distribution of possible next
indices, i.e. $p(s_i \vert s_{<i})$ to compute the likelihood of the full representation as
$p(s)=\prod_i p(s_i \vert s_{<i})$.
This allows us to directly maximize the log-likelihood of the data
representations:
\begin{equation}
  \mathcal{L}_{\text{Transformer}} = \mathbb{E}_{x \sim p(x)} \left[ -\log p(s) \right].
\end{equation}

\paragraph{Conditioned Synthesis}
In many image synthesis tasks a user demands control over the generation
process by providing additional information from which an example shall be
synthesized.
This information, which we will call $c$, could be a single label describing the
overall image class or even another image itself.
The task is then to learn the likelihood of the sequence given this information $c$:
\begin{equation}
  \label{eq:conddensity}
  p(s|c) = \prod_i p(s_i|s_{<i}, c).
\end{equation}
If the conditioning information $c$ has spatial extent, we first learn another \emph{VQGAN} to obtain
again an index-based representation $r \in \{0, \dots, \vert \codebook_c \vert
-1 \}^{h_c \times w_c}$ with the newly obtained codebook $\codebook_c$
Due to the autoregressive structure of the transformer, we can then simply prepend $r$ to $s$ and
restrict the computation of the negative log-likelihood to entries
$p(s_i \vert s_{<i}, r)$. This ``decoder-only'' strategy has also been
successfully used for text-summarization tasks \cite{liu2018generating}.

\paragraph{Generating High-Resolution Images}
\attentionslide
The attention mechanism of the transformer puts limits on the sequence length
$h \cdot w$ of its inputs $s$. While we can adapt the number of downsampling
blocks $m$ of our \emph{VQGAN} to reduce images of size $H\times W$ to $h=H/2^m
\times w=W/2^m$, we observe degradation of the reconstruction quality beyond a
critical value of $m$, which depends on the considered dataset. To generate
images in the megapixel regime, we therefore have to work patch-wise and crop images to restrict the length of $s$ to a maximally
feasible size during training. To sample images, we then use the transformer in a
sliding-window manner as illustrated in Fig.~\ref{fig:attentionslide}.
Our \emph{VQGAN} ensures that the available context is still
sufficient to faithfully model images, as long as either the statistics of the
dataset are approximately spatially invariant or spatial conditioning
information is available. In practice, this is not a restrictive requirement,
because when it is violated, \ie unconditional image synthesis on
aligned data, we can simply condition on image coordinates, similar to
\cite{lin2019cocogan}.

\vspace{-0.5em}
\section{Experiments}
\label{sec:experiments}
\enlargethispage{1\baselineskip}
\vspace{-0.5em}
\noindent This section evaluates the ability of our approach to retain the
advantages of transformers over their convolutional counterparts
(Sec.~\ref{sec:psnailvstrafo}) while integrating the effectiveness of
convolutional architectures to enable high-resolution image synthesis
(Sec.~\ref{sec:four}). Furthermore, in Sec.~\ref{sec:wurst}, we investigate how codebook quality affects our approach.
We close the analysis by providing a quantitative comparison to a wide range of existing approches for generative image synthesis in Sec.~\ref{subsec:quantitative}. 
Based on initial experiments, we usually set $\vert
\codebook \vert=1024$ and train all subsequent transformer models to predict
sequences of length $16\cdot 16$, as this is the maximum feasible length to
train a GPT2-medium architecture (307 M parameters)
\cite{Radford2019LanguageMA} on a GPU with 12GB VRAM. More details on architectures and hyperparameters can be found in the appendix (Tab.~\ref{tab:vqaearchitecture} and Tab.~\ref{tab:hyperparameters}).

\subsection{Attention Is All You Need in the Latent Space}
\label{sec:psnailvstrafo}

\noindent Transformers show state-of-the-art results on a wide variety of tasks,
including autoregressive image modeling. However, evaluations of previous works
were limited to transformers working directly on (low-resolution) pixels
\cite{parmar2018image,child2019generating,ho2019axial}, or to deliberately shallow pixel encodings \cite{chen2020generative}.
This raises the question if our approach retains
the advantages of transformers over convolutional approaches.

To answer this question, we use a variety of conditional and unconditional
tasks and compare the performance between our transformer-based approach and a
convolutional approach. For each task, we train a \emph{VQGAN} with $m=4$
downsampling blocks, and, if needed,
another one for the conditioning information,
and then train both a transformer and a PixelSNAIL \cite{chen2018pixelsnail} model on the same
representations, as the latter has been used
in previous state-of-the-art two-stage approaches \cite{razavi2019generating}.
For a thorough comparison, we vary the model capacities between
85M and 310M parameters and adjust the number of layers in each model to match
one another. We observe that PixelSNAIL trains roughly twice as fast as the
transformer and thus, for a fair comparison, report the negative log-likelihood
both for the same amount of training time (\emph{P-SNAIL time}) and for the same amount of training
steps (\emph{P-SNAIL steps}).
\vspace{-1.1em}
\paragraph{Results} Tab.~\ref{tab:gptvspixelsnail} reports results
for unconditional image modeling on \emph{ImageNet} (IN) \cite{deng2009imagenet}, \emph{Restricted ImageNet} (RIN) \cite{santurkar2019computer},
consisting of a subset of animal classes from ImageNet, \emph{LSUN Churches and
Towers} (LSUN-CT) \cite{yu15lsun}, and for conditional image modeling of RIN
conditioned on depth maps obtained with the approach of \cite{Ranftl2020} (D-RIN) and of landscape images
collected from Flickr conditioned on semantic layouts (S-FLCKR) obtained with the
approach of \cite{chen2017deeplab}. Note that for the semantic layouts, we train the
first-stage using a cross-entropy reconstruction loss due to their discrete
nature.
The results shows that the transformer consistently
outperforms PixelSNAIL across all tasks when trained for the same amount of
time and the gap increases even further when trained for the same number of
steps. These results demonstrate that gains of transformers carry over to our
proposed two-stage setting.
\pixelsnailvstransformersmall

\subsection{A Unified Model for Image Synthesis Tasks}
\label{sec:four}
\firstpagefigurehalf
\noindent
The versatility and generality of the transformer architecture makes it a
promising candidate for image synthesis. In the conditional case, additional
information $c$ such as class labels or segmentation maps are used and the goal
is to learn the distribution of images as described in
Eq.~\eqref{eq:conddensity}. Using the same setting as in
Sec.~\ref{sec:psnailvstrafo} (i.e.\ image size $256 \times 256$, latent size
$16 \times 16$), we perform various conditional image synthesis experiments:
\enlargethispage{1\baselineskip}

\noindent (i): \textbf{Semantic image synthesis}, where we condition on semantic
segmentation masks of ADE20K \cite{zhou2016semantic}, a web-scraped
landscapes dataset (S-FLCKR) and COCO-Stuff \cite{caesar2018cvpr}.
Results are depicted in Figure~\ref{fig:firstpage},~\ref{fig:bigsamples} and Fig.~\ref{fig:convcondplot}.

\noindent (ii): \textbf{Structure-to-image}, where we use either depth or edge
information to synthesize images from both RIN and IN (see
Sec.~\ref{sec:psnailvstrafo}). The resulting depth-to-image and edge-to-image translations
are visualized in Fig.~\ref{fig:firstpage} and Fig.~\ref{fig:convcondplot}.

\noindent (iii): \textbf{Pose-guided synthesis:} Instead of using the semantically
rich information of either segmentation or depth maps, Fig.~\ref{fig:firstpage}
shows that the same approach as for the previous experiments can be used to
build a shape-conditional generative model on the DeepFashion
\cite{liuLQWTcvpr16DeepFashion} dataset.

\enlargethispage{1\baselineskip}
\noindent (iv): \textbf{Stochastic superresolution}, where low-resolution images
serve as the conditioning information and are thereby upsampled. We train our model for an 
upsampling factor of 8 on ImageNet and show results in
Fig.~\ref{fig:convcondplot}. 

\noindent (v): \textbf{Class-conditional image synthesis:} Here, the conditioning
information $c$ is a single index describing the class label of interest.
Results for the RIN and IN dataset are
demonstrated in Fig.~\ref{fig:firstpage} and Fig.~\ref{fig:morecinsamples}, respectively. %

\noindent All of these examples make use of the same methodology. Instead of requiring
task specific architectures or modules, the flexibility of the transformer
allows us to learn appropriate interactions for each task, while the
\emph{VQGAN} --- which can be \emph{reused}
across different tasks --- leads to short sequence lengths. In combination, the
presented approach can be understood as an efficient, general purpose mechanism
for conditional image synthesis. Note that additional results for each experiment can be found in the appendix, Sec.~\ref{sec:addition}.
\paragraph{High-Resolution Synthesis} The sliding window approach introduced in Sec.~\ref{sec:trafo}
enables image synthesis beyond a resolution of $256 \times 256$ pixels.
We evaluate this approach on unconditional image generation on LSUN-CT and
FacesHQ (see Sec.~\ref{sec:wurst}) and  conditional synthesis on D-RIN,
COCO-Stuff and S-FLCKR, where we show results in
Fig.~\ref{fig:frontpage},~\ref{fig:convcondplot} and the supplementary (Fig.~\ref{fig:bigsamplesbig}-\ref{fig:unconditionalchurches}). Note that this approach can in principle be used to generate images of arbitrary ratio and size, given that the image statistics of the dataset of interest are approximately spatially invariant or spatial information is available. 
Impressive results can be achieved by applying this method to image generation from semantic layouts on S-FLCKR, where a strong \emph{VQGAN} can be learned with $m=5$, so that its codebook together with the conditioning information provides the transformer with enough context for image generation in the megapixel regime. 
\bigsamples
\convcondplot
\enlargethispage{1\baselineskip}
\subsection{Building Context-Rich Vocabularies}
\label{sec:wurst}
\fstudyfaces
\noindent
How important are context-rich vocabularies? To investigate this question, we
ran experiments where the transformer architecture is kept fixed while the
amount of context encoded into the representation of the first stage is varied
through the number of downsampling blocks of our \emph{VQGAN}.
We specify the amount of context encoded in terms of reduction factor in the
side-length between image inputs and the resulting representations, \ie a first
stage encoding images of size $H \times W$ into discrete codes of size $H/f \times W/f$ is denoted by a factor $f$. For $f=1$, we reproduce the
approach of \cite{chen2020generative} and replace our \emph{VQGAN} by a k-means clustering of RGB
values with $k=512$. \\
During training, we always crop images to obtain inputs of size $16 \times 16$
for the transformer, \ie when modeling images with a
factor $f$ in the first stage, we use crops of size $16f \times 16f$. 
To sample from the models, we
always apply them in a sliding window manner as described in
Sec.~\ref{sec:approach}.

\noindent \textbf{Results} Fig.~\ref{fig:fstudyfaces} shows results for unconditional
synthesis of faces on \emph{FacesHQ}, the combination of \emph{CelebA-HQ}
\cite{karras2018progressive} and \emph{FFHQ} \cite{karras2019stylebased}. It clearly demonstrates the benefits of
powerful \emph{VQGAN}s by increasing the effective receptive field of the
transformer. For small receptive fields, or equivalently small $f$, the model
cannot capture coherent structures. For an intermediate value of $f=8$, the
overall structure of images can be approximated, but inconsistencies of facial
features such as a half-bearded face and of viewpoints in different parts of
the image arise. Only our full setting of $f=16$ can synthesize high-fidelity
samples. For analogous results in the conditional setting on S-FLCKR, we refer
to the appendix (Fig.~\ref{fig:rfwfstudy} and Sec.~\ref{supp:contextrich}).

\noindent
To assess the effectiveness of our approach quantitatively, we compare results between
training a transformer directly on pixels, and training it on top of a
\emph{VQGAN}'s
latent code with $f=2$, given a fixed computational budget. Again, we follow
\cite{chen2020generative} and learn a dictionary of $512$ RGB values on
CIFAR10 to operate directly on pixel space and 
train the same transformer architecture on top of our
\emph{VQGAN} with a latent code of size $16\times16=256$. We
observe improvements of $18.63\%$ for FIDs and $14.08\times$ faster
sampling of images.

\subsection{Benchmarking Image Synthesis Results}
\label{subsec:quantitative}
\semanticsynthesis
\noindent In this section we investigate how our approach quantitatively compares to existing models for generative image synthesis. 
In particular, we assess the performance of our model in terms of FID and compare to a variety of established
models (GANs, VAEs, Flows, AR, Hybrid). The results on \emph{semantic synthesis}
are shown in Tab.~\ref{tab:semanticsynth}, where we compare to
\cite{park2019semantic, wang2018pix2pixHD, katiyar2021improving, ChenK17}, and
the results on \emph{unconditional face synthesis} are shown in
Tab.~\ref{tab:facesynth}.
While some task-specialized GAN models report better FID scores, our approach
provides a unified model that works well across a wide range of tasks while
retaining the ability to encode and reconstruct images. It thereby bridges the
gap between purely adversarial and likelihood-based approaches.

\noindent Autoregressive models are typically sampled with
a decoding strategy \cite{DBLP:conf/iclr/HoltzmanBDFC20} such as beam-search, top-k or nucleus sampling.
For most of our results, including those in Tab.~\ref{tab:semanticsynth}, we use
top-k sampling with $k=100$ unless stated otherwise. For the results on face
synthesis in Tab.~\ref{tab:facesynth}, we computed scores for
$k\in\{100,200,300,400,500\}$ and report the best results, obtained with
$k=400$ for CelebA-HQ and $k=300$ for FFHQ. Fig.~\ref{fig:facesfidandinceptionplots} in the
supplementary shows FID and Inception scores as a function of $k$.

\celebahq
\vspace{-1em}
\paragraph{Class-Conditional Synthesis on ImageNet}
\cinsynthesisupdate
\morecinsamples
\fidandinceptionplots
\noindent To address a direct comparison with the previous state-of-the-art
for autoregressive modeling of
class-conditional image synthesis on ImageNet, VQVAE-2 \cite{razavi2019generating}, 
we train a class-conditional ImageNet transformer
on $256 \times 256$ images, using a \emph{VQGAN} with $\dim \mathcal{Z}=16384$
and $f=16$, and additionally compare to BigGAN \cite{brock2018large}, IDDPM
\cite{nichol2021improved}, DCTransformer \cite{nash2021generating} and 
ADM \cite{dhariwal2021diffusion} in Tab.~\ref{tab:cinsynthesisupdate}. Note that
our model uses $\simeq 10 \times$ \emph{less} parameters than
VQVAE-2, which has an estimated parameter count of $13.5 \text{B}$ (estimate
based on \cite{seonghyeon2020}).

Samples of this model for different ImageNet classes are shown in Fig.~\ref{fig:morecinsamples}.
We observe that the adversarial training of the corresponding \emph{VQGAN} enables sampling of
high-quality images with realistic textures, of comparable or higher quality than existing approaches
such as BigGAN and VQVAE-2, see also Fig.~\ref{fig:cinqualtwo}-\ref{fig:cinqualone} in the supplementary. 

\noindent Quantitative results are summarized in Tab.~\ref{tab:cinsynthesisupdate}. We report FID and Inception Scores 
for the best $k$/$p$ in top-k/top-p sampling. Following \cite{razavi2019generating}, we can further
increase quality via classifier-rejection, which keeps only the best $m$-out-of-$n$ samples in terms of
the classifier's score, \ie with an acceptance rate of $\nicefrac{m}{n}$.
We use a ResNet-101 classifier \cite{DBLP:journals/corr/HeZRS15}.

We observe that our model outperforms other autoregressive approaches (VQVAE-2, DCTransformer)
in terms of FID and IS, surpasses BigGAN and IDDPM even for low rejection rates and 
yields scores close to the state of the art for higher rejection rates, see also Fig.~\ref{fig:fidandinceptionplots}. %

\vspace{-1.0em}
\paragraph{How good is the \emph{VQGAN}?}
\imagenetreconstruction
Reconstruction FIDs obtained via the codebook provide an estimate on the
achievable FID of the generative model trained on it.
To quantify the performance gains of our \emph{VQGAN} over discrete VAEs 
trained without perceptual and adversarial losses (\eg VQVAE-2, DALL-E \cite{ramesh2021zeroshot}), 
we evaluate this metric on ImageNet and report results in
Tab.~\ref{tab:recfidimagenet}.
Our \emph{VQGAN} outperforms non-adversarial models while providing
significantly more compression (seq. length of $256$ vs. $5120=32^2+64^2$ for VQVAE-2, $256$ vs $1024$ for DALL-E). 
As expected, larger versions of \emph{VQGAN} (either in terms of larger codebook sizes or
increased code lengths) further improve performance. Using the same hierarchical codebook setting as in VQVAE-2 with our model provides the best reconstruction FID, albeit at the cost of a very long and thus impractical sequence. 
\noindent
The qualitative comparison corresponding to the results in
Tab.~\ref{tab:recfidimagenet} can be found in Fig.~\ref{fig:vqvaegan}.

\section{Conclusion}
\noindent This paper adressed the fundamental challenges that previously confined transformers to
low-resolution images. We
proposed an approach which represents images as a composition of perceptually
rich image constituents and thereby overcomes the infeasible quadratic
complexity when modeling images directly in pixel space. Modeling constituents
with a CNN architecture and their compositions with a transformer architecture
taps into the full potential of their complementary strengths and thereby
allowed us to represent the first results on high-resolution image synthesis
with a transformer-based architecture. In experiments, our approach
demonstrates the efficiency of convolutional inductive biases and the
expressivity of transformers by synthesizing images in the megapixel range and
outperforming state-of-the-art convolutional approaches. Equipped with a
general mechanism for conditional synthesis, it offers many
opportunities for novel neural rendering approaches.
\blfootnote{\noindent This work has been supported by the German Research Foundation
(DFG) projects 371923335, 421703927 and a hardware donation from NVIDIA
corporation.}

\newpage
\clearpage
\appendix
\onecolumn

\begin{center}
  \textbf{
    \Large Taming Transformers for High-Resolution \\\vspace{0.3em} Image Synthesis} \\
  \Large 
-- \\
 \textbf{\large Supplementary Material} \\
\hspace{1cm}
\end{center}

\FloatBarrier

\noindent
The supplementary material for our work \emph{Taming Transformers for High-Resolution Image Synthesis} is structured as follows:
First, Sec.~\ref{sec:changelog} summarizes changes to a previous version of this
paper. In Sec.~\ref{sec:impldetails}, we present hyperparameters and
architectures which were used to train our models. Next, extending the
discussion of Sec.~\ref{sec:wurst}, Sec.~\ref{supp:contextrich} presents
additional evidence for the importance of perceptually rich codebooks and its
interpretation as a trade-off between reconstruction fidelity and sampling capability.
Additional results on high-resolution image synthesis for a wide range of tasks
are then presented in Sec.~\ref{sec:addition}, and Sec.~\ref{supp:nns} shows
nearest neighbors of samples. Finally, Sec.~\ref{sec:orderingsupp}
contains results regarding the ordering of image representations.

\section{Changelog}
\label{sec:changelog}
\cinsynthesis
\noindent We summarize changes between this version
\footnote{\url{https://arxiv.org/abs/2012.09841v3}}
of the paper and its previous version 
\footnote{\url{https://arxiv.org/abs/2012.09841v2}}.

In the previous version, Eq.~\eqref{eq:origvqloss} had a weighting term $\beta$
on the commitment loss, and Tab.~\ref{tab:hyperparameters} reported a value of
$\beta=0.25$ for all models. However, due to a bug in the implementation,
$\beta$ was never used and all models have been trained with $\beta=1.0$. Thus,
we removed $\beta$ in Eq.~\eqref{eq:origvqloss}.

We updated class-conditional synthesis results on ImageNet in
Sec.~\ref{subsec:quantitative}. The previous
results, included here in Tab.~\ref{tab:prevresults} for completeness, were based
on a slightly different implementation where the transformer did not predict
the distribution of the first token but used a histogram for it. The new model
has been trained for 2.4 million steps with a batch size of 16 accumulated over 8
batches, which took 45.8 days on a single A100 GPU. The previous model
had been trained for 1.0 million steps.
Furthermore, the FID values were based on 50k (18k) samples against 50k (18k)
training examples (to compare with MSP). For better comparison with other
works, the current version reports FIDs based on 50k samples against all
training examples of ImageNet using \texttt{torch-fidelity}
\cite{obukhov2020torchfidelity}. We updated all qualitative figures
showing samples from this model and added visualizations of the effect
of tuning top-$k$/$p$ or rejection rate in Fig.~\ref{fig:cinqualtwo}-\ref{fig:cinbatchfive}.

To provide a better overview, we also include results from works that became
available after the previous version of our work. Specifically, we include
results on reconstruction quality of the VQVAE from \cite{ramesh2021zeroshot}
in Tab.~\ref{tab:recfidimagenet} and Fig.~\ref{fig:vqvaegan} (which replaces
the previous qualitative comparison), and results on class-conditional ImageNet
sampling from
\cite{nichol2021improved,nash2021generating,dhariwal2021diffusion} in
Tab.~\ref{tab:cinsynthesisupdate}. Note that with the exception of BigGAN and
BigGAN-deep \cite{brock2018large}, no models or sampling results are available
for the methods we compare to in Tab.~\ref{tab:cinsynthesisupdate}. Thus, we
can only report the numbers from the respective papers but cannot re-evaluate
them with the same code. We follow the common evaluation protocol for
class-conditional ImageNet synthesis from \cite{brock2018large} and evaluate
50k samples from the model against the whole training split of ImageNet.
However, it is not clear how different implementations resize the training
images. In our code, we use the largest center-crop and resize it bilinearly
with anti-aliasing to $256 \times 256$ using \texttt{Pillow}
\cite{hugo_van_kemenade_2021_4659051}. FID and Inception Scores are then
computed with \texttt{torch-fidelity} \cite{obukhov2020torchfidelity}.

We updated face-synthesis results in Tab.~\ref{tab:facesynth} based on a
slightly different implementation as in the case of class-conditional ImageNet
results and improve the previous results slightly. In addition, we evaluate
the ability of our NLL-based training to detect overfitting. We train larger
models (FFHQ (big) and CelebA-HQ (big) in Tab.~\ref{tab:hyperparameters}) on
the face datasets, and show nearest neighbors of samples obtained from
checkpoints with the best NLL on the validation split and the training split in
Sec.~\ref{supp:nns}. We also added Fig.~\ref{fig:facesfidandinceptionplots}, which
visualizes the effect of tuning $k$ in top-k sampling on FID and IS.

\facesfidandinceptionplots

\section{Implementation Details}
\label{sec:impldetails}
\noindent The hyperparameters for all experiments presented in the main paper
and supplementary material can be found in Tab.~\ref{tab:hyperparameters}.
Except for the \emph{c-IN (big)}, \emph{COCO-Stuff} and \emph{ADE20K} models, these hyperparameters are set such that
each transformer model can be trained with a batch-size of at least 2 on a GPU
with 12GB VRAM, but we generally train on 2-4 GPUs with an accumulated VRAM of
48 GB. If hardware permits, 16-bit precision training is enabled.
\paragraph{VQGAN Architecture} The architecture of our convolutional encoder and decoder models used in the \emph{VQGAN} experiments is described in Tab.~\ref{tab:vqaearchitecture}. Note that we adopt the compression rate by tuning the number of downsampling steps $m$.
Further note that $\lambda$ in Eq.~\ref{eq:ourvqloss} is set to zero in an initial warm-up phase. Empirically, we found that longer warm-ups generally lead to better reconstructions. As a rule of thumb, we recommend setting $\lambda = 0$ for at least one epoch.

\paragraph{Transformer Architecture}
Our transformer model is identical to the GPT2 architecture
\cite{Radford2019LanguageMA} and we vary its capacity mainly through varying the amount of layers (see Tab.~\ref{tab:hyperparameters}). 
Furthermore, we generally produce samples with a temperature $t=1.0$ and a top-$k$ cutoff at $k=100$ (with higher top-$k$ values for larger codebooks).
\vqarchitecture
\hyperparameters
\section{On Context-Rich Vocabularies}
\label{supp:contextrich}

\noindent
Sec.~\ref{sec:wurst} investigated the effect of the downsampling factor $f$ used
for encoding images. As demonstrated in Fig.~\ref{fig:fstudyfaces}, large
factors are crucial for our approach, since they enable the transformer to
model long-range interactions efficiently. However, since larger $f$ correspond
to larger compression rates, the reconstruction quality of the \emph{VQGAN} starts
to decrease after a certain point, which is analyzed in
Fig.~\ref{fig:quantfacefstudy}. The left part shows the reconstruction error
(measured by LPIPS \cite{zhang2018perceptual})
versus the negative log-likelihood obtained by the transformer for values of
$f$ ranging from $1$ to $64$. The latter provides a measure of the ability to
model the distribution of the image representation, which increases with $f$.
The reconstruction error on the other hand decreases with $f$ and the
qualitative results on the right part show that beyond a critical value of $f$,
in this case $f=16$, reconstruction errors become severe. At this point,
even when the image representations are modeled faithfully, as suggested by a
low negative log-likelihood, sampled images are of low-fidelity, because the
reconstruction capabilities provide an upper bound on the quality that can be
achieved.

Hence, Fig.~\ref{fig:quantfacefstudy} shows that we must learn perceptually
rich encodings, \ie encodings with a large $f$ and perceptually faithful
reconstructions. This is the goal of our \emph{VQGAN} and
Fig.~\ref{fig:vqvaegan} compares its reconstruction capabilities against the
VQVAE \cite{oord2018neural} used in DALL-E \cite{ramesh2021zeroshot}.  We
observe that for $f=8$ and $8192$ codebook entries, both the VQVAE and
\emph{VQGAN} capture the global structure faithfully. However, the textures
produced by the VQVAE are blurry, whereas those of the \emph{VQGAN} are crisp
and realistic looking (\eg the stone texture and the fur and tail of the
squirrel). When we increase the compression rate of the \emph{VQGAN} further to
$f=16$, we see that some reconstructed parts are not perfectly aligned with the
input anymore (\eg the paw of the squirrel), but, especially with slightly
larger codebooks, the reconstructions still look realistic. This demonstrates
how the \emph{VQGAN} provides high-fidelity reconstructions at large factors,
and thereby enables efficient high-resolution image synthesis with
transformers.

To illustrate how the choice of $f$ depends on the dataset,
Fig.~\ref{fig:rfwfstudy} presents results on S-FLCKR.
In the left part, it shows, analogous to Fig.~\ref{fig:fstudyfaces}, how the
quality of samples increases with increasing $f$. However, in the right part, it
shows that reconstructions remain faithful perceptually faithful even for
$f32$, which is in contrast to the corresponding results on faces
in Fig.~\ref{fig:quantfacefstudy}. These results might be explained by a higher
perceptual sensitivity to facial features as compared to textures, and allow us
to generate high-resolution landscapes even more efficiently with $f=32$.
\quantfacefstudy
\vqvaegan
\rfwfstudy

\section{Additional Results}
\label{sec:addition}

\paragraph{Qualitative Comparisons}
The qualitative comparison corresponding to Tab.~\ref{tab:cinsynthesisupdate} and Tab.~\ref{tab:prevresults} can
be found in Fig.~\ref{fig:cinqualtwo},~\ref{fig:cinqualthree},~\ref{fig:cinqualbirds} and
\ref{fig:cinqualone}. Since no models are available for VQVAE-2 and MSP, we
extracted results directly from the
supplementary\footnote{\url{https://drive.google.com/file/d/1H2nr_Cu7OK18tRemsWn_6o5DGMNYentM/view?usp=sharing}}
and from the provided samples\footnote{\url{https://bit.ly/2FJkvhJ}},
respectively. For BigGAN, we produced the samples via the provided
model\footnote{\url{https://tfhub.dev/deepmind/biggan-deep-256/1}}. 
Similarly, the qualitative comparison with the best competitor model (SPADE) for semantic synthesis on standard benchmarks (see Tab.~\ref{tab:semanticsynth}) can be found in Fig.~\ref{fig:ourvsspadeade} (ADE20K) and Fig.~\ref{fig:ourvsspadecoco} (COCO-Stuff)\footnote{samples were reproduced with the authors' official implementation available at \url{https://github.com/nvlabs/spade/}}. 

\paragraph{Comparison to Image-GPT}
To further evaluate the effectiveness of our approach, we compare to the state-of-the-art generative transformer model on images, ImageGPT \cite{chen2020generative}. By using immense amounts of compute the authors demonstrated that transformer models can be applied to the pixel-representation of images and thereby achieved impressive results both in representation learning and image synthesis. However, as their approach is confined to pixel-space, it does not scale beyond a resolution of $192 \times 192$.
As our approach leverages a strong compression method to obtain context-rich representations of images and \emph{then} learns a transformer model, we can synthesize images of much higher resolution. We compare both approaches in Fig.~\ref{fig:ourvsigpt} and Fig.~\ref{fig:ourvsigpttwo}, where completions of images are depicted. Both plots show that our approach is able to synthesize consistent completions of dramatically increased fidelity. The results of \cite{chen2020generative} are obtained from \url{https://openai.com/blog/image-gpt/}.

\cinqualitativetwo
\cinqualitativethree
\cinqualitativebirds
\cinqualitativeone

\rejectioncin
\topkcin
\toppcin

\cinbatchthree
\cinbatchsix
\cinbatchone
\cinbatchtwo
\cinbatchfour
\cinbatchfive

\ourvsigpt
\ourvsigpttwo

\paragraph{Additional High-Resolution Results}
Fig.~\ref{fig:bigsamplesbig},~\ref{fig:bigsamplesbigtwo},~\ref{fig:bigsamplesbigthree} and Fig.~\ref{fig:bigsamplesbigvertical} contain additional HR results on the S-FLCKR dataset for both $f=16$ ($m=4$) and $f=32$ ($m=5$) (semantically guided). In particular, we provide an enlarged version of Fig.~\ref{fig:bigsamples} from the main text, which had to be scaled down due to space constraints. Additionally, we use our sliding window approach (see Sec.~\ref{sec:approach}) to produce high-resolution samples for the depth-to-image setting on RIN in Fig.~\ref{fig:convdepth} and Fig.~\ref{fig:convdepthtwo}, edge-to-image on IN in Fig.~\ref{fig:convedges}, stochastic superresolution on IN in Fig.~\ref{fig:convsr}, more examples on semantically guided landscape synthesis on S-FLCKR in Fig.~\ref{fig:convf16landscapes} with $f=16$ and in Fig.~\ref{fig:convf32landscapes} with $f=32$, and unconditional image generation on LSUN-CT (see Sec.~\ref{sec:psnailvstrafo}) in Fig.~\ref{fig:unconditionalchurches}. Moreover, for images of size $256 \times 256$, we provide results for generation from semantic layout on (i) ADE20K in Fig.~\ref{fig:ourvsspadeade} and (ii) COCO-Stuff in Fig.~\ref{fig:ourvsspadecoco}, depth-to-image on IN in Fig.~\ref{fig:depthinsamples}, pose-guided person generation in Fig.~\ref{fig:dfsamples} and class-conditional synthesis on RIN in Fig.~\ref{fig:rinclasscond}.
\bigsamplesbig
\bigsamplesbigtwo
\bigsamplesbigthree
\bigsamplesbigvertical
\convdepth
\convdepthtwo
\artsyedges
\convsr
\convlandscapessixteen
\convlandscapesthirtytwo

\unconditionalchurches

\ourvsspadeade
\ourvsspadecoco

\indepthsamples
\sticktoreal
\rinclasscond

\section{Nearest Neighbors of Samples}
\label{supp:nns}
\noindent One advantage of likelihood-based generative models over, \eg, GANs is the
ability to evaluate NLL on training data and validation data to detect
overfitting. To test this, we trained large models for face synthesis, which
can easily overfit them, and retained two checkpoints on each dataset: One for
the best validation NLL (at the 10th and 13th epoch for FFHQ and CelebA-HQ,
respectively), and another for the best training NLL (at epoch 1000). We then
produced samples from both checkpoints and retrieved nearest neighbors from the
training data based on the LPIPS similarity metric \cite{zhang2018perceptual}. The
results are shown in Fig.~\ref{fig:facesnns}, where it can be observed that the
checkpoints with best training NLL (best train NLL) reproduce the training
examples, whereas samples from the checkpoints with best validation NLL (best
val. NLL) depict new faces which are not found in the training data.

Based on these results, we can conclude that early-stopping based on validation
NLL can prevent overfitting. Furthermore, the bottleneck for our approach on
face synthesis is given by the dataset size since it has the capacity to almost
perfectly fit the training data. Unfortunately, FID scores cannot detect such
an overfitting. Indeed, the best train NLL checkpoints achieve FID scores of
3.86 on CelebA-HQ and 2.68 on FFHQ, compared to 10.2 and 9.6 for the best val.
NLL checkpoints. While validation NLL provides a way to detect overfitting for
likelihood-based models, it is not clear if early-stopping based on it is
optimal if one is mainly interested in the quality of samples. To address this
and the evaluation of GANs, new metrics will be required which can
differentiate between models that produce new, high-quality samples and those
that simply reproduce the training data.

Our class-conditional ImageNet model does not display overfitting according to
validation NLL, and the nearest neighbors shown in Fig.~\ref{fig:cinneighbors}
also provide evidence that the model produces new, high-quality samples.

\facesnns
\cinneighbors

\section{On the Ordering of Image Representations}
\label{sec:orderingsupp}

\noindent For the ``classical'' domain of transformer models, NLP, the order
of tokens is defined by the language at hand. For images and their discrete
representations, in contrast, it is not clear which linear ordering to use.
In particular, our sliding-window approach depends on a row-major ordering and
we thus investigate the performance of the following five different
permutations of the input sequence of codebook indices: (i) \textbf{row
major}, or \emph{raster scan order}, where the image representation
is unrolled from top left to bottom right.
(ii) \textbf{spiral out}, which incorporates the prior assumption that most images show a
\emph{centered} object.
(iii) \textbf{z-curve}, also known as \emph{z-order} or
\emph{morton curve}, which introduces the
prior of \emph{preserved locality} when mapping a 2D image
representation onto a 1D sequence.
(iv) \textbf{subsample}, where prefixes correspond to subsampled
representations, see also \cite{MenickK19}.
(v) \textbf{alternate}, which is related to \emph{row major}, but alternates the
direction of unrolling every row.
(vi) \textbf{spiral in}, a reversed version of \emph{spiral out} which provides
the most context for predicting the center of the image.
A graphical visualization of these permutation variants is shown in Fig.~\ref{fig:permutations}. Given a \emph{VQGAN} trained on ImageNet, we train a transformer for each permutation in a controlled setting, i.e.\
we fix initialization and computational budget.

\paragraph{Results} Fig.\ref{fig:permutations} depicts the evolution of negative log-likelihood for
each variant as a function of training iterations, with final values given by
(i) 4.767, (ii) 4.889, (iii) 4.810, (iv) 5.015, (v) 4.812, (vi) 4.901.
Interestingly, \emph{row major} performs best in terms of this metric, whereas the more hierarchical
\emph{subsample} prior does not induce any helpful bias.
We also include qualitative samples in Fig.~\ref{fig:permutationsbatches} and
observe that the two worst performing models in terms of NLL (\emph{subsample} and \emph{spiral in}) tend to produce more
textural samples, while the other variants synthesize samples with much more
recognizable structures. Overall, we can conclude that the autoregressive codebook modeling
is \emph{not} permutation-invariant, but the common \emph{row major}
ordering \cite{oord2016conditional,chen2020generative} outperforms other
orderings.

\permutations
\permuationsbatches

\newpage
\clearpage
{\small
\bibliographystyle{ieee_fullname}
\bibliography{ms}
}
\appendix
\onecolumn

\FloatBarrier

\end{document}

%% file: cvpr_figures.tex
\providecommand{\impath}[1]{}
\providecommand{\impatha}[1]{}
\providecommand{\impathb}[1]{}
\providecommand{\impathc}[1]{}
\providecommand{\impathd}[1]{}
\providecommand{\impathe}[1]{}
\providecommand{\imwidth}{}
\providecommand{\imwidtha}{}
\providecommand{\imwidthb}{}

\newcommand{\cinneighbors}{
\begin{figure*}
\centering
\renewcommand{\impath}[1]{img/qualitative_cin/nn/##1} 
\begin{adjustbox}{max width=0.825\linewidth, max height=0.725\textheight}
\begin{tabular}{c c}
\toprule
\multirow{4}{*}{\rotatebox[origin=c]{90}{$k=250$, $p=1.0$, $a=1.0$}} 
 & \includegraphics[width=0.8\textwidth, align=c]{\impath{topk250_9}} \\
 & \includegraphics[width=0.8\textwidth]{\impath{topk250_90}} \\
 & \includegraphics[width=0.8\textwidth]{\impath{topk250_200}} \\
 & \includegraphics[width=0.8\textwidth]{\impath{topk250_985}} \\
\midrule

\multirow{4}{*}{\rotatebox[origin=c]{90}{$k=973$, $p=0.88$, $a=1.0$}} 
 & \includegraphics[width=0.8\textwidth]{\impath{topp088_9}} \\
 & \includegraphics[width=0.8\textwidth]{\impath{topp088_90}} \\
 & \includegraphics[width=0.8\textwidth]{\impath{topp088_200}} \\
 & \includegraphics[width=0.8\textwidth]{\impath{topp088_985}} \\
\midrule 
 
\multirow{4}{*}{\rotatebox[origin=c]{90}{mixed $k$, $p=1.0$, $a=0.05$}} 
 & \includegraphics[width=0.8\textwidth]{\impath{a_01_9}} \\
 & \includegraphics[width=0.8\textwidth]{\impath{a_01_90}} \\
 & \includegraphics[width=0.8\textwidth]{\impath{a_01_200}} \\
 & \includegraphics[width=0.8\textwidth]{\impath{a_01_985}} \\
\midrule

\multirow{4}{*}{\rotatebox[origin=c]{90}{mixed $k$, $p=1.0$, $a=0.005$}} 
 & \includegraphics[width=0.8\textwidth]{\impath{a_10_9}} \\
 & \includegraphics[width=0.8\textwidth]{\impath{a_10_90}} \\
 & \includegraphics[width=0.8\textwidth]{\impath{a_10_200}} \\
 & \includegraphics[width=0.8\textwidth]{\impath{a_10_985}} \\
\bottomrule

\end{tabular}
\end{adjustbox}
\caption{\label{fig:cinneighbors} Nearest neighbors for our class-conditional ImageNet model ($256 \times 256$ pix), based on the LPIPS \cite{zhang2018unreasonable} distance. The left column shows a sample from our model, while the 
10 examples to the right show the nearest neighbors from the corresponding class (increasing distance) in the training dataset. Our model produces new, unseen high-quality images, not present in the training data.}
\end{figure*}
}

\newcommand{\cinbatchone}{
\begin{figure*}
\centering
\renewcommand{\impath}[1]{img/qualitative_cin/batches/##1} 
\begin{adjustbox}{max width=\linewidth, max height=\textheight}
\includegraphics[width=\textwidth]{\impath{topk_250_a1_0}}
\end{adjustbox}
\caption{\label{fig:cinbatchone} Random samples on $256 \times 256$ class-conditional ImageNet with $k=250$, $p=1.0$, acceptance rate 1.0. FID: 15.98, IS: $78.6 \pm 1.1$. Please see \url{https://git.io/JLlvY} for an uncompressed version.}
\end{figure*}
}

\newcommand{\cinbatchtwo}{
\begin{figure*}
\centering
\renewcommand{\impath}[1]{img/qualitative_cin/batches/##1} 
\begin{adjustbox}{max width=\linewidth, max height=\textheight}
\includegraphics[width=\textwidth]{\impath{topp_088_a1_0}}
\end{adjustbox}
\caption{\label{fig:cinbatchtwo} Random samples on $256 \times 256$ class-conditional ImageNet with $k=973$, $p=0.88$, acceptance rate 1.0. FID: 15.78, IS: $74.3 \pm 1.8$. Please see \url{https://git.io/JLlvY} for an uncompressed version.}
\end{figure*}
}

\newcommand{\cinbatchthree}{
\begin{figure*}
\centering
\renewcommand{\impath}[1]{img/qualitative_cin/batches/##1} 
\begin{adjustbox}{max width=\linewidth, max height=\textheight}
\includegraphics[width=\textwidth]{\impath{mixed_k_a1_0}}
\end{adjustbox}
\caption{\label{fig:cinbatchthree} Random samples on $256 \times 256$ class-conditional ImageNet with $k \in [100, 200, 250, 300, 350, 400, 500, 600, 800, 973]$
, $p=1.0$, acceptance rate 1.0. FID: 17.04, IS: $70.6 \pm 1.8$. Please see \url{https://git.io/JLlvY} for an uncompressed version.}
\end{figure*}
}

\newcommand{\cinbatchfour}{
\begin{figure*}
\centering
\renewcommand{\impath}[1]{img/qualitative_cin/batches/##1} 
\begin{adjustbox}{max width=\linewidth, max height=\textheight}
\includegraphics[width=\textwidth]{\impath{mixed_k_a0_005}}
\end{adjustbox}
\caption{\label{fig:cinbatchfour} Random samples on $256 \times 256$ class-conditional ImageNet with $k \in [100, 200, 250, 300, 350, 400, 500, 600, 800, 973]$
, $p=1.0$, acceptance rate 0.005. FID: 6.59, IS: $402.7 \pm 2.9$. Please see \url{https://git.io/JLlvY} for an uncompressed version.}
\end{figure*}
}

\newcommand{\cinbatchfive}{
\begin{figure*}
\centering
\renewcommand{\impath}[1]{img/qualitative_cin/batches/##1} 
\begin{adjustbox}{max width=\linewidth, max height=\textheight}
\includegraphics[width=\textwidth]{\impath{mixed_k_a0_05}}
\end{adjustbox}
\caption{\label{fig:cinbatchfive} Random samples on $256 \times 256$ class-conditional ImageNet with $k \in [100, 200, 250, 300, 350, 400, 500, 600, 800, 973]$
, $p=1.0$, acceptance rate 0.05. FID: 5.88, IS: $304.8 \pm 3.6$. Please see \url{https://git.io/JLlvY} for an uncompressed version.}
\end{figure*}
}

\newcommand{\cinbatchsix}{
\begin{figure*}
\centering
\renewcommand{\impath}[1]{img/qualitative_cin/batches/##1} 
\begin{adjustbox}{max width=\linewidth, max height=\textheight}
\includegraphics[width=\textwidth]{\impath{topk_600_a0_05}}
\end{adjustbox}
\caption{\label{fig:cinbatchsix} Random samples on $256 \times 256$ class-conditional ImageNet with $k=600$, $p=1.0$, acceptance rate 0.05. FID: 5.20, IS: $280.3 \pm 5.5$. Please see \url{https://git.io/JLlvY} for an uncompressed version.}
\end{figure*}
}

\newcommand{\morecinsamples}{
\begin{figure*}[thbp]
	\centering
    \renewcommand{\imwidth}{0.142\textwidth}
    \renewcommand{\impath}[1]{img/qualitative_cin/moreours/##1} 
    \setlength{\tabcolsep}{0pt}
    \renewcommand{\arraystretch}{0}
	\begin{small}
 	\begin{tabular}{ccccccc}
	\includegraphics[width=\imwidth]{\impath{000841}} &	        
	\includegraphics[width=\imwidth]{\impath{000232}} &	        
	\includegraphics[width=\imwidth]{\impath{000926}} &	        
	\includegraphics[width=\imwidth]{\impath{000215}} &	        
	\includegraphics[width=\imwidth]{\impath{000005}} &	        
	\includegraphics[width=\imwidth]{\impath{000006}} &	 
	\includegraphics[width=\imwidth]{\impath{000027}} \\
	
	\includegraphics[width=\imwidth]{\impath{000029}} &	        
	\includegraphics[width=\imwidth]{\impath{000023}} &	        
	\includegraphics[width=\imwidth]{\impath{000014}} &	        
	\includegraphics[width=\imwidth]{\impath{000751}} &	        
	\includegraphics[width=\imwidth]{\impath{000029_2}} &	        
	\includegraphics[width=\imwidth]{\impath{000029_1}} &	 
	\includegraphics[width=\imwidth]{\impath{000030}} \\
	
	\includegraphics[width=\imwidth]{\impath{000218}} &	        
	\includegraphics[width=\imwidth]{\impath{000365}} &	        
	\includegraphics[width=\imwidth]{\impath{000079}} &	        
	\includegraphics[width=\imwidth]{\impath{000473}} &	        
	\includegraphics[width=\imwidth]{\impath{000781}} &	        
	\includegraphics[width=\imwidth]{\impath{000877}} &	 
	\includegraphics[width=\imwidth]{\impath{000062}} \\
	\end{tabular}
	\end{small}
	\caption{\label{fig:morecinsamples} Samples from our class-conditional ImageNet model trained on $256\times256$ images.}
\end{figure*}
}

\newcommand{\rejectioncin}{
\begin{figure*}[thbp]
	\centering
    \renewcommand{\imwidth}{0.3\textwidth}
    \renewcommand{\impath}[1]{img/qualitative_cin/reject/##1} 
    \setlength{\tabcolsep}{1pt}
    \renewcommand{\arraystretch}{1}
	\begin{small}
 	\begin{tabular}{ccc} 	  	
 	\toprule
 	\multicolumn{3}{c}{933: \emph{cheeseburger}} \\
 	acc. rate 1.0 & acc. rate 0.5 & acc. rate 0.1 \\
	\includegraphics[width=\imwidth]{\impath{tk_973_shrink_0_05_933}} &	        
	\includegraphics[width=\imwidth]{\impath{tk_973_shrink_0_10_933}} &	        
	\includegraphics[width=\imwidth]{\impath{tk_973_shrink_0_50_933}} \\
	\midrule
 	\multicolumn{3}{c}{992: \emph{agaric}} \\
 	acc. rate 1.0 & acc. rate 0.5 & acc. rate 0.1 \\
	\includegraphics[width=\imwidth]{\impath{tk_973_shrink_0_05_992}} &
	\includegraphics[width=\imwidth]{\impath{tk_973_shrink_0_10_992}} &
	\includegraphics[width=\imwidth]{\impath{tk_973_shrink_0_50_992}} \\
	\midrule
 	\multicolumn{3}{c}{200: \emph{tibetian terrier}} \\
 	acc. rate 1.0 & acc. rate 0.5 & acc. rate 0.1 \\
	\includegraphics[width=\imwidth]{\impath{tk_973_shrink_0_05_200}} &
	\includegraphics[width=\imwidth]{\impath{tk_973_shrink_0_10_200}} &
	\includegraphics[width=\imwidth]{\impath{tk_973_shrink_0_50_200}} \\
  	\bottomrule
	\end{tabular}
	\end{small}
	\caption{\label{fig:rejectioncin} Visualizing the effect of increased rejection rate (\ie lower acceptance rate) by using a ResNet-101 classifier trained on ImageNet and samples from our class-conditional ImageNet model. Higher rejection rates tend to produce images showing more central, recognizable objects compared to the unguided samples. Here, $k=973$, $p=1.0$ are fixed for all samples. Note that $k=973$ is the \emph{effective} size
of the \emph{VQGAN}'s codebook, \ie it describes how many entries of the codebook with $\dim \mathcal{Z} = 16384$ are
actually used.}
\end{figure*}
}

\newcommand{\topkcin}{
\begin{figure*}[thbp]
	\centering
    \renewcommand{\imwidth}{0.3\textwidth}
    \renewcommand{\impath}[1]{img/qualitative_cin/topk/##1} 
    \setlength{\tabcolsep}{1pt}
    \renewcommand{\arraystretch}{1}
	\begin{small}
 	\begin{tabular}{ccc} 	  	
 	\toprule
 	\multicolumn{3}{c}{933: \emph{cheeseburger}} \\
 	$k=973$ & $k=300$ & $k=100$ \\
	\includegraphics[width=\imwidth]{\impath{top_k_973_933}} &	        
	\includegraphics[width=\imwidth]{\impath{top_k_300_933}} &	        
	\includegraphics[width=\imwidth]{\impath{top_k_100_933}} \\
	\midrule
 	\multicolumn{3}{c}{992: \emph{agaric}} \\
 	$k=973$ & $k=300$ & $k=100$ \\
	\includegraphics[width=\imwidth]{\impath{top_k_973_992}} &
	\includegraphics[width=\imwidth]{\impath{top_k_300_992}} &
	\includegraphics[width=\imwidth]{\impath{top_k_100_992}} \\
	\midrule
 	\multicolumn{3}{c}{200: \emph{tibetian terrier}} \\
 	$k=973$ & $k=300$ & $k=100$ \\
	\includegraphics[width=\imwidth]{\impath{top_k_973_200}} &
	\includegraphics[width=\imwidth]{\impath{top_k_300_200}} &
	\includegraphics[width=\imwidth]{\impath{top_k_100_200}} \\
  	\bottomrule
	\end{tabular}
	\end{small}
	\caption{\label{fig:topkcin} 
Visualizing the effect of varying $k$ in top-k sampling (\ie truncating the probability distribution per image token) by using a ResNet-101 classifier trained on ImageNet and samples from our class-conditional ImageNet model. Lower values of $k$ produce more uniform, low-entropic images compared to samples obtained with full $k$. Here, an acceptance rate of 1.0 and $p=1.0$ are fixed for all samples.	Note that $k=973$ is the \emph{effective} size
of the \emph{VQGAN}'s codebook, \ie it describes how many entries of the codebook with $\dim \mathcal{Z} = 16384$ are
actually used.}
\end{figure*}
}

\newcommand{\toppcin}{
\begin{figure*}[thbp]
	\centering
    \renewcommand{\imwidth}{0.3\textwidth}
    \renewcommand{\impath}[1]{img/qualitative_cin/topp/##1} 
    \setlength{\tabcolsep}{1pt}
    \renewcommand{\arraystretch}{1}
	\begin{small}
 	\begin{tabular}{ccc} 	  	
 	\toprule
 	\multicolumn{3}{c}{933: \emph{cheeseburger}} \\
 	$p=1.0$ & $p=0.96$ & $p=0.84$ \\
	\includegraphics[width=\imwidth]{\impath{1_00_933}} &	        
	\includegraphics[width=\imwidth]{\impath{0_96_933}} &	        
	\includegraphics[width=\imwidth]{\impath{0_84_933}} \\
	\midrule
 	\multicolumn{3}{c}{992: \emph{agaric}} \\
 	$p=1.0$ & $p=0.96$ & $p=0.84$ \\
	\includegraphics[width=\imwidth]{\impath{1_00_992}} &
	\includegraphics[width=\imwidth]{\impath{0_96_992}} &
	\includegraphics[width=\imwidth]{\impath{0_84_992}} \\
	\midrule
 	\multicolumn{3}{c}{200: \emph{tibetian terrier}} \\
 	$p=1.0$ & $p=0.96$ & $p=0.84$ \\
	\includegraphics[width=\imwidth]{\impath{1_00_200}} &
	\includegraphics[width=\imwidth]{\impath{0_96_200}} &
	\includegraphics[width=\imwidth]{\impath{0_84_200}} \\
  	\bottomrule
	\end{tabular}
	\end{small}
	\caption{\label{fig:toppcin} Visualizing the effect of varying $p$ in top-p sampling (or \emph{nucleus sampling} \cite{DBLP:conf/iclr/HoltzmanBDFC20}) by using a ResNet-101 classifier trained on ImageNet and samples from our class-conditional ImageNet model. Lowering $p$ has similar effects as decreasing $k$, see Fig.~\ref{fig:topkcin}. Here, an acceptance rate of 1.0 and $k=973$ are fixed for all samples.}
\end{figure*}
}

\newcommand{\fidandinceptionplots}{
\begin{figure*}[btph]
\centering
\includegraphics[width=0.33\textwidth]{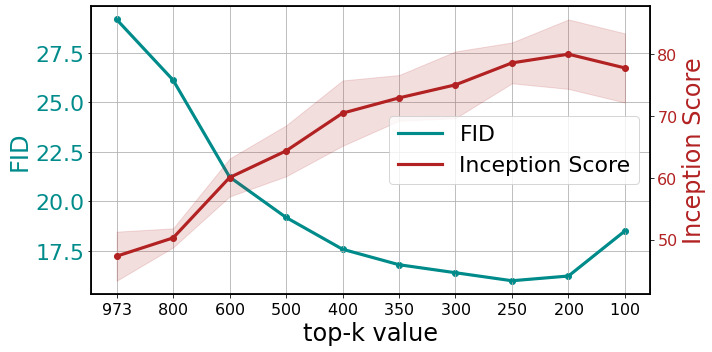}
\includegraphics[width=0.33\textwidth]{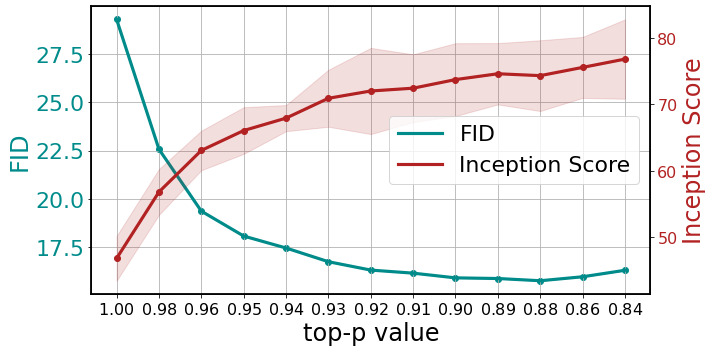}
\includegraphics[width=0.33\textwidth]{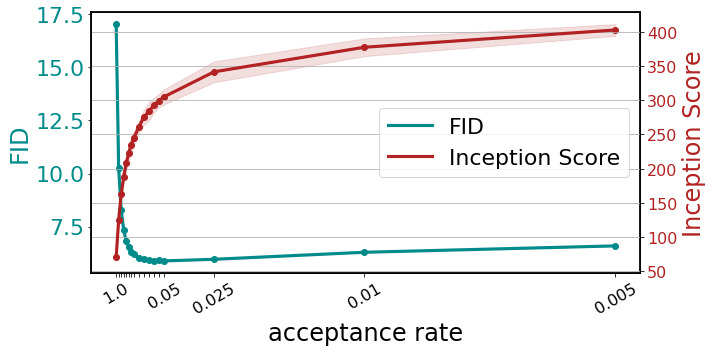}
\vspace{-0.5em}
\caption{\label{fig:fidandinceptionplots} FID and Inception Score as a function of top-k, nucleus and rejection filtering. \vspace{-0.5em}}
\end{figure*}
}

\newcommand{\cinqualitativebirds}{
\begin{figure*}[tbp]
  \renewcommand{\impatha}[1]{img/qualitative_cin/ours/##1}
  \renewcommand{\impathb}[1]{img/qualitative_cin/vqvae2/##1}
  \renewcommand{\impathc}[1]{img/qualitative_cin/biggan/##1}
  \renewcommand{\impathd}[1]{img/qualitative_cin/msp/##1}
  \renewcommand{\imwidth}{0.24\textwidth}
  \setlength{\tabcolsep}{1pt}
  \centering
  \begin{tabular}{cccc}
    \textbf{ours} & VQVAE-2 \cite{razavi2019generating} & BigGAN \cite{brock2018large} & MSP \cite{msp} \\
    \toprule
	\includegraphics[width=\imwidth, align=c]{\impatha{0011}} &	         
	\includegraphics[width=\imwidth, align=c]{\impathb{011}} &	         
	\includegraphics[width=\imwidth, align=c]{\impathc{011}} &
	\includegraphics[width=\imwidth, align=c]{\impathd{011}} 	\vspace*{1em} \\

	\includegraphics[width=\imwidth, align=c]{\impatha{0022}} &	         
	\includegraphics[width=\imwidth, align=c]{\impathb{022}} &	         
	\includegraphics[width=\imwidth, align=c]{\impathc{022}} &
	\includegraphics[width=\imwidth, align=c]{\impathd{022}} \\
    \bottomrule  
  \end{tabular}    
  \caption{Qualitative assessment of various models for class-conditional image synthesis on ImageNet. Depicted classes: \textsl{11: goldfinch} (top) and \textsl{22: bald eagle} (bottom).}
  \label{fig:cinqualbirds}
\end{figure*}
}

\newcommand{\cinqualitativeone}{
\begin{figure*}[tbp]
  \renewcommand{\impatha}[1]{img/qualitative_cin/ours/##1}
  \renewcommand{\impathb}[1]{img/qualitative_cin/vqvae2/##1}
  \renewcommand{\impathc}[1]{img/qualitative_cin/biggan/##1}
  \renewcommand{\impathd}[1]{img/qualitative_cin/msp/##1}
  \renewcommand{\imwidth}{0.24\textwidth}
  \setlength{\tabcolsep}{1pt}
  \centering
  \begin{tabular}{cccc}
    \textbf{ours} & VQVAE-2 \cite{razavi2019generating} & BigGAN \cite{brock2018large} & MSP \cite{msp} \\
    \toprule
	\includegraphics[width=\imwidth, align=c]{\impatha{0000}} &	         
	\includegraphics[width=\imwidth, align=c]{\impathb{000}} &	         
	\includegraphics[width=\imwidth, align=c]{\impathc{000}} &
	\includegraphics[width=\imwidth, align=c]{\impathd{000}} 	\vspace*{1em} \\

	\includegraphics[width=\imwidth, align=c]{\impatha{0009}} &	         
	\includegraphics[width=\imwidth, align=c]{\impathb{009}} &	         
	\includegraphics[width=\imwidth, align=c]{\impathc{009}} &
	\includegraphics[width=\imwidth, align=c]{\impathd{009}} \\
    \bottomrule  
  \end{tabular}    
  \caption{Qualitative assessment of various models for class-conditional image synthesis on ImageNet. Depicted classes: \textsl{0: tench} (top) and \textsl{9: ostrich} (bottom).}
  \label{fig:cinqualone}
\end{figure*}
}

\newcommand{\cinqualitativetwo}{
\begin{figure*}[tbp]
  \renewcommand{\impatha}[1]{img/qualitative_cin/ours/##1}
  \renewcommand{\impathb}[1]{img/qualitative_cin/vqvae2/##1}
  \renewcommand{\impathc}[1]{img/qualitative_cin/biggan/##1}
  \renewcommand{\impathd}[1]{img/qualitative_cin/msp/##1}
  \renewcommand{\imwidth}{0.24\textwidth}
  \setlength{\tabcolsep}{1pt}
  \centering
  \begin{tabular}{cccc}
    \textbf{ours} & VQVAE-2 \cite{razavi2019generating} & BigGAN \cite{brock2018large} & MSP \cite{msp} \\
    \toprule
	\includegraphics[width=\imwidth, align=c]{\impatha{0028}} &	         
	\includegraphics[width=\imwidth, align=c]{\impathb{028}} &	         
	\includegraphics[width=\imwidth, align=c]{\impathc{028}} &
	\includegraphics[width=\imwidth, align=c]{\impathd{028}} 	\vspace*{1em} \\

	\includegraphics[width=\imwidth, align=c]{\impatha{0097}} &	         
	\includegraphics[width=\imwidth, align=c]{\impathb{097}} &	         
	\includegraphics[width=\imwidth, align=c]{\impathc{097}} &
	\includegraphics[width=\imwidth, align=c]{\impathd{097}} \\
    \bottomrule  
  \end{tabular}    
  \caption{Qualitative assessment of various models for class-conditional image synthesis on ImageNet. Depicted classes: \textsl{28: spotted salamander} (top) and \textsl{97: drake} (bottom). We report class labels as in VQVAE-2 \cite{razavi2019generating}.}
  \label{fig:cinqualtwo}
\end{figure*}
}

\newcommand{\cinqualitativethree}{
\begin{figure*}[tbp]
  \renewcommand{\impatha}[1]{img/qualitative_cin/ours/##1}
  \renewcommand{\impathb}[1]{img/qualitative_cin/vqvae2/##1}
  \renewcommand{\impathc}[1]{img/qualitative_cin/biggan/##1}
  \renewcommand{\impathd}[1]{img/qualitative_cin/msp/##1}
  \renewcommand{\imwidth}{0.24\textwidth}
  \setlength{\tabcolsep}{1pt}
  \centering
  \begin{tabular}{cccc}
    \textbf{ours} & VQVAE-2 \cite{razavi2019generating} & BigGAN \cite{brock2018large} & MSP \cite{msp} \\
    \toprule
	\includegraphics[width=\imwidth, align=c]{\impatha{0108}} &	         
	\includegraphics[width=\imwidth, align=c]{\impathb{108}} &	         
	\includegraphics[width=\imwidth, align=c]{\impathc{108}} &
	\includegraphics[width=\imwidth, align=c]{\impathd{108}} 	\vspace*{1em} \\

	\includegraphics[width=\imwidth, align=c]{\impatha{0141}} &	         
	\includegraphics[width=\imwidth, align=c]{\impathb{141}} &	         
	\includegraphics[width=\imwidth, align=c]{\impathc{141}} &
	\includegraphics[width=\imwidth, align=c]{\impathd{141}} \\
    \bottomrule  
  \end{tabular}    
  \caption{Qualitative assessment of various models for class-conditional image synthesis on ImageNet. Depicted classes: \textsl{108: sea anemone} (top) and \textsl{141: redshank} (bottom). We report class labels as in VQVAE-2 \cite{razavi2019generating}.}
  \label{fig:cinqualthree}
\end{figure*}
}

\newcommand{\convlandscapessixteen}{
\begin{figure*}[tbp]
  \renewcommand{\impath}[1]{img/convolutional/semantic/landscapes/##1}
  \renewcommand{\imwidtha}{0.23\textwidth}
  \renewcommand{\imwidthb}{0.38\textwidth}
  \setlength{\tabcolsep}{2pt}
  \centering
  \begin{tabular}{ c cc}
	conditioning & \multicolumn{2}{c}{samples}\\
    \toprule
	\includegraphics[width=\imwidtha, align=c]{\impath{f16/mountainlake_cond}} \phantom {a}&     
	
	\includegraphics[width=\imwidthb, align=c]{\impath{f16/mountainlake3}} &	         
	\includegraphics[width=\imwidthb, align=c]{\impath{f16/mountainlake2}} \\

	\includegraphics[width=\imwidtha, align=c]{\impath{f16/lake}} \phantom {a}&     
	
	\includegraphics[width=\imwidthb, align=c]{\impath{f16/lake_s1}} &	         
	\includegraphics[width=\imwidthb, align=c]{\impath{f16/lake_s2}} \\

	\includegraphics[width=\imwidtha, align=c]{\impath{f16/tree}} \phantom {a}&     
	
	\includegraphics[width=\imwidthb, align=c]{\impath{f16/tree_s3}} &	         
	\includegraphics[width=\imwidthb, align=c]{\impath{f16/tree_s1}} \\

	\includegraphics[width=\imwidtha, align=c]{\impath{f16/mlake2}} \phantom {a}&     
	
	\includegraphics[width=\imwidthb, align=c]{\impath{f16/mlake2_s2}} &	         
	\includegraphics[width=\imwidthb, align=c]{\impath{f16/mlake2_s5}} \\
    \bottomrule  
  \end{tabular}    
  \caption{Samples generated from semantic layouts on S-FLCKR with $f=16$, using the sliding attention window.}
  \label{fig:convf16landscapes}
\end{figure*}
}

\newcommand{\convlandscapesthirtytwo}{
\begin{figure*}[tbp]
  \renewcommand{\impath}[1]{img/convolutional/semantic/landscapes/##1}
  \renewcommand{\imwidtha}{0.23\textwidth}
  \renewcommand{\imwidthb}{0.38\textwidth}
  \setlength{\tabcolsep}{2pt}
  \centering
  \begin{tabular}{ c cc}
	conditioning & \multicolumn{2}{c}{samples}\\
    \toprule
	\includegraphics[width=\imwidtha, align=c]{\impath{f32/river}} \phantom {a}&     
	
	\includegraphics[width=\imwidthb, align=c]{\impath{f32/river_s2}} &	         
	\includegraphics[width=\imwidthb, align=c]{\impath{f32/river_s4}} \\

	\includegraphics[width=\imwidtha, align=c]{\impath{f32/rfw4233cond}} \phantom {a}&     
	
	\includegraphics[width=\imwidthb, align=c]{\impath{f32/rfw512_4233_s1}} &	         
	\includegraphics[width=\imwidthb, align=c]{\impath{f32/rfw512_4233_s3}} \\

	\includegraphics[width=\imwidtha, align=c]{\impath{f32/wind}} \phantom {a}&     
	
	\includegraphics[width=\imwidthb, align=c]{\impath{f32/wind_s1}} &	         
	\includegraphics[width=\imwidthb, align=c]{\impath{f32/wind_s5}} \\
	
	\includegraphics[width=\imwidtha, align=c]{\impath{f32/creek}} \phantom {a}&     
	
	\includegraphics[width=\imwidthb, align=c]{\impath{f32/creek_s1}} &	         
	\includegraphics[width=\imwidthb, align=c]{\impath{f32/creek_s3}} \\
	
    \bottomrule  
  \end{tabular}    
  \caption{Samples generated from semantic layouts on S-FLCKR with $f=32$, using the sliding attention window.}
  \label{fig:convf32landscapes}
\end{figure*}
}

\newcommand{\convcondplot}{
\begin{figure}[tbp]
  \renewcommand{\impatha}[1]{img/convolutional/depth/rin_old/##1}
  \renewcommand{\impathb}[1]{img/convolutional/superres/##1}
  \renewcommand{\impathc}[1]{img/convolutional/semantic/landscapes/##1}
  \renewcommand{\impathd}[1]{img/convolutional/class/##1}
  \renewcommand{\impathe}[1]{img/edges_f8/##1}
  \renewcommand{\imwidth}{0.16\textwidth}
  \renewcommand{\imwidtha}{0.115\textwidth}
  \renewcommand{\imwidthb}{0.19\textwidth}
  \setlength{\tabcolsep}{0pt}
  \centering
  \begin{tabular}{ c cc}
	\includegraphics[width=\imwidth, align=c]{\impatha{depth654cond}} \phantom {a}&     
	
	\includegraphics[width=\imwidth, align=c]{\impatha{depth654s1}} &	         
	\includegraphics[width=\imwidth, align=c]{\impatha{depth654s2}} \\

	\includegraphics[width=\imwidth, align=c]{\impathb{bottles_lr}} \phantom {a}&     
	
	\includegraphics[width=\imwidth, align=c]{\impathb{bottles_sr1}} &	         
	\includegraphics[width=\imwidth, align=c]{\impathb{bottles_sr2}} \\
	
	\includegraphics[width=\imwidth, align=c]{\impathc{f16/mountainlake_cond}} \phantom {a}&     	
	\includegraphics[width=\imwidth, align=c]{\impathc{f16/mountainlake3}} &	         
	\includegraphics[width=\imwidth, align=c]{\impathc{f16/mountainlake2}} \\
	
	\includegraphics[width=\imwidth, align=c]{\impathc{f32/rfw4233cond}} \phantom {a}&     	
	\includegraphics[width=\imwidth, align=c]{\impathc{f32/rfw512_4233_s1}} &	         
	\includegraphics[width=\imwidth, align=c]{\impathc{f32/rfw512_4233_s3}} \\

	\includegraphics[width=\imwidth, align=c]{\impathe{c3_745673}} \phantom {a}&
	\includegraphics[width=\imwidth, align=c]{\impathe{c3_s0}} &	         
	\includegraphics[width=\imwidth, align=c]{\impathe{c3_s2}} \\
  \end{tabular}    
  \vspace{0.015em}
  \caption{Applying the sliding attention window approach
  (Fig.~\ref{fig:attentionslide}) to various conditional image synthesis tasks.
  Top: Depth-to-image on RIN, 2nd row: Stochastic superresolution on IN, 3rd
  and 4th row: Semantic synthesis on S-FLCKR, bottom: Edge-guided synthesis on
  IN. The resulting images vary between $368 \times 496$ and $1024 \times 576$, hence they are best viewed zoomed in.}
  \label{fig:convcondplot}
\end{figure}
}

\newcommand{\convdepth}{
\begin{figure*}[tbp]
  \renewcommand{\impatha}[1]{img/convolutional/depth/rin_old/##1}
  \renewcommand{\impath}[1]{img/convolutional/depth/rin_f16/##1}
  \renewcommand{\imwidth}{0.24\textwidth}
  \setlength{\tabcolsep}{2pt}
  \centering
  \begin{tabular}{ c ccc}
	conditioning & \multicolumn{3}{c}{samples}\\
    \toprule
	\includegraphics[width=\imwidth, align=c]{\impatha{depth654cond}} \phantom {a}&     
	
	\includegraphics[width=\imwidth, align=c]{\impatha{depth654s1}} &	         
	\includegraphics[width=\imwidth, align=c]{\impatha{depth654s2}} &
	\includegraphics[width=\imwidth, align=c]{\impatha{depth654s4}} \\

	\includegraphics[width=\imwidth, align=c]{\impath{dog1_cond}} \phantom {a}&     
	
	\includegraphics[width=\imwidth, align=c]{\impath{dog1_s1}} &	         
	\includegraphics[width=\imwidth, align=c]{\impath{dog1_s3}} &
	\includegraphics[width=\imwidth, align=c]{\impath{dog1_s4}} \\

	\includegraphics[width=\imwidth, align=c]{\impath{crab_cond}} \phantom {a}&     
	
	\includegraphics[width=\imwidth, align=c]{\impath{crab_s1}} &	         
	\includegraphics[width=\imwidth, align=c]{\impath{crab_s4}} &
	\includegraphics[width=\imwidth, align=c]{\impath{crab_s3}} \\

	\includegraphics[width=\imwidth, align=c]{\impath{grasshopper_cond}} \phantom {a}&     
	
	\includegraphics[width=\imwidth, align=c]{\impath{grasshopper_s1}} &	         
	\includegraphics[width=\imwidth, align=c]{\impath{grasshopper_s2}} &
	\includegraphics[width=\imwidth, align=c]{\impath{grasshopper_s3}} \\
	
	\includegraphics[width=\imwidth, align=c]{\impath{bird_cond}} \phantom {a}&     
	
	\includegraphics[width=\imwidth, align=c]{\impath{bird_s1}} &	         
	\includegraphics[width=\imwidth, align=c]{\impath{bird_s2}} &
	\includegraphics[width=\imwidth, align=c]{\impath{bird_s3}} \\
	
    \bottomrule  
  \end{tabular}    
  \caption{Depth-guided neural rendering on RIN with $f=16$ using the sliding
  attention window.}
  \label{fig:convdepth}
\end{figure*}
}

\newcommand{\convdepthtwo}{
\begin{figure*}[tbp]
  \renewcommand{\impatha}[1]{img/convolutional/depth/rin_old/##1}
  \renewcommand{\impath}[1]{img/convolutional/depth/rin_f16/##1}
  \renewcommand{\imwidth}{0.24\textwidth}
  \setlength{\tabcolsep}{2pt}
  \centering
  \begin{tabular}{ c ccc}
	conditioning & \multicolumn{3}{c}{samples}\\
    \toprule
	
	\includegraphics[width=\imwidth, align=c]{\impath{monkey_cond}} \phantom {a}&     
	
	\includegraphics[width=\imwidth, align=c]{\impath{monkey_s1}} &	         
	\includegraphics[width=\imwidth, align=c]{\impath{monkey_s4}} &
	\includegraphics[width=\imwidth, align=c]{\impath{monkey_s3}} \\
	
	\includegraphics[width=\imwidth, align=c]{\impath{linedog_cond}} \phantom {a}&     
	
	\includegraphics[width=\imwidth, align=c]{\impath{linedog_s1}} &	         
	\includegraphics[width=\imwidth, align=c]{\impath{linedog_s2}} &
	\includegraphics[width=\imwidth, align=c]{\impath{linedog_s3}} \\

	\includegraphics[width=\imwidth, align=c]{\impath{goose_cond}} \phantom {a}&     
	
	\includegraphics[width=\imwidth, align=c]{\impath{goose_s1}} &	         
	\includegraphics[width=\imwidth, align=c]{\impath{goose_s2}} &
	\includegraphics[width=\imwidth, align=c]{\impath{goose_s3}} \\

	\includegraphics[width=\imwidth, align=c]{\impath{rundog_cond}} \phantom {a}&     
	
	\includegraphics[width=\imwidth, align=c]{\impath{rundog_s4}} &	         
	\includegraphics[width=\imwidth, align=c]{\impath{rundog_s1}} &
	\includegraphics[width=\imwidth, align=c]{\impath{rundog_s3}} \\
	
	\includegraphics[width=\imwidth, align=c]{\impath{facedog_cond}} \phantom {a}&     
	
	\includegraphics[width=\imwidth, align=c]{\impath{facedog_s4}} &	         
	\includegraphics[width=\imwidth, align=c]{\impath{facedog_s2}} &
	\includegraphics[width=\imwidth, align=c]{\impath{facedog_s3}} \\
    \bottomrule  
  \end{tabular}    
  \caption{Depth-guided neural rendering on RIN with $f=16$ using the sliding
  attention window.}
  \label{fig:convdepthtwo}
\end{figure*}
}

\newcommand{\artsyedges}{
\begin{figure*}[tbp]
  \renewcommand{\impath}[1]{img/convolutional/edges/in_f16/##1}
  \renewcommand{\imwidth}{0.32\textwidth}
  \setlength{\tabcolsep}{2pt}
  \centering
  \begin{tabular}{ c cc}
	conditioning & \multicolumn{2}{c}{samples}\\
    \toprule

	\includegraphics[width=\imwidth, align=c]{\impath{helmet_cond}} \phantom {a}&     
	
	\includegraphics[width=\imwidth, align=c]{\impath{helmet_s1}} &	         
	\includegraphics[width=\imwidth, align=c]{\impath{helmet_s3}} \\
	
	\includegraphics[width=\imwidth, align=c]{\impath{tbox_cond}} \phantom {a}&     
	
	\includegraphics[width=\imwidth, align=c]{\impath{tbox_s1}} &	         
	\includegraphics[width=\imwidth, align=c]{\impath{tbox_s2}} \\
	
	\includegraphics[width=\imwidth, align=c]{\impath{shot_cond}} \phantom {a}&     
	
	\includegraphics[width=\imwidth, align=c]{\impath{shot_s2}} &	         
	\includegraphics[width=\imwidth, align=c]{\impath{shot_s4}} \\
    \bottomrule  
  \end{tabular}    
  \caption{Intentionally limiting the receptive field can lead to interesting creative applications like this one: Edge-to-Image synthesis on IN with $f=8$, using the sliding attention window.}
  \label{fig:convedges}
\end{figure*}
}

\newcommand{\convsr}{
\begin{figure*}[tbp]
  \renewcommand{\impath}[1]{img/convolutional/superres/##1}
  \renewcommand{\imwidth}{0.24\textwidth}
  \setlength{\tabcolsep}{2pt}
  \centering
  \begin{tabular}{ c ccc}
	conditioning & \multicolumn{3}{c}{samples}\\
    \toprule
	\includegraphics[width=\imwidth, align=c]{\impath{goose_lr}} \phantom {a}&     
	
	\includegraphics[width=\imwidth, align=c]{\impath{goose_hr1}} &	         
	\includegraphics[width=\imwidth, align=c]{\impath{goose_hr2}} &
	\includegraphics[width=\imwidth, align=c]{\impath{goose_hr3}} \\
	
	\includegraphics[width=\imwidth, align=c]{\impath{dog_lr}} \phantom {a}&     
	
	\includegraphics[width=\imwidth, align=c]{\impath{dog_hr1}} &	         
	\includegraphics[width=\imwidth, align=c]{\impath{dog_hr2}} &
	\includegraphics[width=\imwidth, align=c]{\impath{dog_hr3}} \\

	\includegraphics[width=\imwidth, align=c]{\impath{lamp_lr}} \phantom {a}&     
	
	\includegraphics[width=\imwidth, align=c]{\impath{lamp_hr1}} &	         
	\includegraphics[width=\imwidth, align=c]{\impath{lamp_hr2}} &
	\includegraphics[width=\imwidth, align=c]{\impath{lamp_hr3}} \\

	\includegraphics[width=\imwidth, align=c]{\impath{pc_lr}} \phantom {a}&     
	
	\includegraphics[width=\imwidth, align=c]{\impath{pc_hr1}} &	         
	\includegraphics[width=\imwidth, align=c]{\impath{pc_hr2}} &
	\includegraphics[width=\imwidth, align=c]{\impath{pc_hr3}} \\

	\includegraphics[width=\imwidth, align=c]{\impath{pot_lr}} \phantom {a}&     
	
	\includegraphics[width=\imwidth, align=c]{\impath{pot_hr1}} &	         
	\includegraphics[width=\imwidth, align=c]{\impath{pot_hr2}} &
	\includegraphics[width=\imwidth, align=c]{\impath{pot_hr3}} \\
    \bottomrule  
  \end{tabular}    
  \caption{Additional results for stochastic superresolution with an $f=16$ model on IN, using the sliding attention window.}
  \label{fig:convsr}
\end{figure*}
}

\newcommand{\fstudyfaces}{
\begin{figure*}[tbp]
  \renewcommand{\impath}[1]{img/fexperiments/faces/##1}
  \renewcommand{\imwidth}{0.095\textwidth}
  \setlength{\tabcolsep}{0pt}
  \centering
  \begin{tabular}{c  c c  c c c  c c c c}
	f1 & \multicolumn{2}{c}{f2} & \multicolumn{3}{c}{f8} & \multicolumn{2}{c}{f16} & \multicolumn{2}{c}{\footnotesize{\emph{downsampling factor}}} \\
    \toprule
   	\includegraphics[width=\imwidth, align=c]{\impath{f1/faces_f1_s0_gs361250}} \phantom {a} &     
   	
	\includegraphics[width=\imwidth, align=c]{\impath{f2/faces_f2_s1}} &     
	\includegraphics[width=\imwidth, align=c]{\impath{f2/faces_f2_s3}} \phantom {a}&         
	
	\includegraphics[width=\imwidth, align=c]{\impath{f8/faces_f8_halfbeard}} &	         
	\includegraphics[width=\imwidth, align=c]{\impath{f8/faces_f8_s2}} &
	\includegraphics[width=\imwidth, align=c]{\impath{f8/faces_f8_s3}} \phantom {a}&     
	
	\includegraphics[width=\imwidth, align=c]{\impath{f16/faces_f16_s7}} &         
	\includegraphics[width=\imwidth, align=c]{\impath{f16/faces_f16_s9}} &         
	\includegraphics[width=\imwidth, align=c]{\impath{f16/faces_f16_s8}} &         
	\includegraphics[width=\imwidth, align=c]{\impath{f16/faces_f16_s10}} \phantom {a}\\

   	\includegraphics[width=\imwidth, align=c]{\impath{f1/faces_f1_s1_gs361250}} \phantom {a} &     
	   	
	\includegraphics[width=\imwidth, align=c]{\impath{f2/faces_f2_s4}} &     
	\includegraphics[width=\imwidth, align=c]{\impath{f2/faces_f2_s2}} \phantom {a}&         

	\includegraphics[width=\imwidth, align=c]{\impath{f8/faces_f8_s4}} &	         
	\includegraphics[width=\imwidth, align=c]{\impath{f8/faces_f8_s5}} &
	\includegraphics[width=\imwidth, align=c]{\impath{f8/faces_f8_s6}} \phantom {a}&     
	
	\includegraphics[width=\imwidth, align=c]{\impath{f16/faces_f16_s14}} &         
	\includegraphics[width=\imwidth, align=c]{\impath{f16/faces_f16_s13}} &         
	\includegraphics[width=\imwidth, align=c]{\impath{f16/faces_f16_s1}} &         
	\includegraphics[width=\imwidth, align=c]{\impath{f16/faces_f16_s19_gs92092}} \phantom {a}\\	
    \bottomrule  
    \footnotesize 1.0 & \multicolumn{2}{c}{\footnotesize 3.86} &
    \multicolumn{3}{c}{\footnotesize 65.81} & \multicolumn{2}{c}{\footnotesize 280.68} & \multicolumn{2}{c}{\footnotesize{\emph{speed-up}}} \\
  \end{tabular}    
  \caption{Evaluating the importance of effective codebook for HQ-Faces
  (CelebA-HQ and FFHQ) for a fixed sequence length $\vert s \vert = 16\cdot16 =
  256$. Globally consistent structures can only be modeled with a context-rich
  vocabulary (right). All samples are generated with temperature $t=1.0$ and
  top-$k$ sampling with $k=100$. Last row reports the speedup over the f1
  baseline which operates directly on pixels and takes 7258 seconds to produce
  a sample on a NVIDIA GeForce GTX Titan X.\vspace{-1em}}
  \label{fig:fstudyfaces}
\end{figure*}
}

\newcommand{\quantfacefstudy}{
\begin{figure*}[p]
\renewcommand{\impath}[2]{img/fexperiments/suppfaces_recs/##1/009287_##2}
\renewcommand{\impatha}[2]{img/fexperiments/suppfaces_samples/##1/002895_##2}
\renewcommand{\impathb}[1]{img/fexperiments/suppfaces_samples/##1/samples}
\renewcommand{\imwidth}{0.142\textwidth}
\setlength{\tabcolsep}{2pt}
\centering
  \begin{minipage}[c]{0.40\textwidth}
    \vspace{1.0em}
\includegraphics[width=0.99\linewidth]{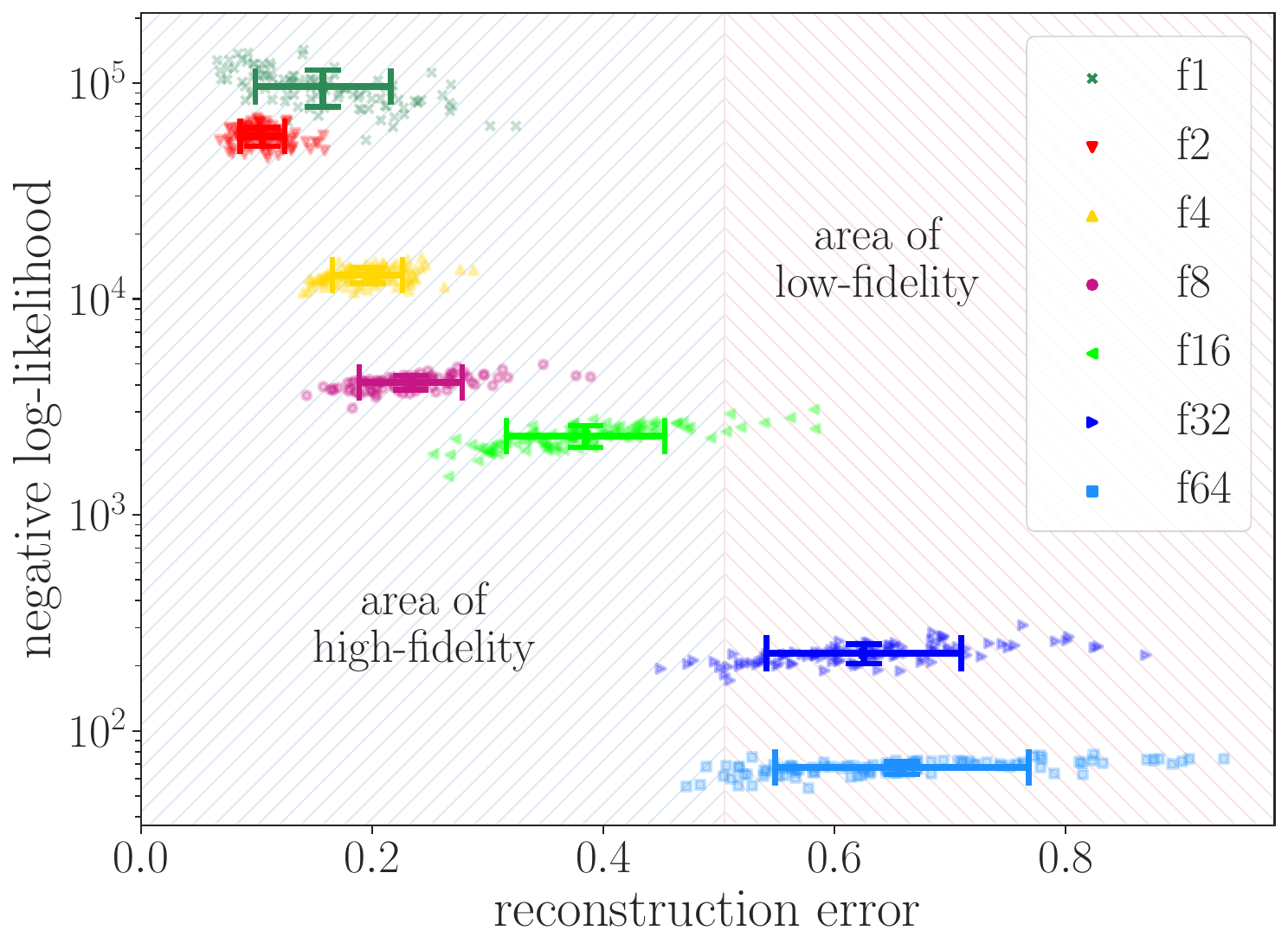}
  \end{minipage}
  \begin{minipage}[c]{0.58\textwidth}
  \begin{tabular}{c c c c c c c}
    & f2 & f4 & f8 & \textbf{f16} & f32 & f64 \\
    \toprule

    \parbox[c]{\imwidth}{\centering \scriptsize recon-struction} &
    \includegraphics[width=\imwidth, align=c]{\impath{2020-11-13T11-28-13_faces_f2_coord}{xrec}} &     
    \includegraphics[width=\imwidth, align=c]{\impath{2020-11-22T13-50-21_faces_f4_coord}{xrec}} &     
    \includegraphics[width=\imwidth, align=c]{\impath{2020-11-12T22-48-25_faces_f8_coord}{xrec}} &
   \includegraphics[width=\imwidth, align=c]{\impath{2020-11-13T21-41-45_faces_f16_coord}{xrec}} &     
   \includegraphics[width=\imwidth, align=c]{\impath{2020-11-22T13-34-58_faces_f32_coord}{xrec}} &     
   \includegraphics[width=\imwidth, align=c]{\impath{2020-11-23T01-08-26_faces_f64_coord}{xrec}} \\    
    \parbox[c]{\imwidth}{\centering \scriptsize sample} &
    \includegraphics[width=\imwidth, align=c]{\impatha{2020-11-13T11-28-13_faces_f2_coord}{xsample}} &     
    \includegraphics[width=\imwidth, align=c]{\impatha{2020-11-22T13-50-21_faces_f4_coord}{xsample}} &     
    \includegraphics[width=\imwidth, align=c]{\impatha{2020-11-12T22-48-25_faces_f8_coord}{xsample}} &
    \includegraphics[width=\imwidth, align=c]{\impatha{2020-11-13T21-41-45_faces_f16_coord}{xsample}} &     
    \includegraphics[width=\imwidth, align=c]{\impatha{2020-11-22T13-34-58_faces_f32_coord}{xsample}} &     
    \includegraphics[width=\imwidth, align=c]{\impatha{2020-11-23T01-08-26_faces_f64_coord}{xsample}} \\    
    \parbox[c]{\imwidth}{\centering \scriptsize samples} &
    \includegraphics[width=\imwidth, align=c]{\impathb{2020-11-13T11-28-13_faces_f2_coord}} &     
    \includegraphics[width=\imwidth, align=c]{\impathb{2020-11-22T13-50-21_faces_f4_coord}} &     
    \includegraphics[width=\imwidth, align=c]{\impathb{2020-11-12T22-48-25_faces_f8_coord}} &
    \includegraphics[width=\imwidth, align=c]{\impathb{2020-11-13T21-41-45_faces_f16_coord}} &     
    \includegraphics[width=\imwidth, align=c]{\impathb{2020-11-22T13-34-58_faces_f32_coord}} &     
    \includegraphics[width=\imwidth, align=c]{\impathb{2020-11-23T01-08-26_faces_f64_coord}} \\    
    \parbox[c]{\imwidth}{\centering \scriptsize rec. error} &
    \tiny $0.11 \pm 0.02$ &
    \tiny $0.20 \pm 0.03$ &
    \tiny $0.23 \pm 0.04$ &
    \tiny $0.38 \pm 0.07$ &
    \tiny $0.63 \pm 0.08$ &
    \tiny $0.66 \pm 0.11$ \\
    \parbox[c]{\imwidth}{\centering \scriptsize nll} &
    \tiny $5.66\cdot 10^4$ &
    \tiny $1.29\cdot 10^4$ &
    \tiny $4.10\cdot 10^3$ &
    \tiny $2.32\cdot 10^3$ &
    \tiny $2.28\cdot 10^2$ &
    \tiny $6.75\cdot 10^1$ \\
    \bottomrule
  \end{tabular}    
  \end{minipage}
\caption{Trade-off between negative log-likelihood (nll) and reconstruction error.
  While context-rich encodings obtained with large factors $f$ allow the
  transformer to effectively model long-range interactions, the reconstructions
  capabilities and hence quality of samples suffer after a critical value
  (here, $f=16$). For more details, see Sec.~\ref{supp:contextrich}.}
\label{fig:quantfacefstudy}
\end{figure*}
}

\newcommand{\vqvaegan}{
\begin{figure*}[p]
    \renewcommand{\imwidth}{\textwidth}
    \includegraphics[width=\imwidth, align=c]{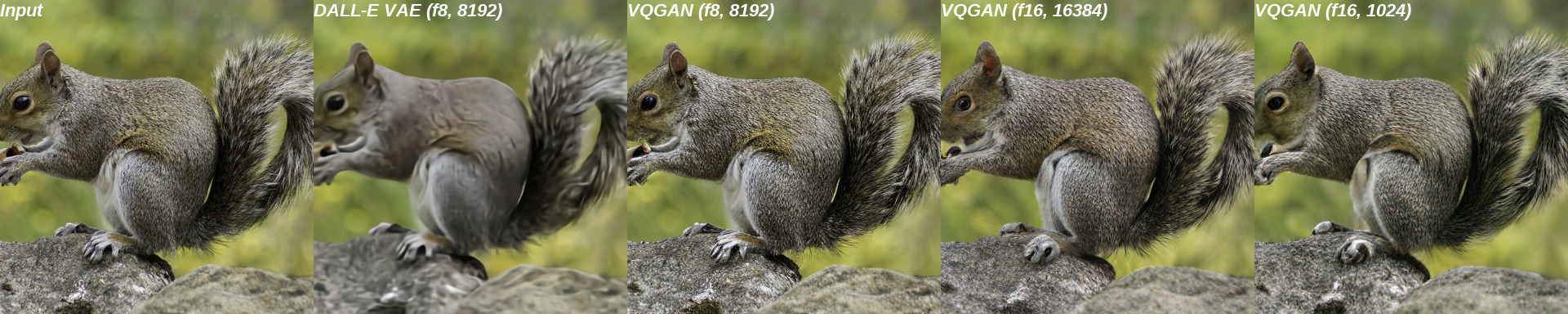}
    \includegraphics[width=\imwidth, align=c]{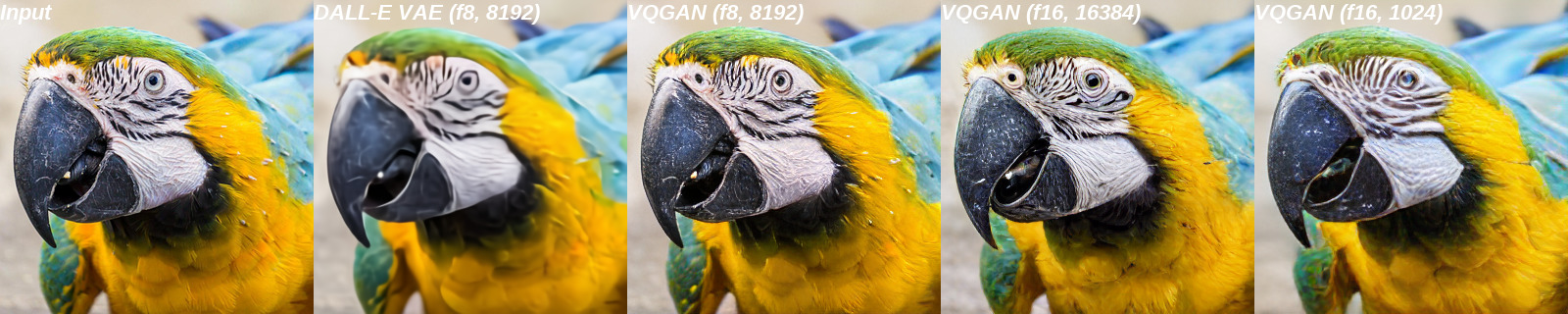}
  \caption{Comparing reconstruction capabilities between VQVAEs and
  \emph{VQGAN}s. Numbers in parentheses denote compression factor and
  codebook size. With the same compression factor and codebook size,
  \emph{VQGAN}s produce more realistic reconstructions compared to blurry
  reconstructions of VQVAEs. This enables increased compression rates
  for \emph{VQGAN} while retaining realistic reconstructions. See Sec.~\ref{supp:contextrich}.}
\label{fig:vqvaegan}
\end{figure*}
}

\newcommand{\rfwfstudy}{
\begin{figure*}[p]
\renewcommand{\impath}[1]{img/fexperiments/supprfw_samples/##1}
\renewcommand{\imwidth}{0.165\textwidth}
\renewcommand{\imwidtha}{0.24\textwidth}
\renewcommand{\impatha}[1]{img/fexperiments/supprfw_recs/2020-11-11T17-24-24_landscapes_f4_ss/##1}
\renewcommand{\impathb}[1]{img/fexperiments/supprfw_recs/2020-11-09T13-31-51_RFWnet2nettransformer_v2/##1}
\renewcommand{\impathc}[1]{img/fexperiments/supprfw_recs/2020-10-16T10-39-45_RFWnet2nettransformer512_restore/##1}
\setlength{\tabcolsep}{0pt}
\centering
  \begin{minipage}[c]{0.575\textwidth}
  \begin{tabular}{c  c c  c c c}
	f4 & \multicolumn{2}{c}{f16} & \multicolumn{3}{c}{f32} \\
    \toprule
    \includegraphics[width=\imwidth, align=c]{\impath{2020-11-11T17-24-24_landscapes_f4_ss/007813_xsample}} \phantom {a} &     

    \includegraphics[width=\imwidth, align=c]{\impath{2020-11-09T13-31-51_RFWnet2nettransformer_v2/007813_xsample}} &     
    \includegraphics[width=\imwidth, align=c]{\impath{2020-11-09T13-31-51_RFWnet2nettransformer_v2/007751_xsample}} \phantom {a}&         

    \includegraphics[width=\imwidth, align=c]{\impath{2020-10-16T10-39-45_RFWnet2nettransformer512_restore/007813_xsample}} &	         
    \includegraphics[width=\imwidth, align=c]{\impath{2020-10-16T10-39-45_RFWnet2nettransformer512_restore/007751_xsample}} &
    \includegraphics[width=\imwidth, align=c]{\impath{2020-10-16T10-39-45_RFWnet2nettransformer512_restore/000235_xsample}} \\

    \includegraphics[width=\imwidth, align=c]{\impath{2020-11-11T17-24-24_landscapes_f4_ss/004225_xsample}} \phantom {a} &     

    \includegraphics[width=\imwidth, align=c]{\impath{2020-11-09T13-31-51_RFWnet2nettransformer_v2/004225_xsample}} &     
    \includegraphics[width=\imwidth, align=c]{\impath{2020-11-09T13-31-51_RFWnet2nettransformer_v2/003462_xsample}} \phantom {a}&         

    \includegraphics[width=\imwidth, align=c]{\impath{2020-10-16T10-39-45_RFWnet2nettransformer512_restore/004225_xsample}} &	         
    \includegraphics[width=\imwidth, align=c]{\impath{2020-10-16T10-39-45_RFWnet2nettransformer512_restore/003462_xsample}} &
    \includegraphics[width=\imwidth, align=c]{\impath{2020-10-16T10-39-45_RFWnet2nettransformer512_restore/002895_xsample}} \\

    \includegraphics[width=\imwidth, align=c]{\impath{2020-11-11T17-24-24_landscapes_f4_ss/005056_xsample}} \phantom {a} &     

    \includegraphics[width=\imwidth, align=c]{\impath{2020-11-09T13-31-51_RFWnet2nettransformer_v2/005056_xsample}} &     
    \includegraphics[width=\imwidth, align=c]{\impath{2020-11-09T13-31-51_RFWnet2nettransformer_v2/000905_xsample}} \phantom {a}&         

    \includegraphics[width=\imwidth, align=c]{\impath{2020-10-16T10-39-45_RFWnet2nettransformer512_restore/005056_xsample}} &	         
    \includegraphics[width=\imwidth, align=c]{\impath{2020-10-16T10-39-45_RFWnet2nettransformer512_restore/000905_xsample}} &
    \includegraphics[width=\imwidth, align=c]{\impath{2020-10-16T10-39-45_RFWnet2nettransformer512_restore/000144_xsample}} \\
    \bottomrule  
  \end{tabular}    
  \end{minipage}
  \begin{minipage}[c]{0.3675\textwidth}
  \hspace{2.5em}
  \begin{tabular}{c c c c}
    factor & \multicolumn{3}{c}{reconstructions} \\
    \toprule
    \parbox[c]{\imwidtha}{\centering f4} &
    \includegraphics[width=\imwidtha, align=c]{\impatha{009287_xrec}} &     
    \includegraphics[width=\imwidtha, align=c]{\impatha{004409_xrec}} &     
    \includegraphics[width=\imwidtha, align=c]{\impatha{004225_xrec}} \\    
    \midrule
    \parbox[c]{\imwidtha}{\centering f16} &
    \includegraphics[width=\imwidtha, align=c]{\impathb{009287_xrec}} &     
    \includegraphics[width=\imwidtha, align=c]{\impathb{004409_xrec}} &     
    \includegraphics[width=\imwidtha, align=c]{\impathb{004225_xrec}} \\    
    \midrule
    \parbox[c]{\imwidtha}{\centering f32} &
    \includegraphics[width=\imwidtha, align=c]{\impathc{009287_xrec}} &     
    \includegraphics[width=\imwidtha, align=c]{\impathc{004409_xrec}} &     
    \includegraphics[width=\imwidtha, align=c]{\impathc{004225_xrec}} \\    
    \bottomrule
  \end{tabular}    
  \end{minipage}
  \caption{Samples on landscape dataset (left) obtained with different factors
  $f$, analogous to Fig.~\ref{fig:fstudyfaces}. In contrast to faces, a factor
  of $f=32$ still allows for faithful reconstructions (right). See also
  Sec.~\ref{supp:contextrich}.}
  \label{fig:rfwfstudy}
\end{figure*}
}

\newcommand{\alaskatop}{
  \vspace{-1em}
	\centering
	\includegraphics[width=1.00\textwidth, trim=0em 0em 0em 0em, clip]{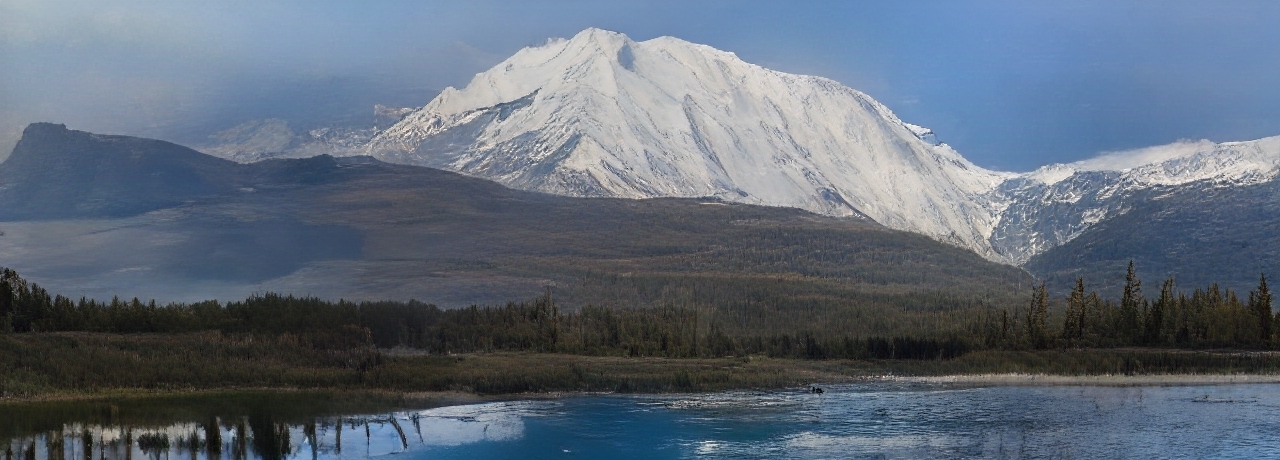}
  \captionof{figure}{
  Our approach enables transformers
  to synthesize high-resolution images like this one, which contains
  1280x460 pixels.}
	\label{fig:frontpage}
  \vspace{1.5em}
}

\newcommand{\modelfigure}{
	\begin{figure*}
		\centering
	\includegraphics[width=0.975\textwidth]{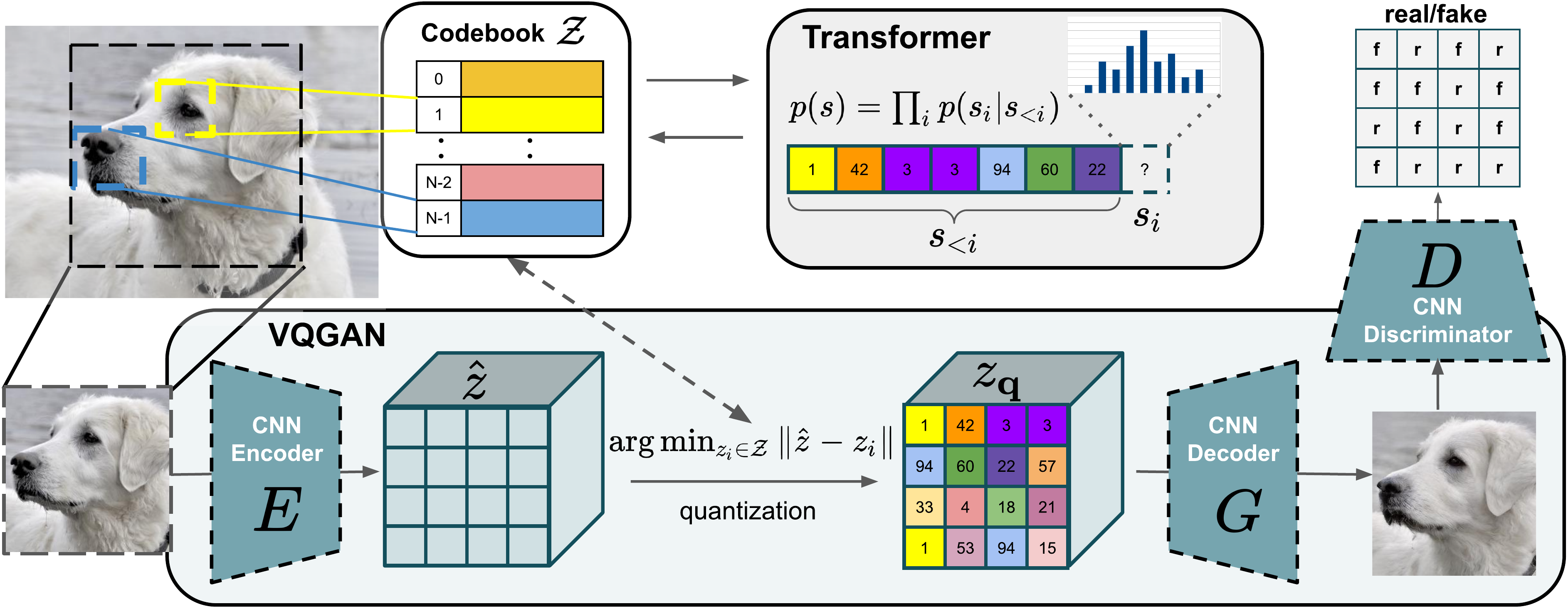}
  \caption{Our approach uses a convolutional \emph{VQGAN} to learn a codebook of context-rich visual parts, whose composition is subsequently modeled with an autoregressive transformer architecture. A discrete codebook provides the interface between these architectures and a patch-based discriminator enables strong compression while retaining high perceptual quality. This method introduces the efficiency of convolutional approaches to transformer based high resolution image synthesis.}
	\label{fig:model}
	\end{figure*}
}

\newcommand{\attentionslide}{
	\begin{figure}[th]
	\centering
	\includegraphics[width=0.485\textwidth]{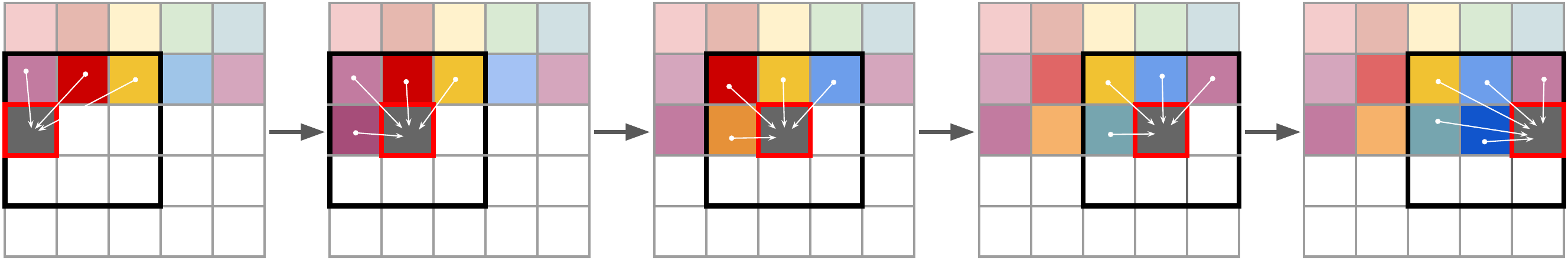}
    \caption{Sliding attention window.}
	\label{fig:attentionslide}
	\end{figure}
}

\newcommand{\bigsamples}{
	\begin{figure}
		\centering
	\includegraphics[width=0.4995\textwidth]{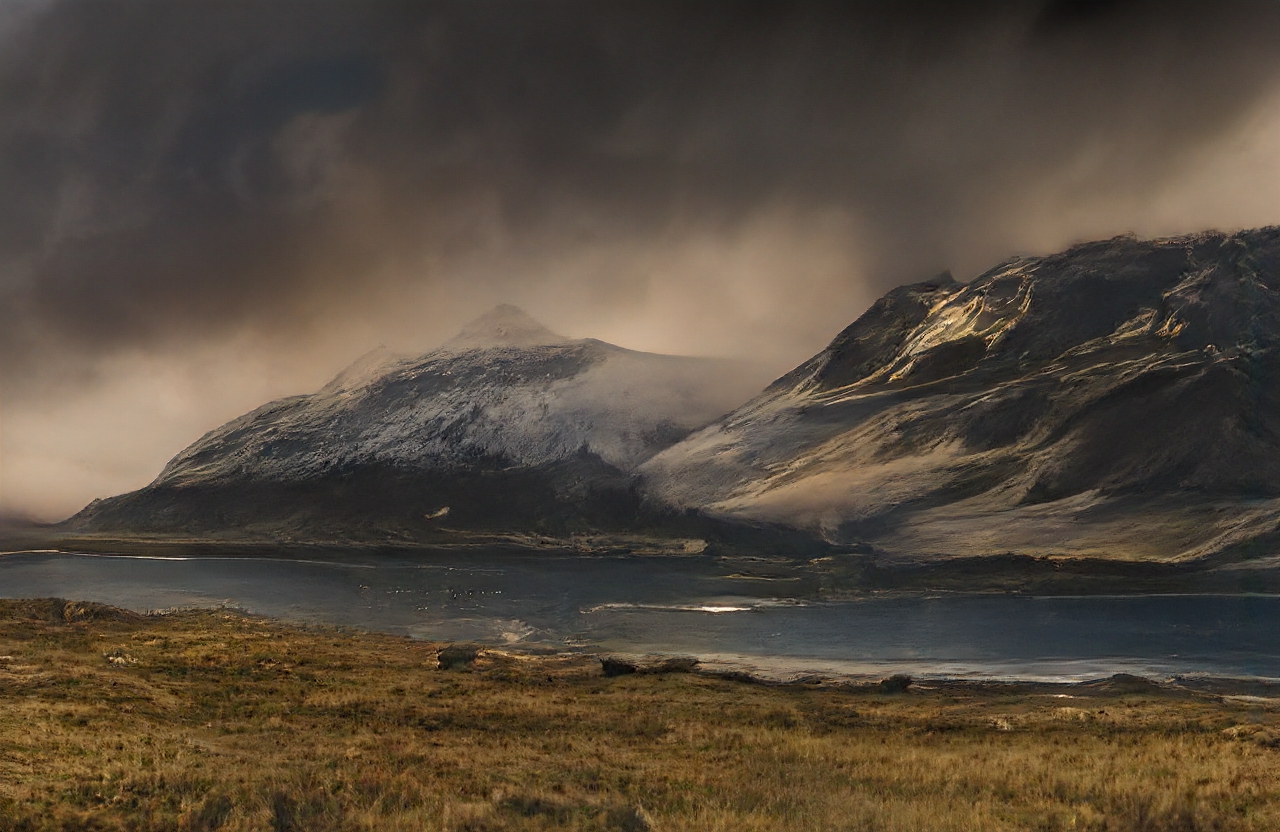}
	\includegraphics[width=0.4995\textwidth]{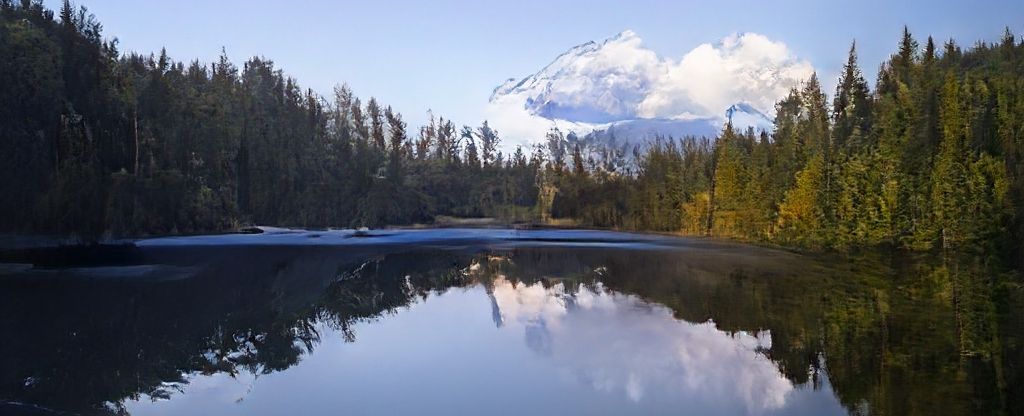}
	\includegraphics[width=0.4995\textwidth]{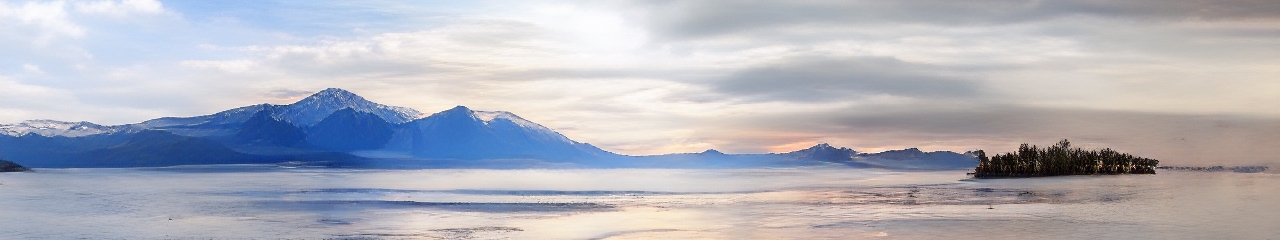}
  \caption{Samples generated from semantic layouts on S-FLCKR. Sizes from
    top-to-bottom: $1280 \times 832$, $1024 \times 416$ and $1280 \times 240$
    pixels. Best viewed zoomed in. A larger visualization can be found in the
    appendix, see Fig~\ref{fig:bigsamplesbig}.}
	\label{fig:bigsamples}
	\end{figure}
}

\newcommand{\bigsamplesbig}{
	\begin{figure}
	\centering
	\includegraphics[width=0.995\textwidth]{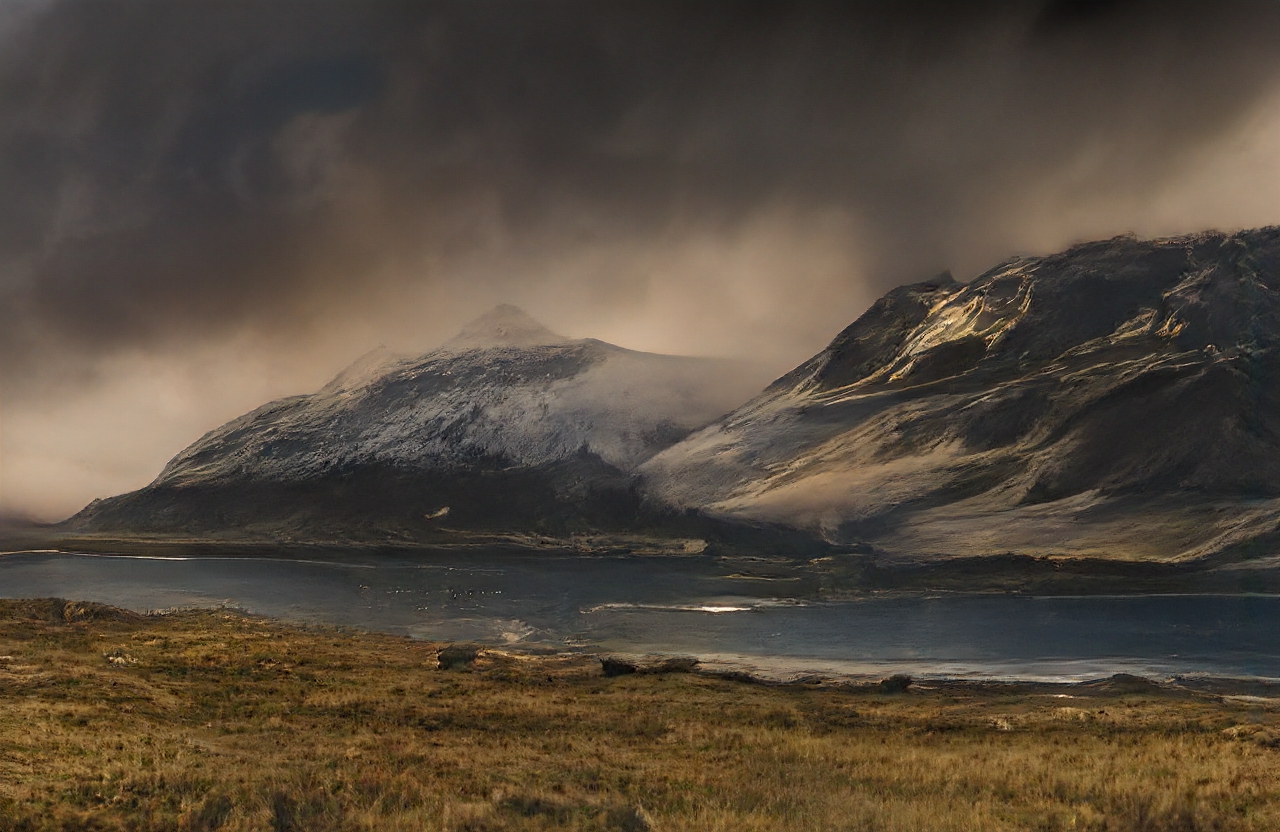}
	\includegraphics[width=0.995\textwidth]{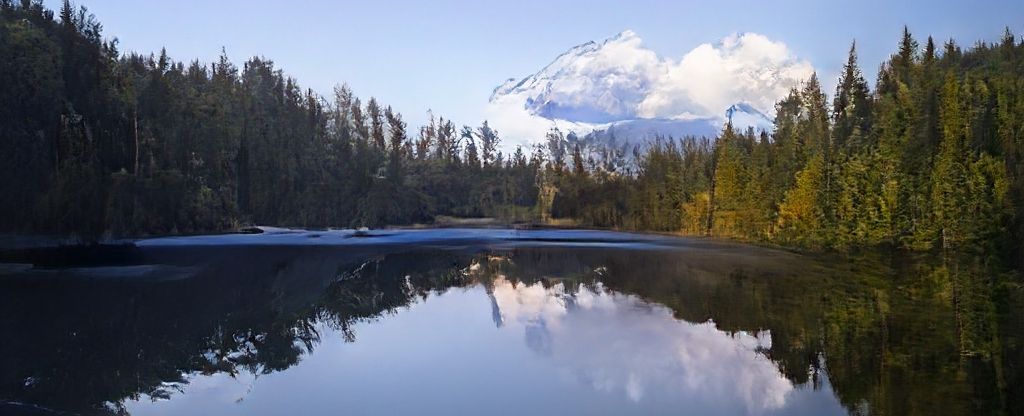}
	\includegraphics[width=0.995\textwidth]{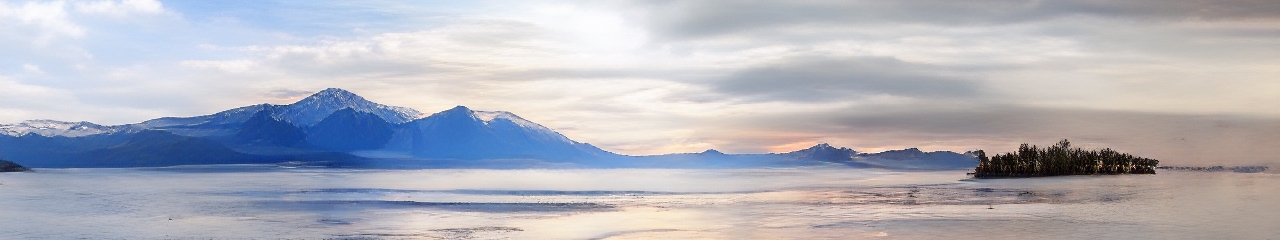}
  \caption{Samples generated from semantic layouts on S-FLCKR. Sizes from
    top-to-bottom: $1280 \times 832$, $1024 \times 416$ and $1280 \times 240$
    pixels.}
	\label{fig:bigsamplesbig}
	\end{figure}
}

\newcommand{\bigsamplesbigtwo}{
	\begin{figure}
	\centering
	\includegraphics[width=0.99\textwidth]{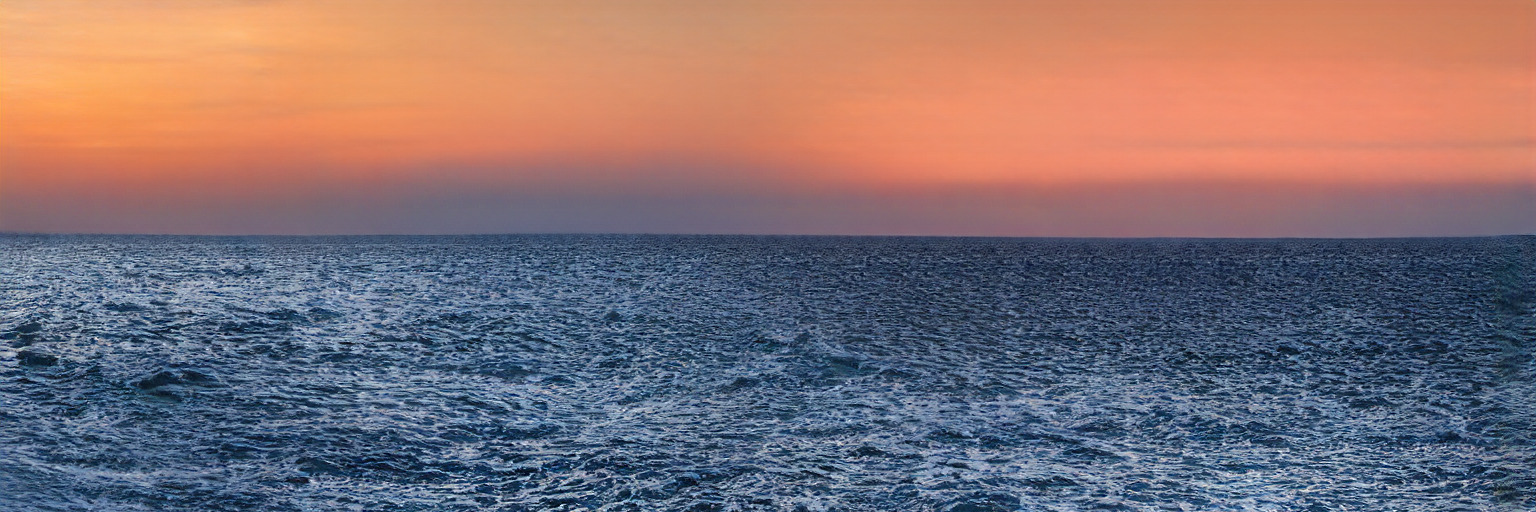}
	\includegraphics[width=0.99\textwidth]{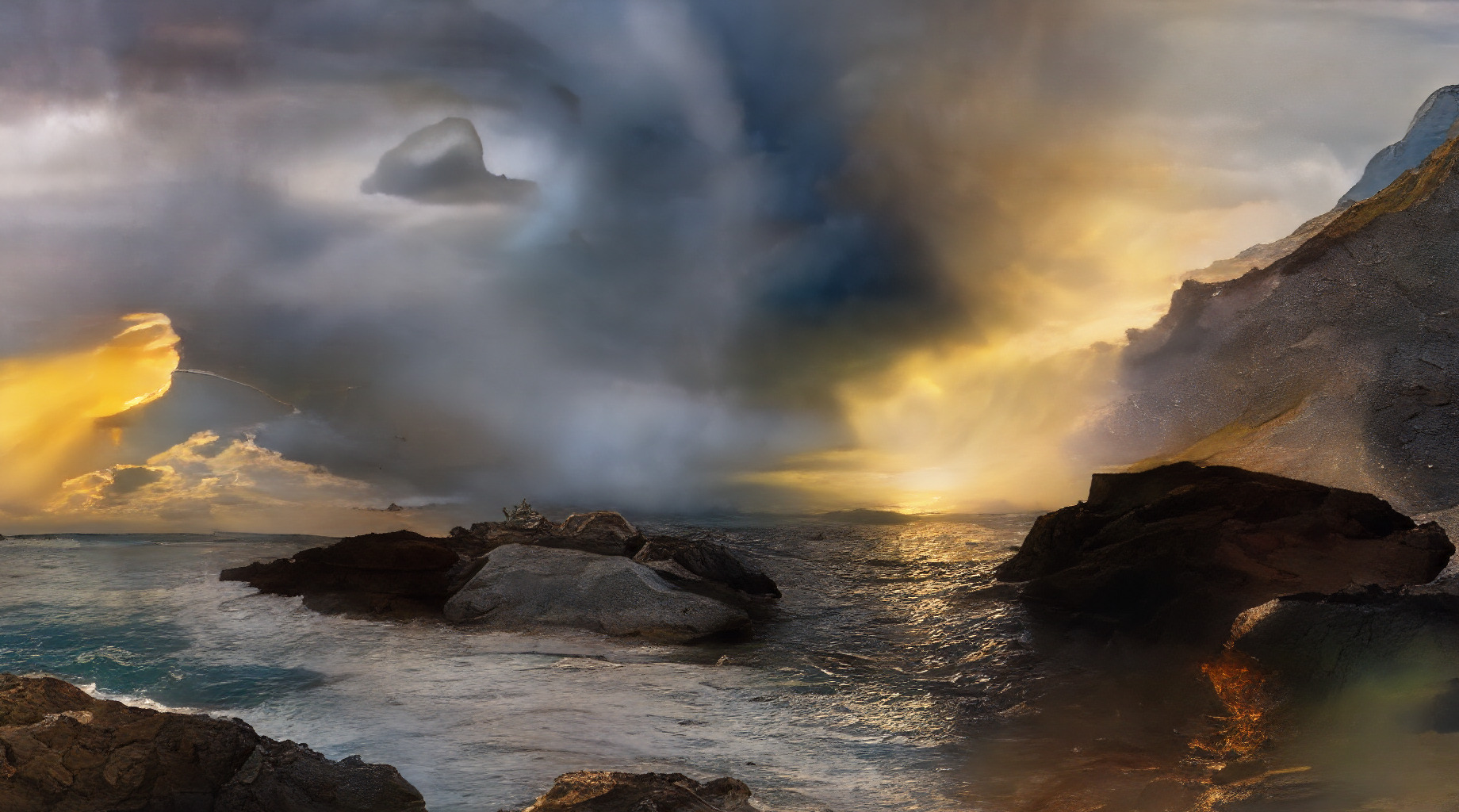}	
	\includegraphics[width=0.99\textwidth]{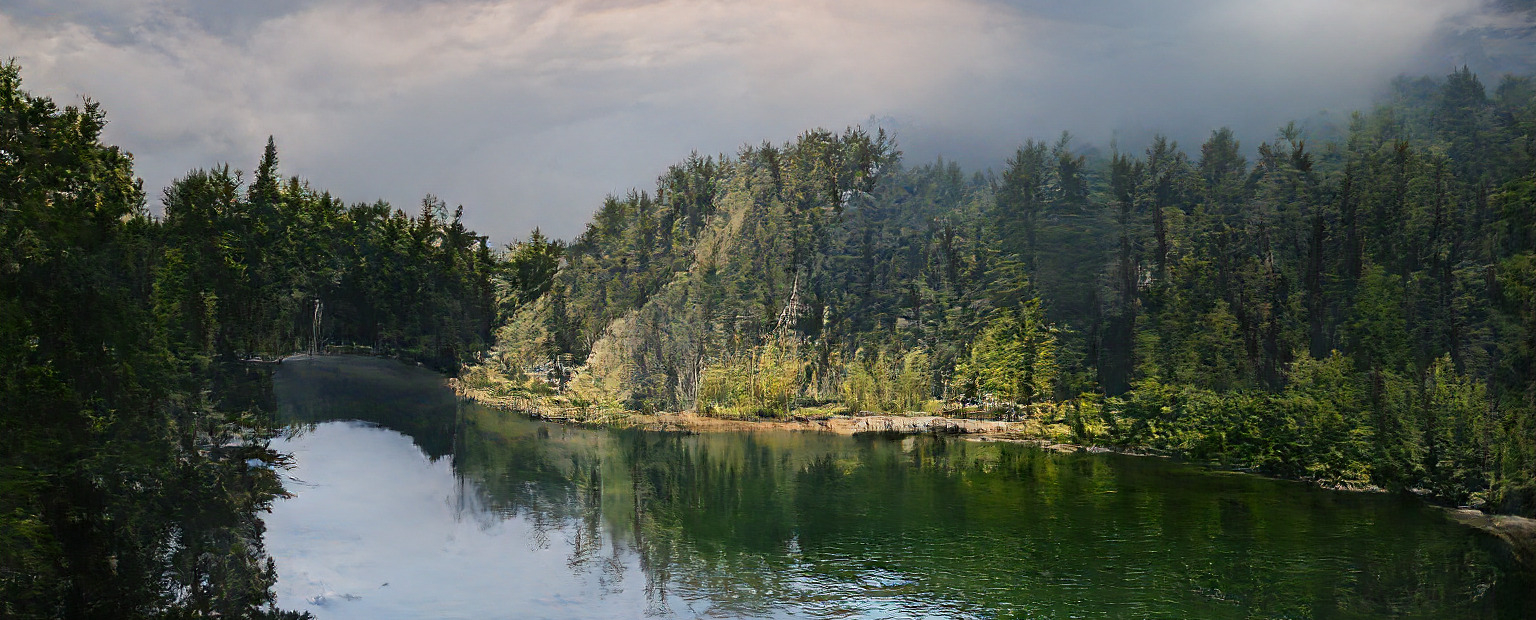}	
  \caption{Samples generated from semantic layouts on S-FLCKR. Sizes from
    top-to-bottom: $1536\times 512$, $1840\times 1024$, and $1536\times 620$
    pixels.}
	\label{fig:bigsamplesbigtwo}
	\end{figure}
}

\newcommand{\bigsamplesbigthree}{
	\begin{figure}
	\centering
	\includegraphics[width=0.99\textwidth]{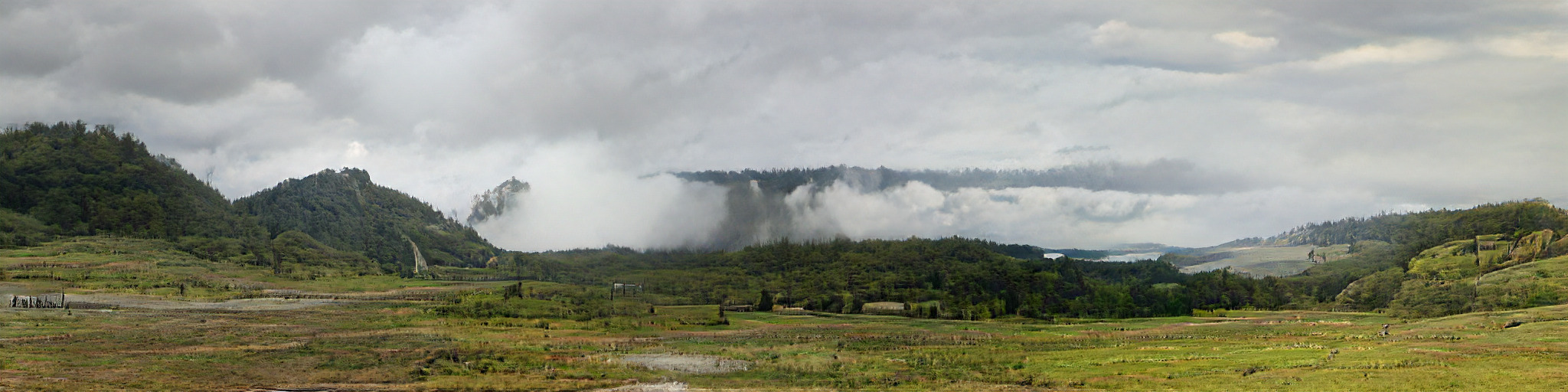}
	\includegraphics[width=0.99\textwidth]{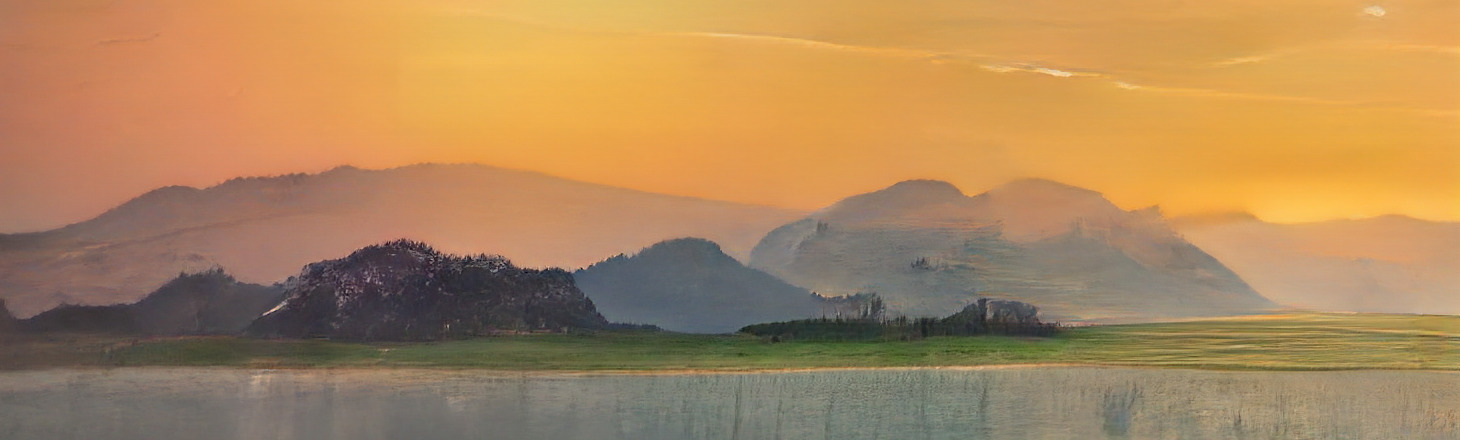}
    \includegraphics[width=0.99\textwidth]{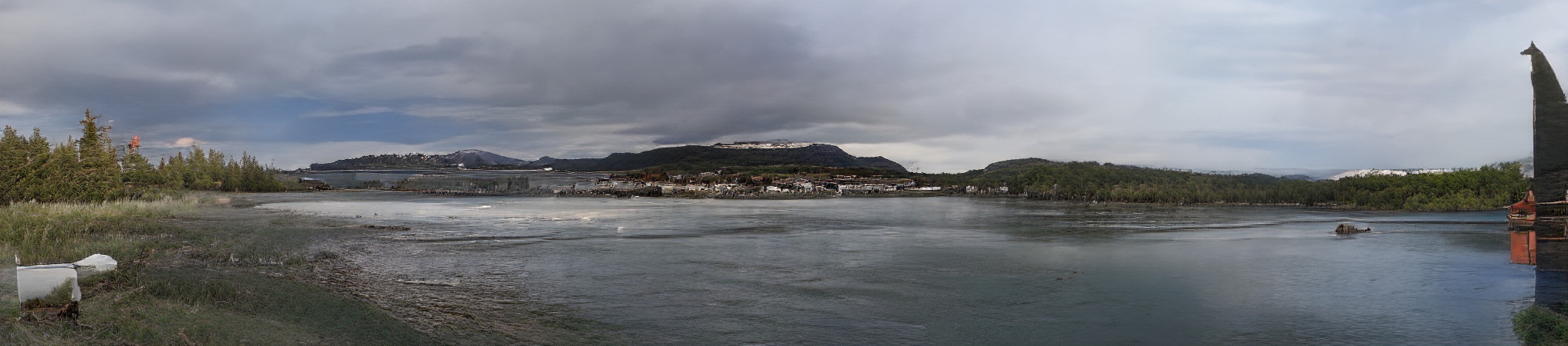}	
	\includegraphics[width=0.99\textwidth]{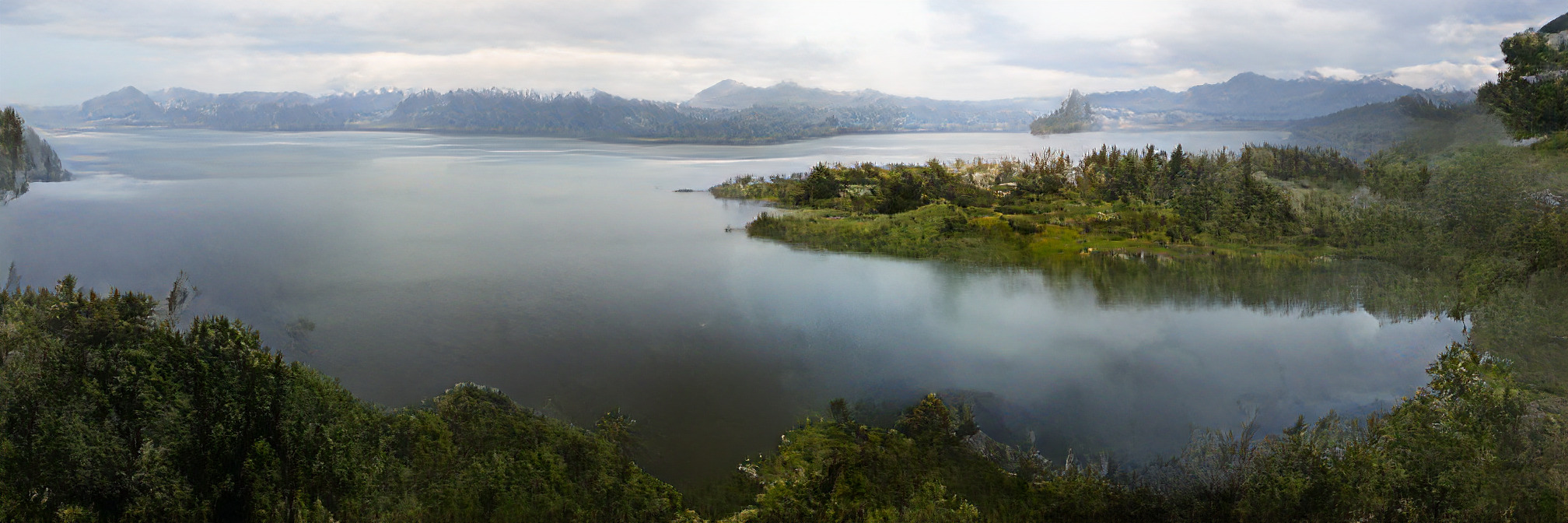}	
  \caption{Samples generated from semantic layouts on S-FLCKR. Sizes from
    top-to-bottom: $2048\times 512$, $1460\times 440$, $2032\times 448$ and
    $2016\times 672$
    pixels.}
	\label{fig:bigsamplesbigthree}
	\end{figure}
}

\newcommand{\bigsamplesbigvertical}{
	\begin{figure}
	\centering
	\includegraphics[height=0.495\textheight]{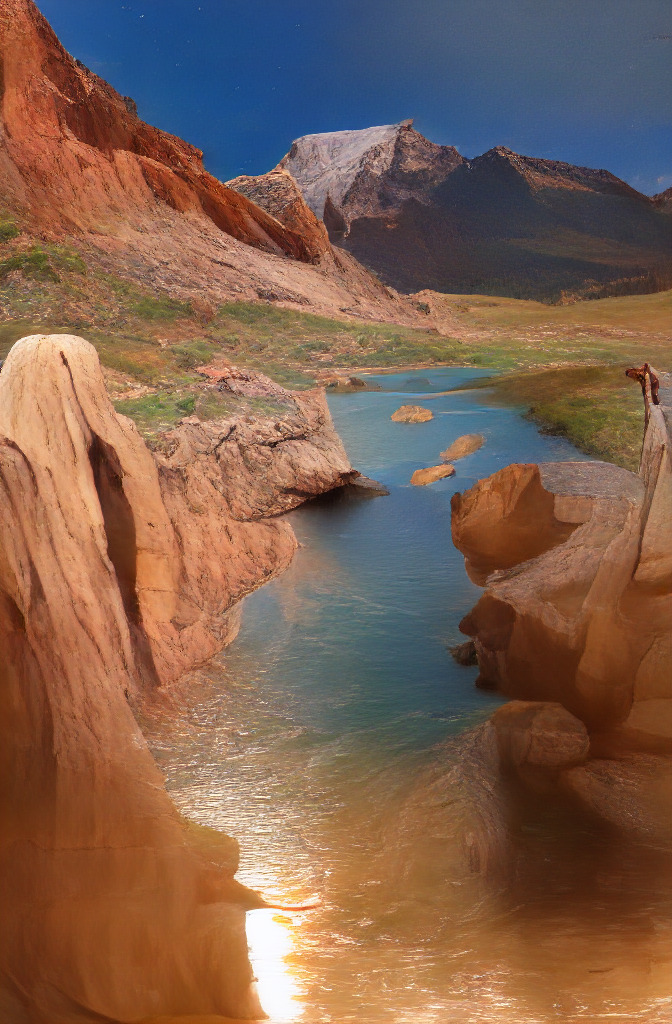}
	\includegraphics[height=0.495\textheight]{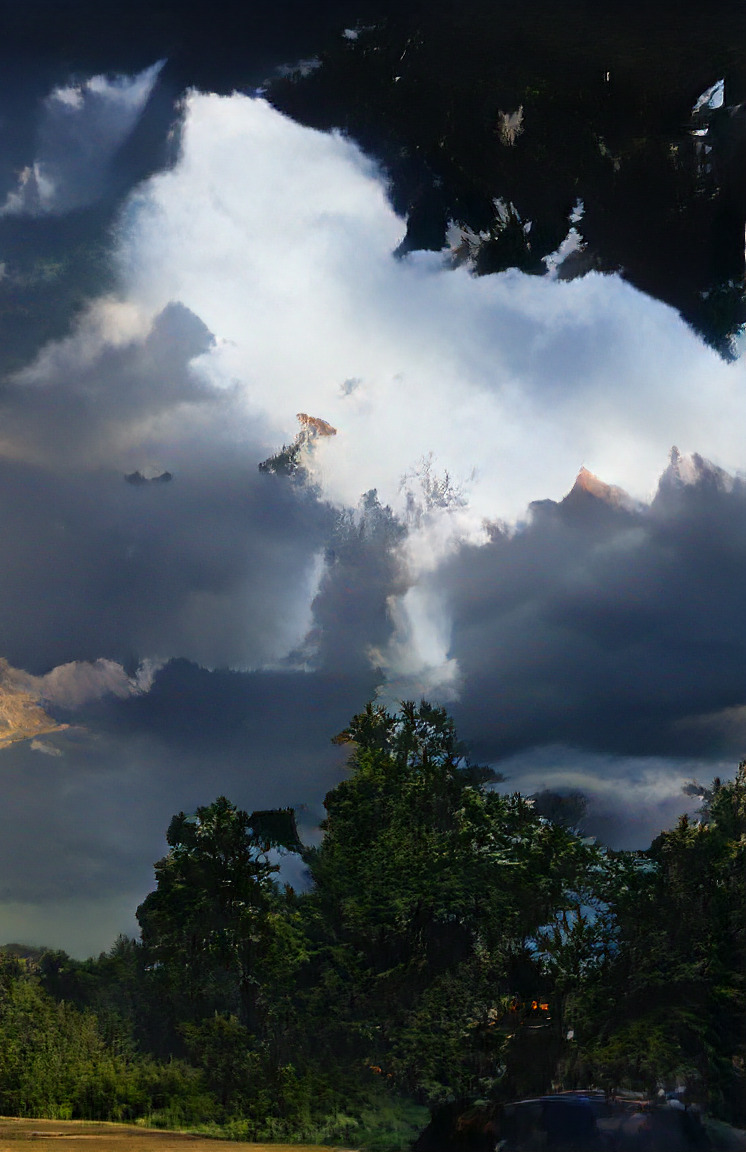}
	\includegraphics[height=0.495\textheight]{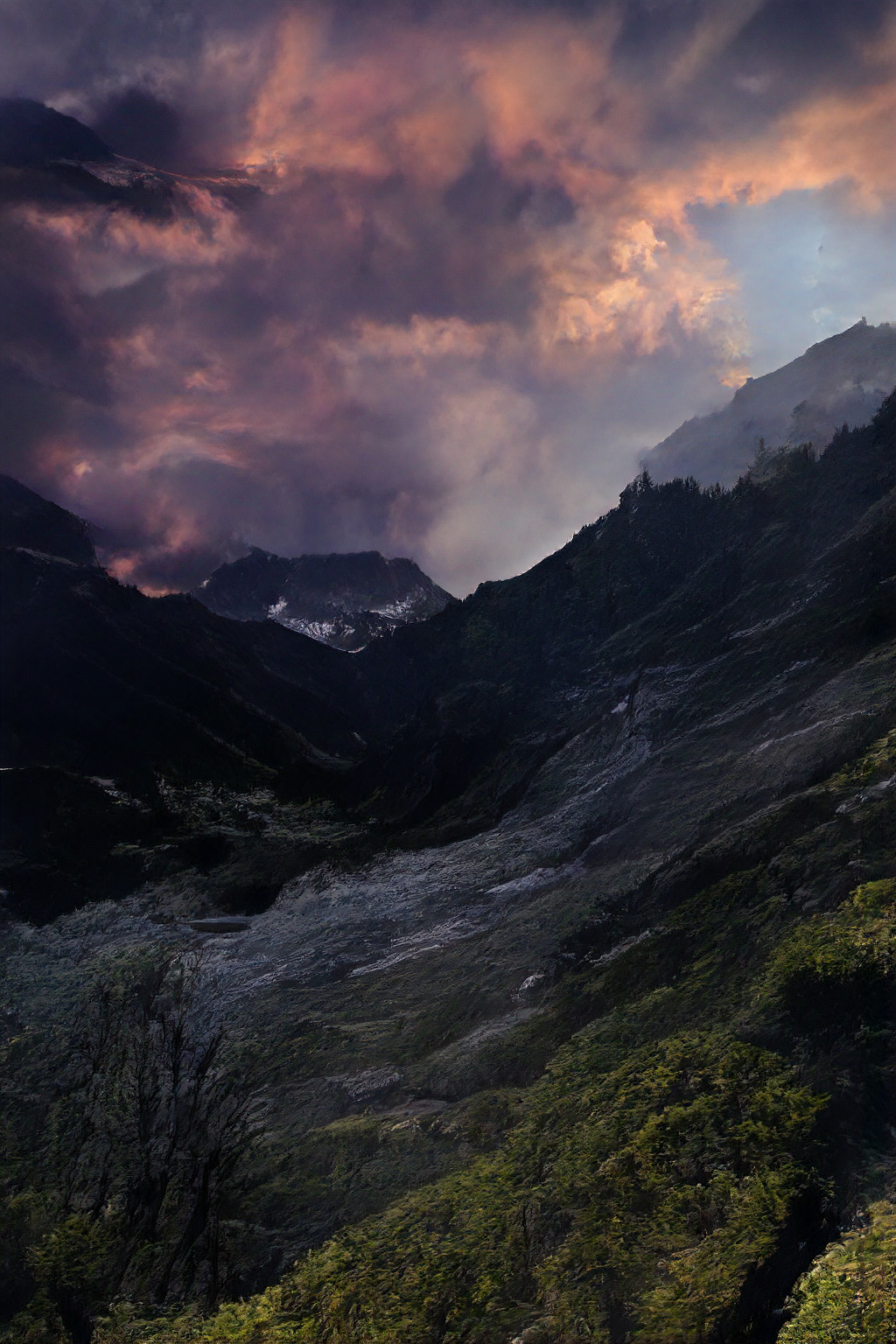}
	\includegraphics[height=0.495\textheight]{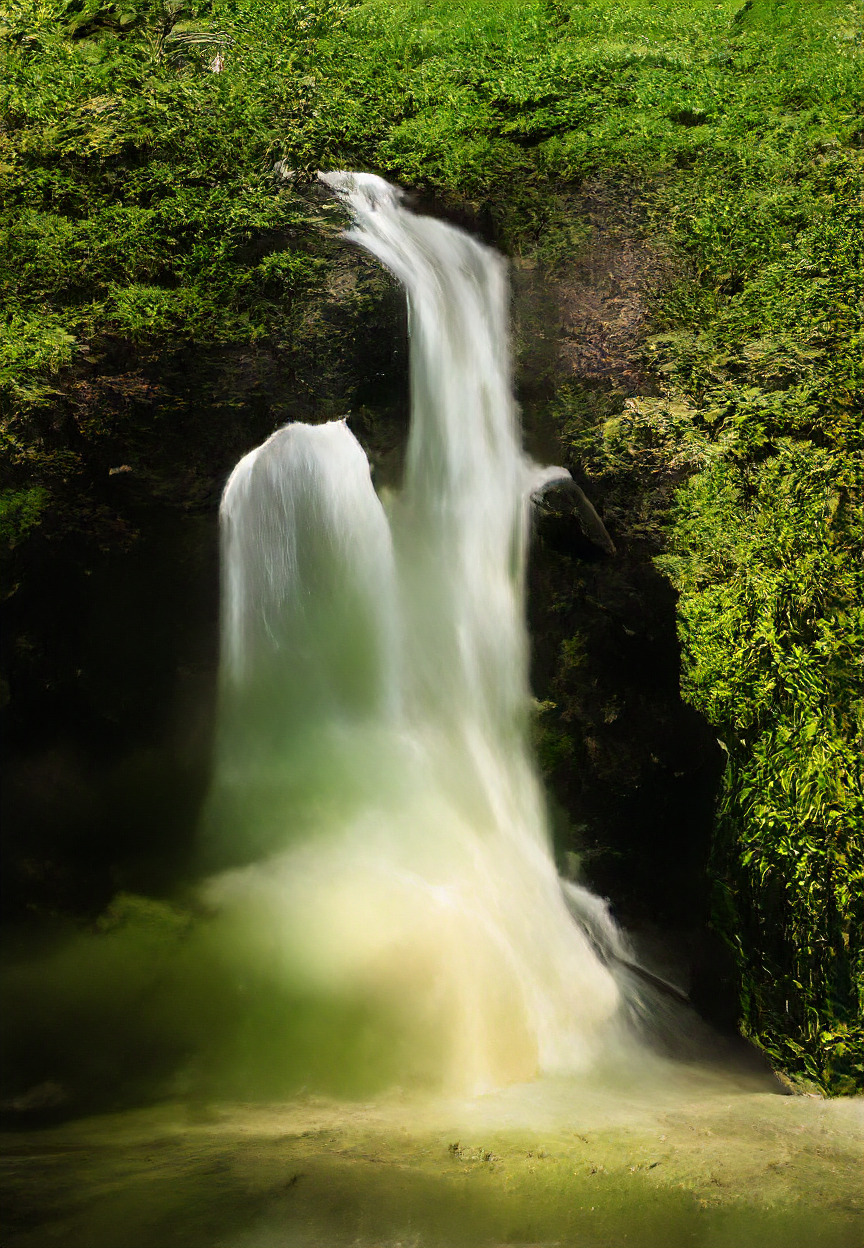}

  \caption{Samples generated from semantic layouts on S-FLCKR. Sizes from
    top-to-bottom: $1280 \times 832$, $1024 \times 416$ and $1280 \times 240$
    pixels.}
	\label{fig:bigsamplesbigvertical}
	\end{figure}
}

\newcommand{\unconditionalchurches}{
	\begin{figure}
	\centering
	\includegraphics[width=0.98\textwidth]{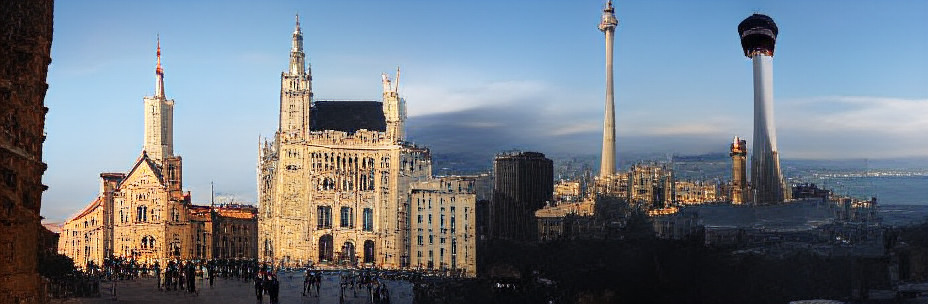}
	\includegraphics[scale=0.36]{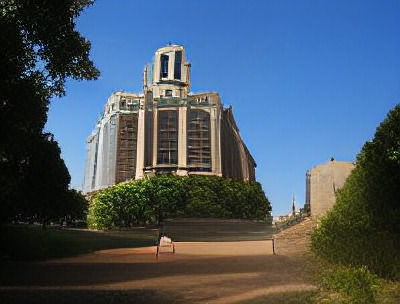}
	\includegraphics[scale=0.36]{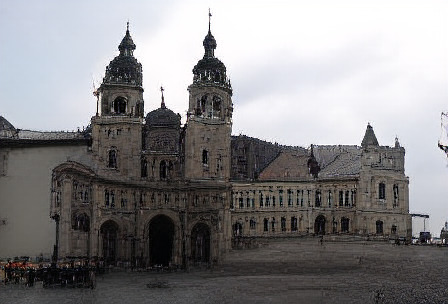}
	\includegraphics[scale=0.36]{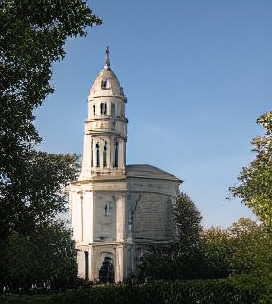}
	
	\includegraphics[scale=0.36]{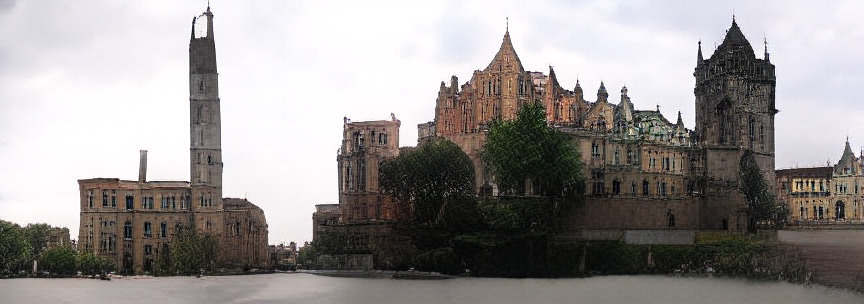}
	\includegraphics[scale=0.36]{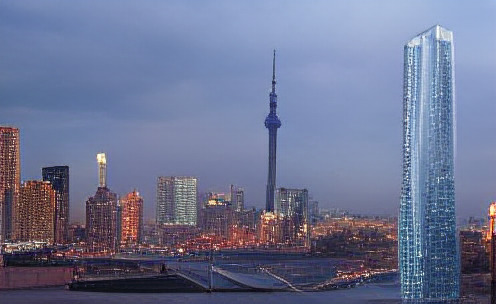}
	
	\includegraphics[scale=0.36]{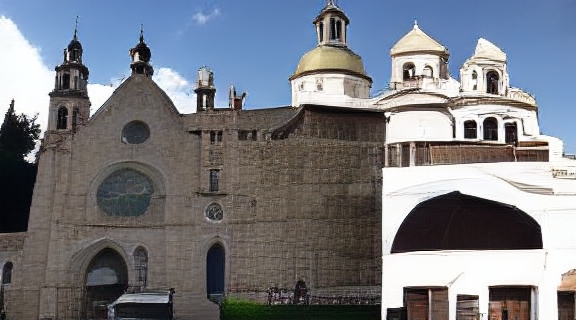}
	\includegraphics[scale=0.36]{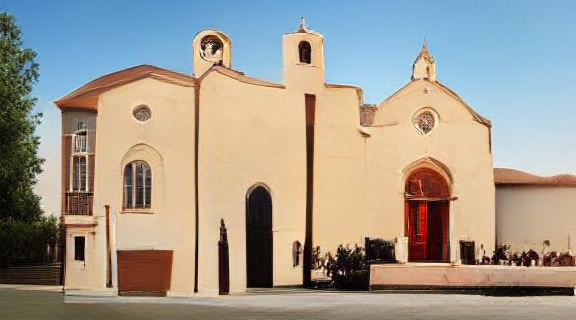}

  \caption{Unconditional samples from a model trained on LSUN Churches \& Towers, using the sliding attention window.}
	\label{fig:unconditionalchurches}

	\end{figure}
}

\newcommand{\ourvsigpt}{
\begin{figure*}[tbp]
  \renewcommand{\impatha}[1]{img/igpt/theirs/##1}
  \renewcommand{\impathb}[1]{img/igpt/ours/##1}
  \renewcommand{\imwidth}{0.24\textwidth}
  \setlength{\tabcolsep}{0pt}
  \centering
  \begin{tabular}{c ccc}
    conditioning & \multicolumn{3}{c}{\textbf{ours} (top) vs iGPT \cite{chen2020generative} (bottom)}\\
    \toprule
    \multirow{2}{*}{\includegraphics[width=\imwidth, align=c]{\impathb{input2}} \phantom{a}}
	&     	
	\includegraphics[width=\imwidth, align=c]{\impathb{igpt_2_ours_s2}} &	         
	\includegraphics[width=\imwidth, align=c]{\impathb{igpt_2_ours_s3}} &
	\includegraphics[width=\imwidth, align=c]{\impathb{igpt_2_ours_s4}} \\

	& 
	\includegraphics[width=\imwidth, align=c]{\impatha{igpt_2_s1}} &	         
	\includegraphics[width=\imwidth, align=c]{\impatha{igpt_2_s2}} &
	\includegraphics[width=\imwidth, align=c]{\impatha{igpt_2_s3}} \\

\midrule
	\multirow{2}{*}{\includegraphics[width=\imwidth, align=c]{\impathb{input4}} \phantom{a}}
	&     	
	\includegraphics[width=\imwidth, align=c]{\impathb{igpt_4_ours_s1}} &	         
	\includegraphics[width=\imwidth, align=c]{\impathb{igpt_4_ours_s2}} &
	\includegraphics[width=\imwidth, align=c]{\impathb{igpt_4_ours_s3}} \\

	& 
	\includegraphics[width=\imwidth, align=c]{\impatha{igpt_4_s1}} &	         
	\includegraphics[width=\imwidth, align=c]{\impatha{igpt_4_s2}} &
	\includegraphics[width=\imwidth, align=c]{\impatha{igpt_4_s3}} \\
    \bottomrule  
  \end{tabular}    
  \caption{Comparing our approach with the pixel-based approach of
  \cite{chen2020generative}. Here, we use our $f=16$ S-FLCKR model to obtain
  high-fidelity image completions of the inputs depicted on the left (half
  completions). For each conditioning, we show three of our samples (top) and
  three of \cite{chen2020generative} (bottom).}
  \label{fig:ourvsigpt}
\end{figure*}
}

\newcommand{\ourvsigpttwo}{
\begin{figure*}[tbp]
  \renewcommand{\impatha}[1]{img/igpt/theirs/##1}
  \renewcommand{\impathb}[1]{img/igpt/ours/##1}
  \renewcommand{\imwidth}{0.24\textwidth}
  \setlength{\tabcolsep}{0pt}
  \centering
  \begin{tabular}{c ccc}
    conditioning & \multicolumn{3}{c}{\textbf{ours} (top) vs iGPT \cite{chen2020generative} (bottom)}\\
    \toprule
    \multirow{2}{*}{\includegraphics[width=\imwidth, align=c]{\impathb{input6}} \phantom{a}}
	&     	
	\includegraphics[width=\imwidth, align=c]{\impathb{igpt_6_ours_s1}} &	         
	\includegraphics[width=\imwidth, align=c]{\impathb{igpt_6_ours_s2}} &
	\includegraphics[width=\imwidth, align=c]{\impathb{igpt_6_ours_s6}} \\

	& 
	\includegraphics[width=\imwidth, align=c]{\impatha{igpt_6_s1}} &	         
	\includegraphics[width=\imwidth, align=c]{\impatha{igpt_6_s2}} &
	\includegraphics[width=\imwidth, align=c]{\impatha{igpt_6_s3}} \\

\midrule
	\multirow{2}{*}{\includegraphics[width=\imwidth, align=c]{\impathb{input3}} \phantom{a}}
	&     	
	\includegraphics[width=\imwidth, align=c]{\impathb{igpt_3_ours_s4}} &	         
	\includegraphics[width=\imwidth, align=c]{\impathb{igpt_3_ours_s2}} &
	\includegraphics[width=\imwidth, align=c]{\impathb{igpt_3_ours_s3}} \\

	& 
	\includegraphics[width=\imwidth, align=c]{\impatha{igpt_3_s1}} &	         
	\includegraphics[width=\imwidth, align=c]{\impatha{igpt_3_s2}} &
	\includegraphics[width=\imwidth, align=c]{\impatha{igpt_3_s3}} \\
    \bottomrule  
  \end{tabular}    
  \caption{Comparing our approach with the pixel-based approach of
  \cite{chen2020generative}. Here, we use our $f=16$ S-FLCKR model to obtain
  high-fidelity image completions of the inputs depicted on the left (half
  completions). For each conditioning, we show three of our samples (top) and
  three of \cite{chen2020generative} (bottom).}
  \label{fig:ourvsigpttwo}
\end{figure*}
}

\newcommand{\ourvsspadecoco}{
\begin{figure*}[tbp]
  \renewcommand{\impath}[1]{img/compare_spade/coco/##1}
  \renewcommand{\imwidth}{0.22\textwidth}
  \setlength{\tabcolsep}{1pt}
  \centering
  \begin{tabular}{c c c c}
	conditioning & ground truth & \textbf{ours} & SPADE \cite{park2019semantic}\\
    \toprule
 	\includegraphics[width=\imwidth, align=c]{\impath{000000017207/ours/000000017207_conditioning_02289-0000}} &	         
	\includegraphics[width=\imwidth, align=c]{\impath{000000017207/ours/000000017207_inputs_02289-0000}} &	         
	\includegraphics[width=\imwidth, align=c]{\impath{000000017207/ours/000000017207}} &
	\includegraphics[width=\imwidth, align=c]{\impath{000000017207/theirs/000000017207}} \\

 	\includegraphics[width=\imwidth, align=c]{\impath{000000033638/ours/000000033638_conditioning_00131-0000}} &	         
	\includegraphics[width=\imwidth, align=c]{\impath{000000033638/ours/000000033638_inputs_00131-0000}} &	         
	\includegraphics[width=\imwidth, align=c]{\impath{000000033638/ours/000000033638}} &
	\includegraphics[width=\imwidth, align=c]{\impath{000000033638/theirs/000000033638}} \\
	
 	\includegraphics[width=\imwidth, align=c]{\impath{000000131131/ours/000000131131_conditioning_00411-0000}} &	         
	\includegraphics[width=\imwidth, align=c]{\impath{000000131131/ours/000000131131_inputs_00411-0000}} &	         
	\includegraphics[width=\imwidth, align=c]{\impath{000000131131/ours/000000131131}} &
	\includegraphics[width=\imwidth, align=c]{\impath{000000131131/theirs/000000131131}} \\
	
 	\includegraphics[width=\imwidth, align=c]{\impath{000000539143/ours/000000539143_conditioning_01180-0000}} &	         
	\includegraphics[width=\imwidth, align=c]{\impath{000000539143/ours/000000539143_inputs_01180-0000}} &	         
	\includegraphics[width=\imwidth, align=c]{\impath{000000539143/ours/000000539143}} &
	\includegraphics[width=\imwidth, align=c]{\impath{000000539143/theirs/000000539143}} \\
	
 	\includegraphics[width=\imwidth, align=c]{\impath{000000521956/ours/000000521956_conditioning_03571-0000}} &	         
	\includegraphics[width=\imwidth, align=c]{\impath{000000521956/ours/000000521956_inputs_03571-0000}} &	         
	\includegraphics[width=\imwidth, align=c]{\impath{000000521956/ours/000000521956}} &
	\includegraphics[width=\imwidth, align=c]{\impath{000000521956/theirs/000000521956}} \\
	
 	\includegraphics[width=\imwidth, align=c]{\impath{000000451150/ours/000000451150_conditioning_01346-0000}} &	         
	\includegraphics[width=\imwidth, align=c]{\impath{000000451150/ours/000000451150_inputs_01346-0000}} &	         
	\includegraphics[width=\imwidth, align=c]{\impath{000000451150/ours/000000451150}} &
	\includegraphics[width=\imwidth, align=c]{\impath{000000451150/theirs/000000451150}} \\
    \bottomrule  
  \end{tabular}    
  \caption{Qualitative comparison to \cite{park2019semantic} on $256 \times 256$ images from the COCO-Stuff dataset.}
  \label{fig:ourvsspadecoco}
\end{figure*}
}

\newcommand{\ourvsspadeade}{
\begin{figure*}[tbp]
  \renewcommand{\impath}[1]{img/compare_spade/ade20k/##1}
  \renewcommand{\imwidth}{0.22\textwidth}
  \setlength{\tabcolsep}{1pt}
  \centering
  \begin{tabular}{c c c c}
	conditioning & ground truth & \textbf{ours} & SPADE \cite{park2019semantic} \\
    \toprule
 	\includegraphics[width=\imwidth, align=c]{\impath{00000407/ours/ADE_val_00000407_conditioning_01228-0000}} &	         
	\includegraphics[width=\imwidth, align=c]{\impath{00000407/ours/ADE_val_00000407_inputs_01228-0000}} &	         
	\includegraphics[width=\imwidth, align=c]{\impath{00000407/ours/ADE_val_00000407_samples_onepix_01228-0002}} &
	\includegraphics[width=\imwidth, align=c]{\impath{00000407/theirs/ADE_val_00000407}} \\

 	\includegraphics[width=\imwidth, align=c]{\impath{00000390/ours/ADE_val_00000390_conditioning_00484-0000}} &	         
	\includegraphics[width=\imwidth, align=c]{\impath{00000390/ours/ADE_val_00000390_inputs_00484-0000}} &	         
	\includegraphics[width=\imwidth, align=c]{\impath{00000390/ours/ADE_val_00000390_samples_onepix_00484-0001}} &
	\includegraphics[width=\imwidth, align=c]{\impath{00000390/theirs/ADE_val_00000390}} \\        
    
 	\includegraphics[width=\imwidth, align=c]{\impath{00000266/ours/ADE_val_00000266_conditioning_01693-0000}} &	         
	\includegraphics[width=\imwidth, align=c]{\impath{00000266/ours/ADE_val_00000266_inputs_01693-0000}} &	         
	\includegraphics[width=\imwidth, align=c]{\impath{00000266/ours/ADE_val_00000266_samples_onepix_01693-0001}} &
	\includegraphics[width=\imwidth, align=c]{\impath{00000266/theirs/ADE_val_00000266}} \\    

 	\includegraphics[width=\imwidth, align=c]{\impath{00001325/ours/ADE_val_00001325_conditioning_01632-0000}} &	         
	\includegraphics[width=\imwidth, align=c]{\impath{00001325/ours/ADE_val_00001325_inputs_01632-0000}} &	         
	\includegraphics[width=\imwidth, align=c]{\impath{00001325/ours/ADE_val_00001325_samples_onepix_01632-0000}} &
	\includegraphics[width=\imwidth, align=c]{\impath{00001325/theirs/ADE_val_00001325}} \\    

 	\includegraphics[width=\imwidth, align=c]{\impath{00000145/ours/ADE_val_00000145_conditioning_01187-0000}} &	         
	\includegraphics[width=\imwidth, align=c]{\impath{00000145/ours/ADE_val_00000145_inputs_01187-0002}} &	         
	\includegraphics[width=\imwidth, align=c]{\impath{00000145/ours/ADE_val_00000145_samples_onepix_01187-0000}} &
	\includegraphics[width=\imwidth, align=c]{\impath{00000145/theirs/ADE_val_00000145}} \\    

 	\includegraphics[width=\imwidth, align=c]{\impath{00000486/ours/ADE_val_00000486_conditioning_01824-0000}} &	         
	\includegraphics[width=\imwidth, align=c]{\impath{00000486/ours/ADE_val_00000486_inputs_01824-0000}} &	         
	\includegraphics[width=\imwidth, align=c]{\impath{00000486/ours/ADE_val_00000486_samples_onepix_01824-0001}} &
	\includegraphics[width=\imwidth, align=c]{\impath{00000486/theirs/ADE_val_00000486}} \\
    \bottomrule  
  \end{tabular}    
  \caption{Qualitative comparison to \cite{park2019semantic} on $256 \times 256$ images from the ADE20K dataset.}
  \label{fig:ourvsspadeade}
\end{figure*}
}

\newcommand{\permutations}{
\begin{figure*}[tbp]
\centering
\begin{tabular}{c c c c c c} 
  \scriptsize{row major} & \scriptsize{spiral in} & \scriptsize{spiral out} & \scriptsize{z-curve} & \scriptsize{subsample} & \scriptsize{alternate} \\
  \includegraphics[width=0.14\linewidth]{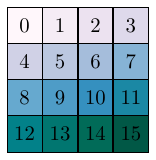} &
  \includegraphics[width=0.14\linewidth]{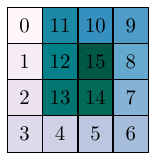} &
  \includegraphics[width=0.14\linewidth]{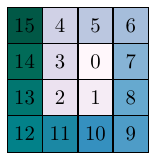} &
  \includegraphics[width=0.14\linewidth]{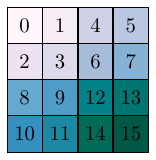} &
  \includegraphics[width=0.14\linewidth]{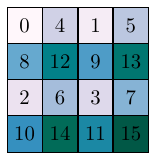} &
  \includegraphics[width=0.14\linewidth]{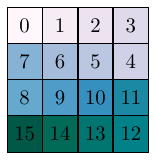} 
\end{tabular}
\includegraphics[width=0.49\linewidth]{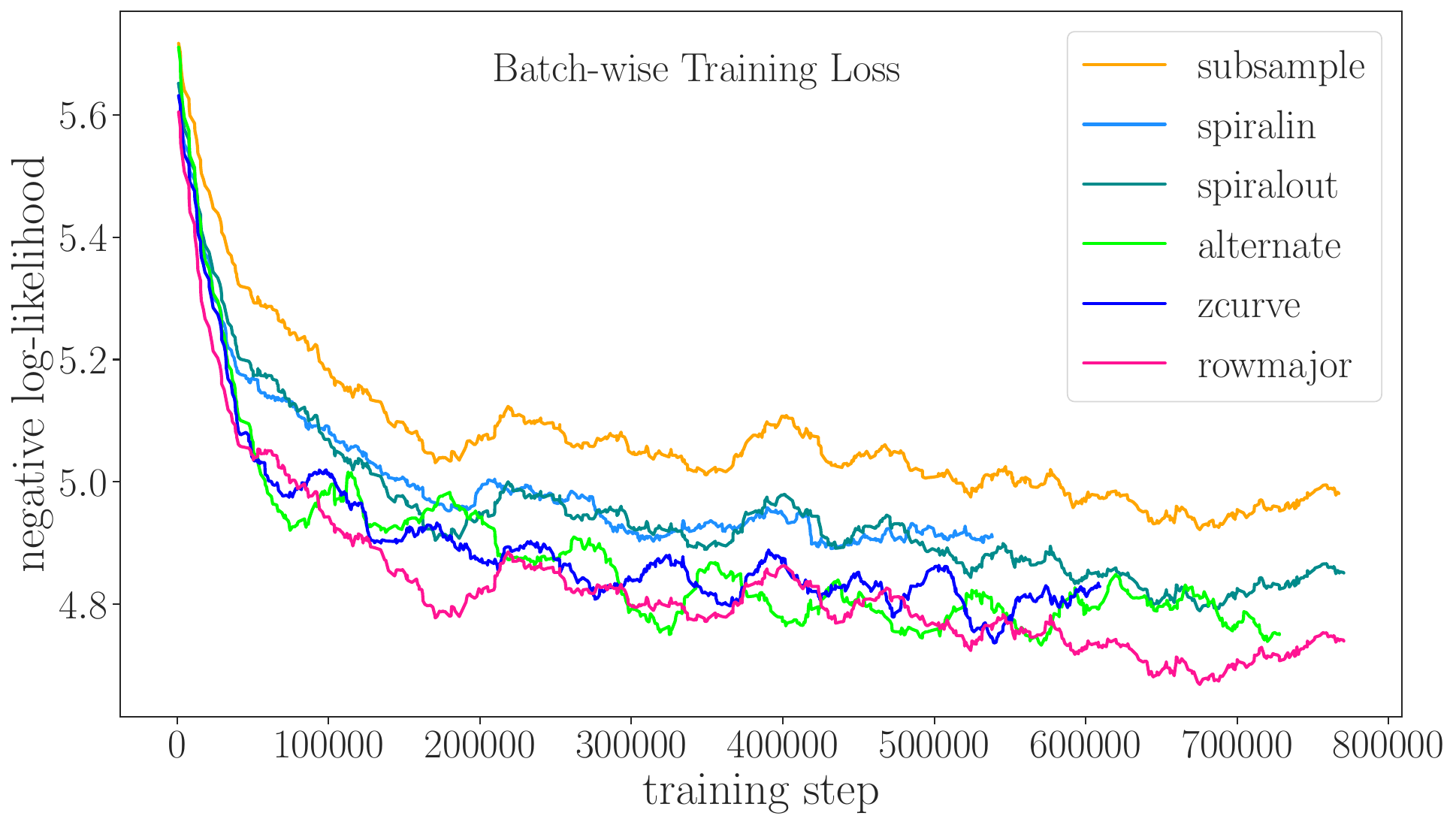}
\includegraphics[width=0.49\linewidth]{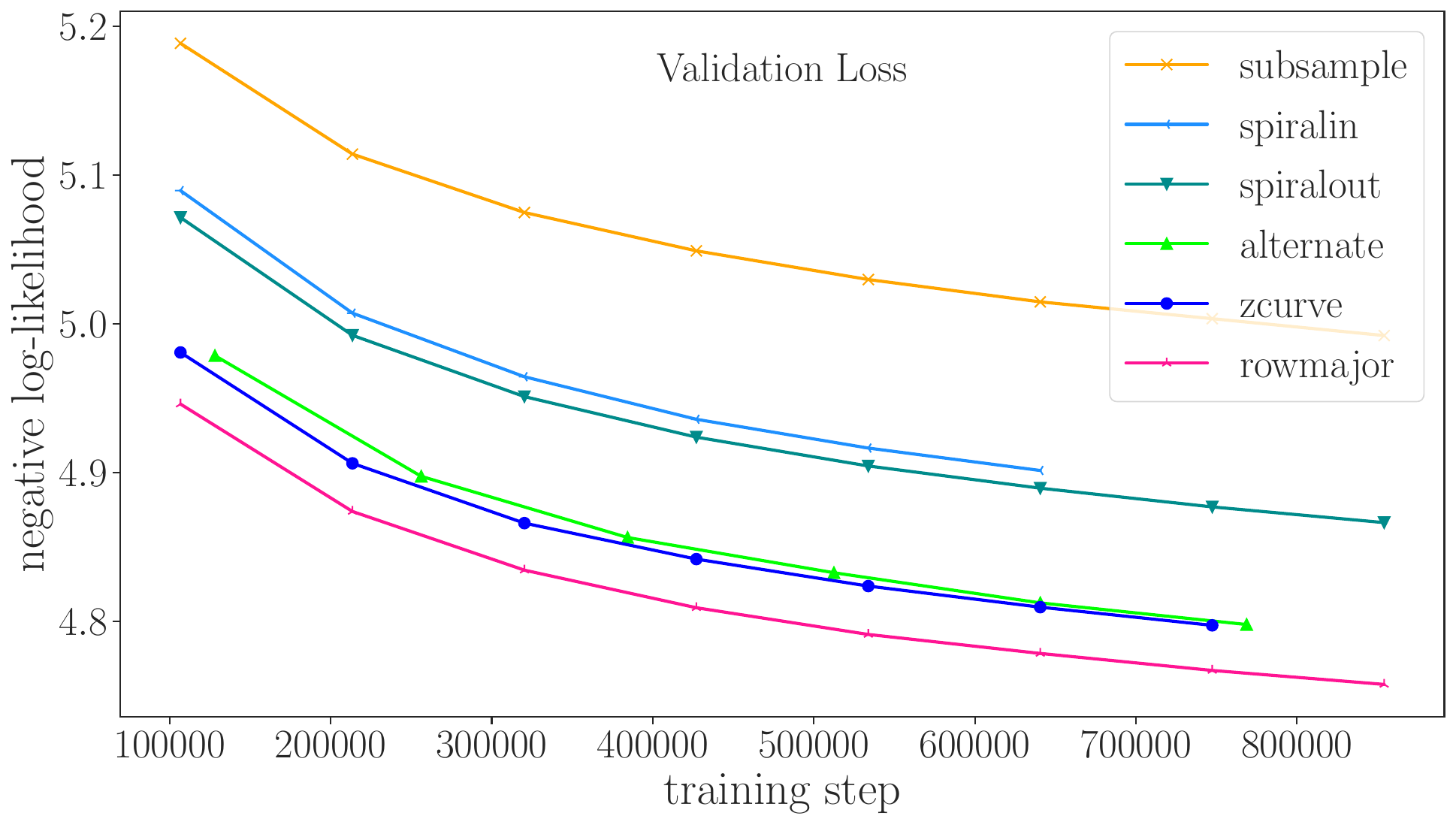}
\caption{Top: All sequence permutations we investigate, illustrated on a
  $4 \times4$ grid. Bottom: The transformer architecture is permutation invariant but next-token
  prediction is not: The average loss on the validation split of ImageNet,
  corresponding to the negative log-likelihood, differs significantly between
  different prediction orderings. Among our choices, the commonly used
  row-major order performs best.}
\label{fig:permutations}
\end{figure*}
}

\newcommand{\indepthsamples}{
\begin{figure}[btp]
  \renewcommand{\impath}[1]{img/depthtoin/##1}
  \renewcommand{\imwidth}{0.1\textwidth}
  \setlength{\tabcolsep}{2pt}
  \centering
  \begin{tabular}{ c  c  c  c  c  c  c  c}
    conditioning & \multicolumn{3}{c}{samples} & conditioning & \multicolumn{3}{c}{samples} \\
    \toprule
	\includegraphics[width=\imwidth, align=c]{\impath{conditioning_16216-0001}} &     
	\includegraphics[width=\imwidth, align=c]{\impath{samples_onepix_16216-0000}} &         
	\includegraphics[width=\imwidth, align=c]{\impath{samples_onepix_16216-0002}} &         
	\includegraphics[width=\imwidth, align=c]{\impath{samples_onepix_16216-0003}} &
	
	\includegraphics[width=\imwidth, align=c]{\impath{conditioning_16331-0001}} &     
	\includegraphics[width=\imwidth, align=c]{\impath{samples_onepix_16331-0000}} &         
	\includegraphics[width=\imwidth, align=c]{\impath{samples_onepix_16331-0002}} &         
	\includegraphics[width=\imwidth, align=c]{\impath{samples_onepix_16331-0003}} \\
	\midrule

	\includegraphics[width=\imwidth, align=c]{\impath{conditioning_17507-0001}} &     
	\includegraphics[width=\imwidth, align=c]{\impath{samples_onepix_17507-0000}} &         
	\includegraphics[width=\imwidth, align=c]{\impath{samples_onepix_17507-0002}} &         
	\includegraphics[width=\imwidth, align=c]{\impath{samples_onepix_17507-0003}} &

	\includegraphics[width=\imwidth, align=c]{\impath{conditioning_09438-0001}} &     
	\includegraphics[width=\imwidth, align=c]{\impath{samples_onepix_09438-0003}} &         
	\includegraphics[width=\imwidth, align=c]{\impath{samples_onepix_09438-0004}} &         
	\includegraphics[width=\imwidth, align=c]{\impath{samples_onepix_09438-0000}} \\

	\midrule
	\includegraphics[width=\imwidth, align=c]{\impath{conditioning_26548-0001}} &     
	\includegraphics[width=\imwidth, align=c]{\impath{samples_onepix_26548-0000}} &         
	\includegraphics[width=\imwidth, align=c]{\impath{samples_onepix_26548-0002}} &         
	\includegraphics[width=\imwidth, align=c]{\impath{samples_onepix_26548-0001}} &
	
	\includegraphics[width=\imwidth, align=c]{\impath{conditioning_22732-0001}} &     
	\includegraphics[width=\imwidth, align=c]{\impath{samples_onepix_22732-0000}} &         
	\includegraphics[width=\imwidth, align=c]{\impath{samples_onepix_22732-0001}} &         
	\includegraphics[width=\imwidth, align=c]{\impath{samples_onepix_22732-0003}} \\
    \bottomrule  
  \end{tabular}    
  \caption{Conditional samples for the depth-to-image model on IN.}
  \label{fig:depthinsamples}
\end{figure}
}

\newcommand{\sticktoreal}{
\begin{figure}[tbp]
  \renewcommand{\impath}[1]{img/deepfashion/##1}
  \renewcommand{\imwidth}{0.1\textwidth}
  \setlength{\tabcolsep}{2pt}
  \centering
  \begin{tabular}{ c  c  c  c  c  c  c  c}
    conditioning & \multicolumn{3}{c}{samples} & conditioning & \multicolumn{3}{c}{samples} \\
    \toprule
	\includegraphics[width=\imwidth, align=c]{\impath{condin2726}} &     
	\includegraphics[width=\imwidth, align=c]{\impath{full2726_1}} &         
	\includegraphics[width=\imwidth, align=c]{\impath{full2726_2}} &         
	\includegraphics[width=\imwidth, align=c]{\impath{full2726_3}} &
	
	\includegraphics[width=\imwidth, align=c]{\impath{condin640}} &     
	\includegraphics[width=\imwidth, align=c]{\impath{full640_1}} &         
	\includegraphics[width=\imwidth, align=c]{\impath{full640_2}} &         
	\includegraphics[width=\imwidth, align=c]{\impath{full640_3}} \\
	\midrule
	\includegraphics[width=\imwidth, align=c]{\impath{condin0}} &     
	\includegraphics[width=\imwidth, align=c]{\impath{full0_1}} &         
	\includegraphics[width=\imwidth, align=c]{\impath{full0_2}} &         
	\includegraphics[width=\imwidth, align=c]{\impath{full0_3}} &
	
	\includegraphics[width=\imwidth, align=c]{\impath{condin2036}} &     
	\includegraphics[width=\imwidth, align=c]{\impath{full2036_1}} &         
	\includegraphics[width=\imwidth, align=c]{\impath{full2036_2}} &         
	\includegraphics[width=\imwidth, align=c]{\impath{full2036_3}} \\
	\midrule
	\includegraphics[width=\imwidth, align=c]{\impath{condin690}} &     
	\includegraphics[width=\imwidth, align=c]{\impath{full690_1}} &         
	\includegraphics[width=\imwidth, align=c]{\impath{full690_2}} &         
	\includegraphics[width=\imwidth, align=c]{\impath{full690_3}} &
	
	\includegraphics[width=\imwidth, align=c]{\impath{condin903}} &     
	\includegraphics[width=\imwidth, align=c]{\impath{full903_1}} &         
	\includegraphics[width=\imwidth, align=c]{\impath{full903_2}} &         
	\includegraphics[width=\imwidth, align=c]{\impath{full903_3}} \\
    \bottomrule  
  \end{tabular}    
  \caption{Conditional samples for the pose-guided synthesis model via keypoints on DeepFashion.}
  \label{fig:dfsamples}
\end{figure}
}

\newcommand{\rinclasscond}{
\begin{figure}[tbp]
  \renewcommand{\impath}[1]{img/classcondrin/##1}
  \renewcommand{\imwidth}{0.1\textwidth}
  \setlength{\tabcolsep}{2pt}
  \centering
  \begin{tabular}{ c  c  c  c  c  c  c  c}
    \tiny{class exemplar} & \multicolumn{3}{c}{samples} &  \tiny{class exemplar} & \multicolumn{3}{c}{samples} \\
    \toprule
	\includegraphics[width=\imwidth, align=c]{\impath{inputs_00750-0001}} &     
	\includegraphics[width=\imwidth, align=c]{\impath{samples_onepix_00750-0000}} &         
	\includegraphics[width=\imwidth, align=c]{\impath{samples_onepix_00750-0001}} &         
	\includegraphics[width=\imwidth, align=c]{\impath{samples_onepix_00750-0003}} &
	
	\includegraphics[width=\imwidth, align=c]{\impath{inputs_08334-0001}} &     
	\includegraphics[width=\imwidth, align=c]{\impath{samples_onepix_08334-0001}} &         
	\includegraphics[width=\imwidth, align=c]{\impath{samples_onepix_08334-0005}} &         
	\includegraphics[width=\imwidth, align=c]{\impath{samples_onepix_08334-0000}} \\
	\midrule

	\includegraphics[width=\imwidth, align=c]{\impath{inputs_03448-0001}} &     
	\includegraphics[width=\imwidth, align=c]{\impath{samples_onepix_03448-0004}} &         
	\includegraphics[width=\imwidth, align=c]{\impath{samples_onepix_03448-0002}} &         
	\includegraphics[width=\imwidth, align=c]{\impath{samples_onepix_03448-0003}} &

	\includegraphics[width=\imwidth, align=c]{\impath{inputs_07187-0001}} &     
	\includegraphics[width=\imwidth, align=c]{\impath{samples_onepix_07187-0004}} &         
	\includegraphics[width=\imwidth, align=c]{\impath{samples_onepix_07187-0002}} &         
	\includegraphics[width=\imwidth, align=c]{\impath{samples_onepix_07187-0003}} \\

	\midrule
	\includegraphics[width=\imwidth, align=c]{\impath{inputs_07332-0001}} &     
	\includegraphics[width=\imwidth, align=c]{\impath{samples_onepix_07332-0000}} &         
	\includegraphics[width=\imwidth, align=c]{\impath{samples_onepix_07332-0004}} &         
	\includegraphics[width=\imwidth, align=c]{\impath{samples_onepix_07332-0001}} &
	
	\includegraphics[width=\imwidth, align=c]{\impath{inputs_09274-0001}} &     
	\includegraphics[width=\imwidth, align=c]{\impath{samples_onepix_09274-0003}} &         
	\includegraphics[width=\imwidth, align=c]{\impath{samples_onepix_09274-0002}} &         
	\includegraphics[width=\imwidth, align=c]{\impath{samples_onepix_09274-0001}} \\

    \bottomrule  
  \end{tabular}    
  \caption{Samples produced by the class-conditional model trained on RIN.}
  \label{fig:rinclasscond}
\end{figure}
}

\newcommand{\permuationsbatches}{
\begin{figure}[tbp]
  \renewcommand{\impath}[1]{img/permutations/##1}
  \renewcommand{\imwidth}{0.075\textwidth}
  \setlength{\tabcolsep}{0pt}
  \centering
  \begin{tabular}{ c  c  c  c  c    c    c  c  c  c  c}
     \multicolumn{5}{c}{Row Major} & & \multicolumn{5}{c}{Subsample}\\
    \toprule
	\includegraphics[width=\imwidth, align=c]{\impath{rowmajor/samples_onepix_01562-0000}} &     
	\includegraphics[width=\imwidth, align=c]{\impath{rowmajor/samples_onepix_01562-0001}} &         
	\includegraphics[width=\imwidth, align=c]{\impath{rowmajor/samples_onepix_01562-0002}} &         
	\includegraphics[width=\imwidth, align=c]{\impath{rowmajor/samples_onepix_01562-0003}} &
	\includegraphics[width=\imwidth, align=c]{\impath{rowmajor/samples_onepix_01562-0004}} &     
    \phantom{x} &
	\includegraphics[width=\imwidth, align=c]{\impath{scalespace/samples_onepix_05096-0000}} &
	\includegraphics[width=\imwidth, align=c]{\impath{scalespace/samples_onepix_05096-0001}} &
	\includegraphics[width=\imwidth, align=c]{\impath{scalespace/samples_onepix_05096-0002}} &         
	\includegraphics[width=\imwidth, align=c]{\impath{scalespace/samples_onepix_05096-0003}} &         
	\includegraphics[width=\imwidth, align=c]{\impath{scalespace/samples_onepix_05096-0004}} \\
	
	\includegraphics[width=\imwidth, align=c]{\impath{rowmajor/samples_onepix_04384-0000}} &     
	\includegraphics[width=\imwidth, align=c]{\impath{rowmajor/samples_onepix_04384-0001}} &         
	\includegraphics[width=\imwidth, align=c]{\impath{rowmajor/samples_onepix_04384-0002}} &         
	\includegraphics[width=\imwidth, align=c]{\impath{rowmajor/samples_onepix_04384-0003}} &
	\includegraphics[width=\imwidth, align=c]{\impath{rowmajor/samples_onepix_04384-0004}} &     
    \phantom{x} &
	\includegraphics[width=\imwidth, align=c]{\impath{scalespace/samples_onepix_05933-0000}} &
	\includegraphics[width=\imwidth, align=c]{\impath{scalespace/samples_onepix_05933-0001}} &
	\includegraphics[width=\imwidth, align=c]{\impath{scalespace/samples_onepix_05933-0002}} &         
	\includegraphics[width=\imwidth, align=c]{\impath{scalespace/samples_onepix_05933-0003}} &         
	\includegraphics[width=\imwidth, align=c]{\impath{scalespace/samples_onepix_05933-0004}} \\

	\includegraphics[width=\imwidth, align=c]{\impath{rowmajor/samples_onepix_10575-0000}} &     
	\includegraphics[width=\imwidth, align=c]{\impath{rowmajor/samples_onepix_10575-0001}} &         
	\includegraphics[width=\imwidth, align=c]{\impath{rowmajor/samples_onepix_10575-0002}} &         
	\includegraphics[width=\imwidth, align=c]{\impath{rowmajor/samples_onepix_10575-0003}} &
	\includegraphics[width=\imwidth, align=c]{\impath{rowmajor/samples_onepix_10575-0004}} &     
    \phantom{x} &
	\includegraphics[width=\imwidth, align=c]{\impath{scalespace/samples_onepix_11371-0000}} &
	\includegraphics[width=\imwidth, align=c]{\impath{scalespace/samples_onepix_11371-0001}} &
	\includegraphics[width=\imwidth, align=c]{\impath{scalespace/samples_onepix_11371-0002}} &         
	\includegraphics[width=\imwidth, align=c]{\impath{scalespace/samples_onepix_11371-0003}} &         
	\includegraphics[width=\imwidth, align=c]{\impath{scalespace/samples_onepix_11371-0004}} \\

	\includegraphics[width=\imwidth, align=c]{\impath{rowmajor/samples_onepix_10781-0000}} &     
	\includegraphics[width=\imwidth, align=c]{\impath{rowmajor/samples_onepix_10781-0001}} &         
	\includegraphics[width=\imwidth, align=c]{\impath{rowmajor/samples_onepix_10781-0002}} &         
	\includegraphics[width=\imwidth, align=c]{\impath{rowmajor/samples_onepix_10781-0003}} &
	\includegraphics[width=\imwidth, align=c]{\impath{rowmajor/samples_onepix_10781-0004}} &     
    \phantom{x} &
	\includegraphics[width=\imwidth, align=c]{\impath{scalespace/samples_onepix_14045-0000}} &
	\includegraphics[width=\imwidth, align=c]{\impath{scalespace/samples_onepix_14045-0001}} &
	\includegraphics[width=\imwidth, align=c]{\impath{scalespace/samples_onepix_14045-0002}} &         
	\includegraphics[width=\imwidth, align=c]{\impath{scalespace/samples_onepix_14045-0003}} &         
	\includegraphics[width=\imwidth, align=c]{\impath{scalespace/samples_onepix_14045-0004}} \\
	
	\includegraphics[width=\imwidth, align=c]{\impath{rowmajor/samples_onepix_14942-0000}} &     
	\includegraphics[width=\imwidth, align=c]{\impath{rowmajor/samples_onepix_14942-0001}} &         
	\includegraphics[width=\imwidth, align=c]{\impath{rowmajor/samples_onepix_14942-0002}} &         
	\includegraphics[width=\imwidth, align=c]{\impath{rowmajor/samples_onepix_14942-0003}} &
	\includegraphics[width=\imwidth, align=c]{\impath{rowmajor/samples_onepix_14942-0004}} &     
    \phantom{x} &
	\includegraphics[width=\imwidth, align=c]{\impath{scalespace/samples_onepix_16437-0000}} &
	\includegraphics[width=\imwidth, align=c]{\impath{scalespace/samples_onepix_16437-0001}} &
	\includegraphics[width=\imwidth, align=c]{\impath{scalespace/samples_onepix_16437-0002}} &         
	\includegraphics[width=\imwidth, align=c]{\impath{scalespace/samples_onepix_16437-0003}} &         
	\includegraphics[width=\imwidth, align=c]{\impath{scalespace/samples_onepix_16437-0004}} \\

   	\midrule
    \multicolumn{5}{c}{Z-Curve} & & \multicolumn{5}{c}{Spiral Out}\\
   	\midrule

	\includegraphics[width=\imwidth, align=c]{\impath{zcurve/samples_onepix_01799-0000}} &     
	\includegraphics[width=\imwidth, align=c]{\impath{zcurve/samples_onepix_01799-0001}} &         
	\includegraphics[width=\imwidth, align=c]{\impath{zcurve/samples_onepix_01799-0002}} &         
	\includegraphics[width=\imwidth, align=c]{\impath{zcurve/samples_onepix_01799-0003}} &
	\includegraphics[width=\imwidth, align=c]{\impath{zcurve/samples_onepix_01799-0004}} &     
  \phantom{x} &
	\includegraphics[width=\imwidth, align=c]{\impath{spiralout/samples_onepix_06109-0000}} &
	\includegraphics[width=\imwidth, align=c]{\impath{spiralout/samples_onepix_06109-0001}} &
	\includegraphics[width=\imwidth, align=c]{\impath{spiralout/samples_onepix_06109-0002}} &         
	\includegraphics[width=\imwidth, align=c]{\impath{spiralout/samples_onepix_06109-0003}} &         
	\includegraphics[width=\imwidth, align=c]{\impath{spiralout/samples_onepix_06109-0004}} \\

	\includegraphics[width=\imwidth, align=c]{\impath{zcurve/samples_onepix_04581-0000}} &     
	\includegraphics[width=\imwidth, align=c]{\impath{zcurve/samples_onepix_04581-0001}} &         
	\includegraphics[width=\imwidth, align=c]{\impath{zcurve/samples_onepix_04581-0002}} &         
	\includegraphics[width=\imwidth, align=c]{\impath{zcurve/samples_onepix_04581-0003}} &
	\includegraphics[width=\imwidth, align=c]{\impath{zcurve/samples_onepix_04581-0004}} &     
  \phantom{x} &
	\includegraphics[width=\imwidth, align=c]{\impath{spiralout/samples_onepix_07032-0000}} &
	\includegraphics[width=\imwidth, align=c]{\impath{spiralout/samples_onepix_07032-0001}} &
	\includegraphics[width=\imwidth, align=c]{\impath{spiralout/samples_onepix_07032-0002}} &         
	\includegraphics[width=\imwidth, align=c]{\impath{spiralout/samples_onepix_07032-0003}} &         
	\includegraphics[width=\imwidth, align=c]{\impath{spiralout/samples_onepix_07032-0004}} \\

	\includegraphics[width=\imwidth, align=c]{\impath{zcurve/samples_onepix_05591-0000}} &     
	\includegraphics[width=\imwidth, align=c]{\impath{zcurve/samples_onepix_05591-0001}} &         
	\includegraphics[width=\imwidth, align=c]{\impath{zcurve/samples_onepix_05591-0002}} &         
	\includegraphics[width=\imwidth, align=c]{\impath{zcurve/samples_onepix_05591-0003}} &
	\includegraphics[width=\imwidth, align=c]{\impath{zcurve/samples_onepix_05591-0004}} &     
  \phantom{x} &
	\includegraphics[width=\imwidth, align=c]{\impath{spiralout/samples_onepix_08449-0000}} &
	\includegraphics[width=\imwidth, align=c]{\impath{spiralout/samples_onepix_08449-0001}} &
	\includegraphics[width=\imwidth, align=c]{\impath{spiralout/samples_onepix_08449-0002}} &         
	\includegraphics[width=\imwidth, align=c]{\impath{spiralout/samples_onepix_08449-0003}} &         
	\includegraphics[width=\imwidth, align=c]{\impath{spiralout/samples_onepix_08449-0004}} \\

	\includegraphics[width=\imwidth, align=c]{\impath{zcurve/samples_onepix_06153-0000}} &     
	\includegraphics[width=\imwidth, align=c]{\impath{zcurve/samples_onepix_06153-0001}} &         
	\includegraphics[width=\imwidth, align=c]{\impath{zcurve/samples_onepix_06153-0002}} &         
	\includegraphics[width=\imwidth, align=c]{\impath{zcurve/samples_onepix_06153-0003}} &
	\includegraphics[width=\imwidth, align=c]{\impath{zcurve/samples_onepix_06153-0004}} &     
  \phantom{x} &
	\includegraphics[width=\imwidth, align=c]{\impath{spiralout/samples_onepix_15989-0000}} &
	\includegraphics[width=\imwidth, align=c]{\impath{spiralout/samples_onepix_15989-0001}} &
	\includegraphics[width=\imwidth, align=c]{\impath{spiralout/samples_onepix_15989-0002}} &         
	\includegraphics[width=\imwidth, align=c]{\impath{spiralout/samples_onepix_15989-0003}} &         
	\includegraphics[width=\imwidth, align=c]{\impath{spiralout/samples_onepix_15989-0004}} \\
	
	\includegraphics[width=\imwidth, align=c]{\impath{zcurve/samples_onepix_08296-0000}} &     
	\includegraphics[width=\imwidth, align=c]{\impath{zcurve/samples_onepix_08296-0001}} &         
	\includegraphics[width=\imwidth, align=c]{\impath{zcurve/samples_onepix_08296-0002}} &         
	\includegraphics[width=\imwidth, align=c]{\impath{zcurve/samples_onepix_08296-0003}} &
	\includegraphics[width=\imwidth, align=c]{\impath{zcurve/samples_onepix_08296-0004}} &     
  \phantom{x} &
	\includegraphics[width=\imwidth, align=c]{\impath{spiralout/samples_onepix_16282-0000}} &
	\includegraphics[width=\imwidth, align=c]{\impath{spiralout/samples_onepix_16282-0001}} &
	\includegraphics[width=\imwidth, align=c]{\impath{spiralout/samples_onepix_16282-0002}} &         
	\includegraphics[width=\imwidth, align=c]{\impath{spiralout/samples_onepix_16282-0003}} &         
	\includegraphics[width=\imwidth, align=c]{\impath{spiralout/samples_onepix_16282-0004}} \\

   	\midrule
    \multicolumn{5}{c}{Alternating} & & \multicolumn{5}{c}{Spiral In}\\
   	\midrule

	\includegraphics[width=\imwidth, align=c]{\impath{alternate/samples_onepix_03371-0000}} &     
	\includegraphics[width=\imwidth, align=c]{\impath{alternate/samples_onepix_03371-0001}} &         
	\includegraphics[width=\imwidth, align=c]{\impath{alternate/samples_onepix_03371-0002}} &         
	\includegraphics[width=\imwidth, align=c]{\impath{alternate/samples_onepix_03371-0003}} &
	\includegraphics[width=\imwidth, align=c]{\impath{alternate/samples_onepix_12800-0000}} &     
  \phantom{x} &
	\includegraphics[width=\imwidth, align=c]{\impath{spiralin/samples_onepix_00618-0000}} &
	\includegraphics[width=\imwidth, align=c]{\impath{spiralin/samples_onepix_00618-0001}} &
	\includegraphics[width=\imwidth, align=c]{\impath{spiralin/samples_onepix_00618-0002}} &         
	\includegraphics[width=\imwidth, align=c]{\impath{spiralin/samples_onepix_00618-0003}} &         
	\includegraphics[width=\imwidth, align=c]{\impath{spiralin/samples_onepix_00618-0004}} \\

	\includegraphics[width=\imwidth, align=c]{\impath{alternate/samples_onepix_03704-0000}} &     
	\includegraphics[width=\imwidth, align=c]{\impath{alternate/samples_onepix_03704-0001}} &         
	\includegraphics[width=\imwidth, align=c]{\impath{alternate/samples_onepix_03704-0002}} &         
	\includegraphics[width=\imwidth, align=c]{\impath{alternate/samples_onepix_03704-0003}} &
	\includegraphics[width=\imwidth, align=c]{\impath{alternate/samples_onepix_12800-0001}} &     
  \phantom{x} &
	\includegraphics[width=\imwidth, align=c]{\impath{spiralin/samples_onepix_09744-0000}} &
	\includegraphics[width=\imwidth, align=c]{\impath{spiralin/samples_onepix_09744-0001}} &
	\includegraphics[width=\imwidth, align=c]{\impath{spiralin/samples_onepix_09744-0002}} &         
	\includegraphics[width=\imwidth, align=c]{\impath{spiralin/samples_onepix_09744-0003}} &         
	\includegraphics[width=\imwidth, align=c]{\impath{spiralin/samples_onepix_09744-0004}} \\

	\includegraphics[width=\imwidth, align=c]{\impath{alternate/samples_onepix_03774-0000}} &     
	\includegraphics[width=\imwidth, align=c]{\impath{alternate/samples_onepix_03774-0001}} &         
	\includegraphics[width=\imwidth, align=c]{\impath{alternate/samples_onepix_03774-0002}} &         
	\includegraphics[width=\imwidth, align=c]{\impath{alternate/samples_onepix_03774-0003}} &
	\includegraphics[width=\imwidth, align=c]{\impath{alternate/samples_onepix_12800-0002}} &     
  \phantom{x} &
	\includegraphics[width=\imwidth, align=c]{\impath{spiralin/samples_onepix_11665-0000}} &
	\includegraphics[width=\imwidth, align=c]{\impath{spiralin/samples_onepix_11665-0001}} &
	\includegraphics[width=\imwidth, align=c]{\impath{spiralin/samples_onepix_11665-0002}} &         
	\includegraphics[width=\imwidth, align=c]{\impath{spiralin/samples_onepix_11665-0003}} &         
	\includegraphics[width=\imwidth, align=c]{\impath{spiralin/samples_onepix_11665-0004}} \\

	\includegraphics[width=\imwidth, align=c]{\impath{alternate/samples_onepix_04642-0000}} &     
	\includegraphics[width=\imwidth, align=c]{\impath{alternate/samples_onepix_04642-0001}} &         
	\includegraphics[width=\imwidth, align=c]{\impath{alternate/samples_onepix_04642-0002}} &         
	\includegraphics[width=\imwidth, align=c]{\impath{alternate/samples_onepix_04642-0003}} &
	\includegraphics[width=\imwidth, align=c]{\impath{alternate/samples_onepix_12800-0003}} &     
  \phantom{x} &
	\includegraphics[width=\imwidth, align=c]{\impath{spiralin/samples_onepix_22726-0000}} &
	\includegraphics[width=\imwidth, align=c]{\impath{spiralin/samples_onepix_22726-0001}} &
	\includegraphics[width=\imwidth, align=c]{\impath{spiralin/samples_onepix_22726-0002}} &         
	\includegraphics[width=\imwidth, align=c]{\impath{spiralin/samples_onepix_22726-0003}} &         
	\includegraphics[width=\imwidth, align=c]{\impath{spiralin/samples_onepix_22726-0004}} \\
	
	\includegraphics[width=\imwidth, align=c]{\impath{alternate/samples_onepix_04700-0000}} &     
	\includegraphics[width=\imwidth, align=c]{\impath{alternate/samples_onepix_04700-0001}} &         
	\includegraphics[width=\imwidth, align=c]{\impath{alternate/samples_onepix_04700-0002}} &         
	\includegraphics[width=\imwidth, align=c]{\impath{alternate/samples_onepix_04700-0003}} &
	\includegraphics[width=\imwidth, align=c]{\impath{alternate/samples_onepix_21390-0000}} &     
  \phantom{x} &
	\includegraphics[width=\imwidth, align=c]{\impath{spiralin/samples_onepix_32097-0000}} &
	\includegraphics[width=\imwidth, align=c]{\impath{spiralin/samples_onepix_32097-0001}} &
	\includegraphics[width=\imwidth, align=c]{\impath{spiralin/samples_onepix_32097-0002}} &         
	\includegraphics[width=\imwidth, align=c]{\impath{spiralin/samples_onepix_32097-0003}} &         
	\includegraphics[width=\imwidth, align=c]{\impath{spiralin/samples_onepix_32097-0004}} \\

    \bottomrule  
  \end{tabular}    
  \caption{Random samples from transformer models trained with different
  orderings for autoregressive prediction as described in
  Sec.~\ref{sec:orderingsupp}.
  }
  \label{fig:permutationsbatches}
\end{figure}
}

\newcommand{\firstpagefigurehalf}{
\begin{figure}[tbp]
  \renewcommand{\impath}[1]{img/##1}
  \renewcommand{\imwidth}{0.12\textwidth}
  \setlength{\tabcolsep}{0pt}
  \centering
  \begin{tabular}{ c  c  c  c}
    \scriptsize{conditioning} & \multicolumn{3}{c}{\scriptsize{samples}} \\
    \toprule

 	\includegraphics[width=\imwidth, align=c]{\impath{completions/witcherin_half}} &     
 	\includegraphics[width=\imwidth, align=c]{\impath{completions/witcherh1}} &         
 	\includegraphics[width=\imwidth, align=c]{\impath{completions/witcherh2}} &         
 	\includegraphics[width=\imwidth, align=c]{\impath{completions/witcherh7}} \\
  \midrule

	\includegraphics[width=\imwidth, align=c]{\impath{depthtorin/conditioning_00443-0001}} &     
	\includegraphics[width=\imwidth, align=c]{\impath{depthtorin/samples_onepix_00443-0000}} &         
	\includegraphics[width=\imwidth, align=c]{\impath{depthtorin/samples_onepix_00443-0002}} &         
	\includegraphics[width=\imwidth, align=c]{\impath{depthtorin/samples_onepix_00443-0003}} \\
	\midrule
		
	\includegraphics[width=\imwidth, align=c]{\impath{ade20k/conditioning_01138-0001}} &     
	\includegraphics[width=\imwidth, align=c]{\impath{ade20k/samples_onepix_01138-0001}} &         
	\includegraphics[width=\imwidth, align=c]{\impath{ade20k/samples_onepix_01138-0002}} &         
	\includegraphics[width=\imwidth, align=c]{\impath{ade20k/samples_onepix_01138-0000}} \\
	\midrule

	\includegraphics[width=\imwidth, align=c]{\impath{deepfashion/condin1511}} &     
	\includegraphics[width=\imwidth, align=c]{\impath{deepfashion/full1511_1}} &         
	\includegraphics[width=\imwidth, align=c]{\impath{deepfashion/full1511_2}} &         
	\includegraphics[width=\imwidth, align=c]{\impath{deepfashion/full1511_3}} \\
	\midrule

	\includegraphics[width=\imwidth, align=c]{\impath{classcondrin/inputs_00062-0001}} &     
	\includegraphics[width=\imwidth, align=c]{\impath{classcondrin/samples_onepix_00062-0005}} &         
	\includegraphics[width=\imwidth, align=c]{\impath{classcondrin/samples_onepix_00062-0003}} &         
	\includegraphics[width=\imwidth, align=c]{\impath{classcondrin/samples_onepix_00062-0001}} \\
    \bottomrule  
  \end{tabular}    
  \caption{Transformers within our setting unify a wide range of image
  synthesis tasks. We show $256\times 256$ synthesis results across different
  conditioning inputs and datasets, all obtained with the same approach to
  exploit inductive biases of effective CNN based \emph{VQGAN} architectures in
  combination with the expressivity of transformer architectures. Top row:
  Completions from unconditional training on ImageNet. 2nd row: Depth-to-Image
  on RIN. 3rd row: Semantically guided synthesis on ADE20K. 4th row: Pose-guided person generation on DeepFashion. Bottom row: Class-conditional samples on RIN.}
  \label{fig:firstpage}
\end{figure}
}

\newcommand{\facesnns}{
\begin{figure}[tbp]
\centering
    \renewcommand{\impath}[1]{img/nns/##1}
    \renewcommand{\imwidth}{0.075\textwidth}
    \setlength{\tabcolsep}{0pt}
    \renewcommand{\arraystretch}{0}
    \centering
    \begin{adjustbox}{max width=0.8 \linewidth, max height=0.7\textheight}
\begin{tabular}{c@{\hskip 0.5em}c@{\hskip 1.0em}cccccccccc}
  \toprule

  & sample & \multicolumn{10}{c}{nearest neighbors} \\
  \midrule

  \multirow{4}{*}{\rotatebox[origin=c]{90}{CelebA-HQ (best val. NLL)}} &
  \includegraphics[width=\imwidth, align=c]{\impath{celebahq_nns_bestval/000045_000000_query}} & 
  \includegraphics[width=\imwidth, align=c]{\impath{celebahq_nns_bestval/000045_000000}} & 
  \includegraphics[width=\imwidth, align=c]{\impath{celebahq_nns_bestval/000045_000001}} & 
  \includegraphics[width=\imwidth, align=c]{\impath{celebahq_nns_bestval/000045_000002}} & 
  \includegraphics[width=\imwidth, align=c]{\impath{celebahq_nns_bestval/000045_000003}} & 
  \includegraphics[width=\imwidth, align=c]{\impath{celebahq_nns_bestval/000045_000004}} & 
  \includegraphics[width=\imwidth, align=c]{\impath{celebahq_nns_bestval/000045_000005}} & 
  \includegraphics[width=\imwidth, align=c]{\impath{celebahq_nns_bestval/000045_000006}} & 
  \includegraphics[width=\imwidth, align=c]{\impath{celebahq_nns_bestval/000045_000007}} & 
  \includegraphics[width=\imwidth, align=c]{\impath{celebahq_nns_bestval/000045_000008}} & 
  \includegraphics[width=\imwidth, align=c]{\impath{celebahq_nns_bestval/000045_000009}} \\

  &
  \includegraphics[width=\imwidth, align=c]{\impath{celebahq_nns_bestval/000021_000000_query}} & 
  \includegraphics[width=\imwidth, align=c]{\impath{celebahq_nns_bestval/000021_000000}} & 
  \includegraphics[width=\imwidth, align=c]{\impath{celebahq_nns_bestval/000021_000001}} & 
  \includegraphics[width=\imwidth, align=c]{\impath{celebahq_nns_bestval/000021_000002}} & 
  \includegraphics[width=\imwidth, align=c]{\impath{celebahq_nns_bestval/000021_000003}} & 
  \includegraphics[width=\imwidth, align=c]{\impath{celebahq_nns_bestval/000021_000004}} & 
  \includegraphics[width=\imwidth, align=c]{\impath{celebahq_nns_bestval/000021_000005}} & 
  \includegraphics[width=\imwidth, align=c]{\impath{celebahq_nns_bestval/000021_000006}} & 
  \includegraphics[width=\imwidth, align=c]{\impath{celebahq_nns_bestval/000021_000007}} & 
  \includegraphics[width=\imwidth, align=c]{\impath{celebahq_nns_bestval/000021_000008}} & 
  \includegraphics[width=\imwidth, align=c]{\impath{celebahq_nns_bestval/000021_000009}} \\

  &
  \includegraphics[width=\imwidth, align=c]{\impath{celebahq_nns_bestval/000025_000000_query}} & 
  \includegraphics[width=\imwidth, align=c]{\impath{celebahq_nns_bestval/000025_000000}} & 
  \includegraphics[width=\imwidth, align=c]{\impath{celebahq_nns_bestval/000025_000001}} & 
  \includegraphics[width=\imwidth, align=c]{\impath{celebahq_nns_bestval/000025_000002}} & 
  \includegraphics[width=\imwidth, align=c]{\impath{celebahq_nns_bestval/000025_000003}} & 
  \includegraphics[width=\imwidth, align=c]{\impath{celebahq_nns_bestval/000025_000004}} & 
  \includegraphics[width=\imwidth, align=c]{\impath{celebahq_nns_bestval/000025_000005}} & 
  \includegraphics[width=\imwidth, align=c]{\impath{celebahq_nns_bestval/000025_000006}} & 
  \includegraphics[width=\imwidth, align=c]{\impath{celebahq_nns_bestval/000025_000007}} & 
  \includegraphics[width=\imwidth, align=c]{\impath{celebahq_nns_bestval/000025_000008}} & 
  \includegraphics[width=\imwidth, align=c]{\impath{celebahq_nns_bestval/000025_000009}} \\

  &
  \includegraphics[width=\imwidth, align=c]{\impath{celebahq_nns_bestval/000035_000000_query}} & 
  \includegraphics[width=\imwidth, align=c]{\impath{celebahq_nns_bestval/000035_000000}} & 
  \includegraphics[width=\imwidth, align=c]{\impath{celebahq_nns_bestval/000035_000001}} & 
  \includegraphics[width=\imwidth, align=c]{\impath{celebahq_nns_bestval/000035_000002}} & 
  \includegraphics[width=\imwidth, align=c]{\impath{celebahq_nns_bestval/000035_000003}} & 
  \includegraphics[width=\imwidth, align=c]{\impath{celebahq_nns_bestval/000035_000004}} & 
  \includegraphics[width=\imwidth, align=c]{\impath{celebahq_nns_bestval/000035_000005}} & 
  \includegraphics[width=\imwidth, align=c]{\impath{celebahq_nns_bestval/000035_000006}} & 
  \includegraphics[width=\imwidth, align=c]{\impath{celebahq_nns_bestval/000035_000007}} & 
  \includegraphics[width=\imwidth, align=c]{\impath{celebahq_nns_bestval/000035_000008}} & 
  \includegraphics[width=\imwidth, align=c]{\impath{celebahq_nns_bestval/000035_000009}} \\

  \midrule

  \multirow{4}{*}{\rotatebox[origin=c]{90}{CelebA-HQ (best train NLL)}} &
  \includegraphics[width=\imwidth, align=c]{\impath{celebahq_nns/000000_000000_query}} & 
  \includegraphics[width=\imwidth, align=c]{\impath{celebahq_nns/000000_000000}} & 
  \includegraphics[width=\imwidth, align=c]{\impath{celebahq_nns/000000_000001}} & 
  \includegraphics[width=\imwidth, align=c]{\impath{celebahq_nns/000000_000002}} & 
  \includegraphics[width=\imwidth, align=c]{\impath{celebahq_nns/000000_000003}} & 
  \includegraphics[width=\imwidth, align=c]{\impath{celebahq_nns/000000_000004}} & 
  \includegraphics[width=\imwidth, align=c]{\impath{celebahq_nns/000000_000005}} & 
  \includegraphics[width=\imwidth, align=c]{\impath{celebahq_nns/000000_000006}} & 
  \includegraphics[width=\imwidth, align=c]{\impath{celebahq_nns/000000_000007}} & 
  \includegraphics[width=\imwidth, align=c]{\impath{celebahq_nns/000000_000008}} & 
  \includegraphics[width=\imwidth, align=c]{\impath{celebahq_nns/000000_000009}} \\

  &
  \includegraphics[width=\imwidth, align=c]{\impath{celebahq_nns/000001_000000_query}} & 
  \includegraphics[width=\imwidth, align=c]{\impath{celebahq_nns/000001_000000}} & 
  \includegraphics[width=\imwidth, align=c]{\impath{celebahq_nns/000001_000001}} & 
  \includegraphics[width=\imwidth, align=c]{\impath{celebahq_nns/000001_000002}} & 
  \includegraphics[width=\imwidth, align=c]{\impath{celebahq_nns/000001_000003}} & 
  \includegraphics[width=\imwidth, align=c]{\impath{celebahq_nns/000001_000004}} & 
  \includegraphics[width=\imwidth, align=c]{\impath{celebahq_nns/000001_000005}} & 
  \includegraphics[width=\imwidth, align=c]{\impath{celebahq_nns/000001_000006}} & 
  \includegraphics[width=\imwidth, align=c]{\impath{celebahq_nns/000001_000007}} & 
  \includegraphics[width=\imwidth, align=c]{\impath{celebahq_nns/000001_000008}} & 
  \includegraphics[width=\imwidth, align=c]{\impath{celebahq_nns/000001_000009}} \\

  &
  \includegraphics[width=\imwidth, align=c]{\impath{celebahq_nns/000002_000000_query}} & 
  \includegraphics[width=\imwidth, align=c]{\impath{celebahq_nns/000002_000000}} & 
  \includegraphics[width=\imwidth, align=c]{\impath{celebahq_nns/000002_000001}} & 
  \includegraphics[width=\imwidth, align=c]{\impath{celebahq_nns/000002_000002}} & 
  \includegraphics[width=\imwidth, align=c]{\impath{celebahq_nns/000002_000003}} & 
  \includegraphics[width=\imwidth, align=c]{\impath{celebahq_nns/000002_000004}} & 
  \includegraphics[width=\imwidth, align=c]{\impath{celebahq_nns/000002_000005}} & 
  \includegraphics[width=\imwidth, align=c]{\impath{celebahq_nns/000002_000006}} & 
  \includegraphics[width=\imwidth, align=c]{\impath{celebahq_nns/000002_000007}} & 
  \includegraphics[width=\imwidth, align=c]{\impath{celebahq_nns/000002_000008}} & 
  \includegraphics[width=\imwidth, align=c]{\impath{celebahq_nns/000002_000009}} \\

  &
  \includegraphics[width=\imwidth, align=c]{\impath{celebahq_nns/000003_000000_query}} & 
  \includegraphics[width=\imwidth, align=c]{\impath{celebahq_nns/000003_000000}} & 
  \includegraphics[width=\imwidth, align=c]{\impath{celebahq_nns/000003_000001}} & 
  \includegraphics[width=\imwidth, align=c]{\impath{celebahq_nns/000003_000002}} & 
  \includegraphics[width=\imwidth, align=c]{\impath{celebahq_nns/000003_000003}} & 
  \includegraphics[width=\imwidth, align=c]{\impath{celebahq_nns/000003_000004}} & 
  \includegraphics[width=\imwidth, align=c]{\impath{celebahq_nns/000003_000005}} & 
  \includegraphics[width=\imwidth, align=c]{\impath{celebahq_nns/000003_000006}} & 
  \includegraphics[width=\imwidth, align=c]{\impath{celebahq_nns/000003_000007}} & 
  \includegraphics[width=\imwidth, align=c]{\impath{celebahq_nns/000003_000008}} & 
  \includegraphics[width=\imwidth, align=c]{\impath{celebahq_nns/000003_000009}} \\

  \midrule

  \multirow{4}{*}{\rotatebox[origin=c]{90}{FFHQ (best val. NLL)}} &
  \includegraphics[width=\imwidth, align=c]{\impath{ffhq_nns_bestval/000013_000000_query}} & 
  \includegraphics[width=\imwidth, align=c]{\impath{ffhq_nns_bestval/000013_000000}} & 
  \includegraphics[width=\imwidth, align=c]{\impath{ffhq_nns_bestval/000013_000001}} & 
  \includegraphics[width=\imwidth, align=c]{\impath{ffhq_nns_bestval/000013_000002}} & 
  \includegraphics[width=\imwidth, align=c]{\impath{ffhq_nns_bestval/000013_000003}} & 
  \includegraphics[width=\imwidth, align=c]{\impath{ffhq_nns_bestval/000013_000004}} & 
  \includegraphics[width=\imwidth, align=c]{\impath{ffhq_nns_bestval/000013_000005}} & 
  \includegraphics[width=\imwidth, align=c]{\impath{ffhq_nns_bestval/000013_000006}} & 
  \includegraphics[width=\imwidth, align=c]{\impath{ffhq_nns_bestval/000013_000007}} & 
  \includegraphics[width=\imwidth, align=c]{\impath{ffhq_nns_bestval/000013_000008}} & 
  \includegraphics[width=\imwidth, align=c]{\impath{ffhq_nns_bestval/000013_000009}} \\

  &
  \includegraphics[width=\imwidth, align=c]{\impath{ffhq_nns_bestval/000009_000000_query}} & 
  \includegraphics[width=\imwidth, align=c]{\impath{ffhq_nns_bestval/000009_000000}} & 
  \includegraphics[width=\imwidth, align=c]{\impath{ffhq_nns_bestval/000009_000001}} & 
  \includegraphics[width=\imwidth, align=c]{\impath{ffhq_nns_bestval/000009_000002}} & 
  \includegraphics[width=\imwidth, align=c]{\impath{ffhq_nns_bestval/000009_000003}} & 
  \includegraphics[width=\imwidth, align=c]{\impath{ffhq_nns_bestval/000009_000004}} & 
  \includegraphics[width=\imwidth, align=c]{\impath{ffhq_nns_bestval/000009_000005}} & 
  \includegraphics[width=\imwidth, align=c]{\impath{ffhq_nns_bestval/000009_000006}} & 
  \includegraphics[width=\imwidth, align=c]{\impath{ffhq_nns_bestval/000009_000007}} & 
  \includegraphics[width=\imwidth, align=c]{\impath{ffhq_nns_bestval/000009_000008}} & 
  \includegraphics[width=\imwidth, align=c]{\impath{ffhq_nns_bestval/000009_000009}} \\

  &
  \includegraphics[width=\imwidth, align=c]{\impath{ffhq_nns_bestval/000007_000000_query}} & 
  \includegraphics[width=\imwidth, align=c]{\impath{ffhq_nns_bestval/000007_000000}} & 
  \includegraphics[width=\imwidth, align=c]{\impath{ffhq_nns_bestval/000007_000001}} & 
  \includegraphics[width=\imwidth, align=c]{\impath{ffhq_nns_bestval/000007_000002}} & 
  \includegraphics[width=\imwidth, align=c]{\impath{ffhq_nns_bestval/000007_000003}} & 
  \includegraphics[width=\imwidth, align=c]{\impath{ffhq_nns_bestval/000007_000004}} & 
  \includegraphics[width=\imwidth, align=c]{\impath{ffhq_nns_bestval/000007_000005}} & 
  \includegraphics[width=\imwidth, align=c]{\impath{ffhq_nns_bestval/000007_000006}} & 
  \includegraphics[width=\imwidth, align=c]{\impath{ffhq_nns_bestval/000007_000007}} & 
  \includegraphics[width=\imwidth, align=c]{\impath{ffhq_nns_bestval/000007_000008}} & 
  \includegraphics[width=\imwidth, align=c]{\impath{ffhq_nns_bestval/000007_000009}} \\

  &
  \includegraphics[width=\imwidth, align=c]{\impath{ffhq_nns_bestval/000000_000000_query}} & 
  \includegraphics[width=\imwidth, align=c]{\impath{ffhq_nns_bestval/000000_000000}} & 
  \includegraphics[width=\imwidth, align=c]{\impath{ffhq_nns_bestval/000000_000001}} & 
  \includegraphics[width=\imwidth, align=c]{\impath{ffhq_nns_bestval/000000_000002}} & 
  \includegraphics[width=\imwidth, align=c]{\impath{ffhq_nns_bestval/000000_000003}} & 
  \includegraphics[width=\imwidth, align=c]{\impath{ffhq_nns_bestval/000000_000004}} & 
  \includegraphics[width=\imwidth, align=c]{\impath{ffhq_nns_bestval/000000_000005}} & 
  \includegraphics[width=\imwidth, align=c]{\impath{ffhq_nns_bestval/000000_000006}} & 
  \includegraphics[width=\imwidth, align=c]{\impath{ffhq_nns_bestval/000000_000007}} & 
  \includegraphics[width=\imwidth, align=c]{\impath{ffhq_nns_bestval/000000_000008}} & 
  \includegraphics[width=\imwidth, align=c]{\impath{ffhq_nns_bestval/000000_000009}} \\

  \midrule

  \multirow{4}{*}{\rotatebox[origin=c]{90}{FFHQ (best train NLL)}} &
  \includegraphics[width=\imwidth, align=c]{\impath{ffhq_nns/000000_000000_query}} & 
  \includegraphics[width=\imwidth, align=c]{\impath{ffhq_nns/000000_000000}} & 
  \includegraphics[width=\imwidth, align=c]{\impath{ffhq_nns/000000_000001}} & 
  \includegraphics[width=\imwidth, align=c]{\impath{ffhq_nns/000000_000002}} & 
  \includegraphics[width=\imwidth, align=c]{\impath{ffhq_nns/000000_000003}} & 
  \includegraphics[width=\imwidth, align=c]{\impath{ffhq_nns/000000_000004}} & 
  \includegraphics[width=\imwidth, align=c]{\impath{ffhq_nns/000000_000005}} & 
  \includegraphics[width=\imwidth, align=c]{\impath{ffhq_nns/000000_000006}} & 
  \includegraphics[width=\imwidth, align=c]{\impath{ffhq_nns/000000_000007}} & 
  \includegraphics[width=\imwidth, align=c]{\impath{ffhq_nns/000000_000008}} & 
  \includegraphics[width=\imwidth, align=c]{\impath{ffhq_nns/000000_000009}} \\

  &
  \includegraphics[width=\imwidth, align=c]{\impath{ffhq_nns/000001_000000_query}} & 
  \includegraphics[width=\imwidth, align=c]{\impath{ffhq_nns/000001_000000}} & 
  \includegraphics[width=\imwidth, align=c]{\impath{ffhq_nns/000001_000001}} & 
  \includegraphics[width=\imwidth, align=c]{\impath{ffhq_nns/000001_000002}} & 
  \includegraphics[width=\imwidth, align=c]{\impath{ffhq_nns/000001_000003}} & 
  \includegraphics[width=\imwidth, align=c]{\impath{ffhq_nns/000001_000004}} & 
  \includegraphics[width=\imwidth, align=c]{\impath{ffhq_nns/000001_000005}} & 
  \includegraphics[width=\imwidth, align=c]{\impath{ffhq_nns/000001_000006}} & 
  \includegraphics[width=\imwidth, align=c]{\impath{ffhq_nns/000001_000007}} & 
  \includegraphics[width=\imwidth, align=c]{\impath{ffhq_nns/000001_000008}} & 
  \includegraphics[width=\imwidth, align=c]{\impath{ffhq_nns/000001_000009}} \\

  &
  \includegraphics[width=\imwidth, align=c]{\impath{ffhq_nns/000002_000000_query}} & 
  \includegraphics[width=\imwidth, align=c]{\impath{ffhq_nns/000002_000000}} & 
  \includegraphics[width=\imwidth, align=c]{\impath{ffhq_nns/000002_000001}} & 
  \includegraphics[width=\imwidth, align=c]{\impath{ffhq_nns/000002_000002}} & 
  \includegraphics[width=\imwidth, align=c]{\impath{ffhq_nns/000002_000003}} & 
  \includegraphics[width=\imwidth, align=c]{\impath{ffhq_nns/000002_000004}} & 
  \includegraphics[width=\imwidth, align=c]{\impath{ffhq_nns/000002_000005}} & 
  \includegraphics[width=\imwidth, align=c]{\impath{ffhq_nns/000002_000006}} & 
  \includegraphics[width=\imwidth, align=c]{\impath{ffhq_nns/000002_000007}} & 
  \includegraphics[width=\imwidth, align=c]{\impath{ffhq_nns/000002_000008}} & 
  \includegraphics[width=\imwidth, align=c]{\impath{ffhq_nns/000002_000009}} \\

  &
  \includegraphics[width=\imwidth, align=c]{\impath{ffhq_nns/000003_000000_query}} & 
  \includegraphics[width=\imwidth, align=c]{\impath{ffhq_nns/000003_000000}} & 
  \includegraphics[width=\imwidth, align=c]{\impath{ffhq_nns/000003_000001}} & 
  \includegraphics[width=\imwidth, align=c]{\impath{ffhq_nns/000003_000002}} & 
  \includegraphics[width=\imwidth, align=c]{\impath{ffhq_nns/000003_000003}} & 
  \includegraphics[width=\imwidth, align=c]{\impath{ffhq_nns/000003_000004}} & 
  \includegraphics[width=\imwidth, align=c]{\impath{ffhq_nns/000003_000005}} & 
  \includegraphics[width=\imwidth, align=c]{\impath{ffhq_nns/000003_000006}} & 
  \includegraphics[width=\imwidth, align=c]{\impath{ffhq_nns/000003_000007}} & 
  \includegraphics[width=\imwidth, align=c]{\impath{ffhq_nns/000003_000008}} & 
  \includegraphics[width=\imwidth, align=c]{\impath{ffhq_nns/000003_000009}} \\

  \bottomrule

\end{tabular}
\end{adjustbox}
  \caption{\label{fig:facesnns} Nearest neighbors for our face-models trained on FFHQ and CelebA-HQ ($256 \times 256$ pix), based on the LPIPS \cite{zhang2018unreasonable} distance. The left column shows a sample from our model, while the 10 examples to the right show the nearest neighbors from the corresponding class (increasing distance) in the training dataset. We evaluate two different model checkpoints for each dataset: Best val. NLL denotes the minimal NLL
  over the course of training, evaluated on unseen testdata. For this checkpoint, both models generate crisp, high-quality samples not present in the training data. However, when drastically overfitting the model, it reproduces
  samples from the training data (best train NLL). Although not an ideal measure of image quality, NLL thus provides a proxy on model selection, whereas FID does not. See also Sec.~\ref{supp:nns}.}
\end{figure}
}

\newcommand{\facesfidandinceptionplots}{
\begin{figure*}
\centering
\includegraphics[width=0.4\textwidth]{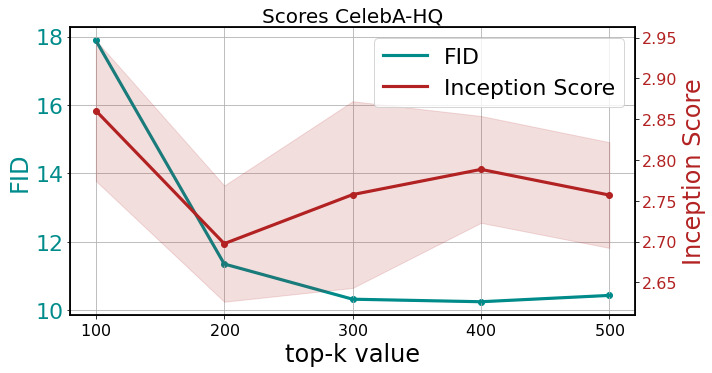}
\includegraphics[width=0.4\textwidth]{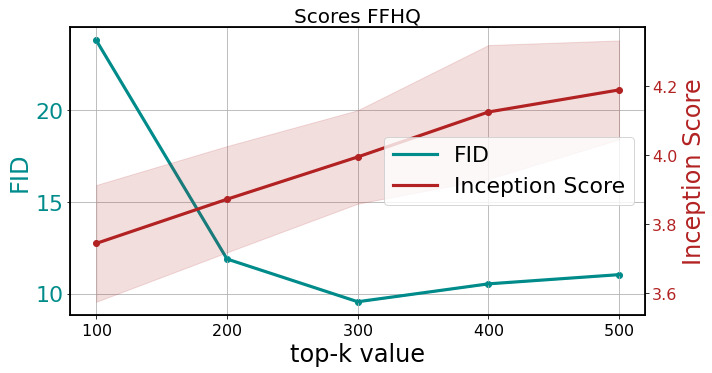}
  \caption{\label{fig:facesfidandinceptionplots} FID and Inception Score as a function of top-k for CelebA-HQ (left) and FFHQ (right).}
\end{figure*}
}

%% file: cvpr_tables.tex
\newcommand{\imagenetreconstruction}{
\begin{table}[t]
\begin{center}
\begin{footnotesize}
\begin{tabular}{ccccc}
  Model & Codebook Size & $\dim \mathcal{Z}$ & FID/val & FID/train \\ 
\toprule
  VQVAE-2 & $64 \times 64$ \& $32 \times 32$ & $512$ & n/a & $\sim 10$\\
  DALL-E \cite{ramesh2021zeroshot} & $32 \times 32$ & $8192$ & 32.01 & 33.88 \\  %
  \emph{VQGAN} & $16 \times 16$ & $1024$ & 7.94 & 10.54 \\
  \emph{VQGAN} & $16 \times 16$ & $16384$ & 4.98 & 7.41\\
  \emph{VQGAN$^*$} & $32 \times 32$ & $8192$ & 1.49 & 3.24\\
  \emph{VQGAN} & $64 \times 64$ \& $32 \times 32$ & $512$ & 1.45 & 2.78 \\
\bottomrule
\end{tabular}
\end{footnotesize}
\caption{\label{tab:recfidimagenet}
  FID on ImageNet between reconstructed validation split and
  original validation (FID/val) and training (FID/train) splits.
  $^*$trained with Gumbel-Softmax reparameterization as in \cite{ramesh2021zeroshot, 
  jang2016categorical}.
  }
\end{center}
\vspace{-2.0em}
\end{table}
}

\newcommand{\semanticsynthesis}{
\begin{table}[t]
\begin{footnotesize}
  \begin{adjustbox}{max width=\linewidth}
\begin{tabular}{ccccc}
  Dataset & \textbf{ours} & SPADE \cite{park2019semantic} & Pix2PixHD (+aug) \cite{wang2018pix2pixHD} & CRN \cite{ChenK17} \\
\toprule
  COCO-Stuff & 22.4 & 22.6/23.9(*) & 111.5 (54.2) & 70.4 \\
  ADE20K & 35.5 & 33.9/35.7(*) & 81.8 (41.5) & 73.3 \\
\bottomrule
\end{tabular}
  \end{adjustbox}
\end{footnotesize}
\caption{\label{tab:semanticsynth}FID score comparison for semantic image
  synthesis ($256 \times 256$ pixels). (*):~Recalculated with our evaluation protocol based on \cite{obukhov2020torchfidelity} on the validation splits of each dataset.
    \vspace{-1.5em}}
\end{table}
}

\newcommand{\cinsynthesis}{
\begin{table}[t]
  \centering
\begin{footnotesize}
  \begin{adjustbox}{max width=\linewidth}
\begin{tabular}{ccccc}
  Dataset & ours-previous (+R) & BigGAN (-deep) & MSP \\
\toprule
  IN $256$, 50K & 19.8 (11.2) & 7.1 (7.3) & n.a. \\
  IN $256$, 18K & 23.5& 9.6 (9.7) & 50.4 \\
\bottomrule
\end{tabular}
    \hspace{2em}
\begin{tabular}{ccc}
  Dataset & ours-previous & ours-new \\
\toprule
  CelebA-HQ $256$ & 10.7 & 10.2 \\
  FFHQ $256$ & 11.4 & 9.6 \\
\bottomrule
\end{tabular}
  \end{adjustbox}
\end{footnotesize}
  \caption{\label{tab:prevresults} Results from a previous version of this
  paper, see also Sec.~\ref{sec:changelog}. Left: Previous results on
  class-conditional ImageNet synthesis with a slightly different implementation
  and evaluated against 50k and 18k training examples instead of the whole
  training split. See Tab.~\ref{tab:cinsynthesisupdate} for new, improved
  results evaluated against the whole training split.
  Right: Previous results on face-synthesis with a slightly different
  implementation compared to the new implementation. See also
  Tab.~\ref{tab:facesynth} for comparison with other methods.
  }
\end{table}
}

\newcommand{\cinsynthesisupdate}{
\begin{table}[t]
\begin{footnotesize}
  \begin{adjustbox}{max width=\linewidth}
\begin{tabular}{c ccc}
  Model & acceptance rate & FID & IS \\
\toprule
mixed $k$, $p=1.0$ & 1.0 & 17.04 & $70.6 \pm 1.8$ \\
$k=973$, $p=1.0$ & 1.0 & 29.20 & $47.3 \pm 1.3$ \\
$k=250$, $p=1.0$ & 1.0 & 15.98 & $78.6 \pm 1.1$ \\
$k=973$, $p=0.88$ & 1.0 & 15.78 & $74.3 \pm 1.8$ \\
$k=600$, $p=1.0$ & 0.05 & 5.20 & $280.3 \pm 5.5$\\ 
\midrule
mixed $k$, $p=1.0$ & 0.5 & 10.26 & $125.5 \pm 2.4$ \\
mixed $k$, $p=1.0$ & 0.25 & 7.35 & $188.6 \pm 3.3$ \\
mixed $k$, $p=1.0$ & 0.05 & 5.88 & $304.8 \pm 3.6$ \\
mixed $k$, $p=1.0$ & 0.005 & 6.59 & $402.7 \pm 2.9$ \\
\midrule
DCTransformer \cite{nash2021generating} & 1.0 & 36.5 & n/a \\
VQVAE-2 \cite{razavi2019generating} & 1.0 & $\sim$31 & $\sim$45 \\
VQVAE-2 & n/a & $\sim$10 & $\sim$330 \\
BigGAN \cite{brock2018large} & 1.0 & 7.53 & $168.6 \pm 2.5$ \\
BigGAN-deep & 1.0 & 6.84 & $203.6 \pm 2.6$ \\
IDDPM \cite{nichol2021improved} & 1.0 & 12.3 & n/a \\
ADM-G, no guid. \cite{dhariwal2021diffusion} & 1.0 & 10.94 & 100.98 \\
ADM-G, 1.0 guid. & 1.0 & 4.59 & 186.7 \\
ADM-G, 10.0 guid. & 1.0 & 9.11 & 283.92 \\
\midrule
val. data & 1.0 & 1.62 & $234.0 \pm 3.9$ \\
\bottomrule
\end{tabular}
  \end{adjustbox}
\end{footnotesize}
\caption{\label{tab:cinsynthesisupdate}
  FID score comparison for class-conditional
  synthesis on $256\times 256$ ImageNet, evaluated between 50k samples and the 
  training split. Classifier-based rejection sampling as in
  VQVAE-2 uses a ResNet-101 \cite{DBLP:journals/corr/HeZRS15} classifier. 
  BigGAN(-deep) evaluated via
  \url{https://tfhub.dev/deepmind} truncated at 1.0. ``Mixed'' $k$ refers to 
  samples generated with different top-k values, here $k \in \{100, 200, 250, 300, 350, 400, 500, 600, 800, 973\}$.
  \vspace{-2em}}
\end{table}
}

\newcommand{\celebahq}{
\begin{table}[t]
\begin{footnotesize}
  \begin{adjustbox}{max width=\linewidth}
  \begin{tabular}{ccc@{\hskip 0.25em}cc}
    \toprule
    \multicolumn{2}{c}{CelebA-HQ $256\times 256$} & & \multicolumn{2}{c}{FFHQ
    $256\times 256$} \\
    \cmidrule{1-2} \cmidrule{4-5}
    Method & FID $\downarrow$ & & Method & FID $\downarrow$ \\
    \cmidrule{1-2} \cmidrule{4-5}
    GLOW \cite{kingma2018glow} & 69.0 & & VDVAE ($t=0.7$) \cite{child2020vdvae} & 38.8 \\
    NVAE \cite{VahdatK20} & 40.3 & & VDVAE ($t=1.0$) & 33.5 \\
    PIONEER (B.) \cite{heljakka2018pioneer} & 39.2 (25.3) & & VDVAE ($t=0.8$) & 29.8 \\
    NCPVAE \cite{aneja2020ncpvae} & 24.8 & & VDVAE ($t=0.9$) & 28.5 \\
    VAEBM \cite{xiao2021vaebm} & 20.4 & & \emph{VQGAN}+P.SNAIL & 21.9 \\
    Style ALAE \cite{PidhorskyiAD20} & 19.2 & & BigGAN & 12.4 \\
    DC-VAE \cite{parmar2020dual} & 15.8 & & \textbf{ours} (k=300) & 9.6 \\
    \textbf{ours} (k=400) & 10.2 & & U-Net GAN (+aug) \cite{SchonfeldSK20} & 10.9 (7.6) \\
    PGGAN \cite{karras2018progressive} & 8.0 & & StyleGAN2 (+aug) \cite{karras2020analyzing} & 3.8 (3.6) \\

\bottomrule
\end{tabular}
  \end{adjustbox}
\end{footnotesize}
  \caption{\label{tab:facesynth}FID score comparison for face image
  synthesis. CelebA-HQ results reproduced from
  \cite{aneja2020ncpvae,parmar2020dual,xiao2021vaebm,heljakka2020towards}, FFHQ
  from \cite{SchonfeldSK20,karras2020training}.
    \vspace{-1.0em}}
\end{table}
}

\newcommand{\pixelsnailvstransformersmall}{
\begin{table}[tbp]
\centering
\begin{tabular}{c ccc}
 & \multicolumn{3}{c}{Negative Log-Likelihood (NLL)} \\
\toprule
\shortstack{\footnotesize{Data} / \\ \footnotesize{\# params}} & \shortstack{\footnotesize{Transformer}\\ \footnotesize{\emph{P-SNAIL steps}}} & \shortstack{\footnotesize{Transformer} \\ \footnotesize{\emph{P-SNAIL time}}} & \shortstack{\footnotesize{PixelSNAIL} \\ \footnotesize{\emph{fixed time}}} \\
\midrule
\footnotesize{RIN / 85M} & \textbf{4.78} & 4.84 & 4.96 \\   %
\footnotesize{LSUN-CT / 310M} & \textbf{4.63} & 4.69 & 4.89 \\   %
\footnotesize{IN / 310M} & \textbf{4.78} & 4.83 & 4.96 \\   %
\midrule
\footnotesize{D-RIN / 180 M} & \textbf{4.70} & 4.78 & 4.88 \\  %
\footnotesize{S-FLCKR / 310 M} & \textbf{4.49} & 4.57 & 4.64 \\   %
\bottomrule
\end{tabular}
\caption{Comparing Transformer and PixelSNAIL architectures across different datasets and model sizes. For all settings,
  transformers outperform the state-of-the-art model from the PixelCNN family,
  PixelSNAIL in terms of NLL. This holds both when comparing NLL at fixed times
  (PixelSNAIL trains roughly 2 times faster) and when trained for a fixed
  number of steps. See Sec.~\ref{sec:psnailvstrafo} for the abbreviations.
}
\label{tab:gptvspixelsnail}
\end{table}
}

\newcommand{\vqarchitecture}{
\begin{table*}[tbp]
\centering
\begin{tabular}{c|c}
Encoder & Decoder \\
\toprule
$ x \in \mathbb{R}^{H \times W \times C}$  & $\quantizedcode \in \mathbb{R}^{h \times w \times n_z}$ \\
Conv2D $\to \mathbb{R}^{H \times W \times C'} $ & Conv2D $\to \mathbb{R}^{h \times w \times C''} $ \\
$m \times $ $\{$ Residual Block, Downsample Block$\}$ $\to \mathbb{R}^{h \times w \times C''} $ & Residual Block $\to \mathbb{R}^{h \times w \times C''} $ \\
Residual Block $\to \mathbb{R}^{h \times w \times C''} $ & Non-Local Block $\to \mathbb{R}^{h \times w \times C''} $ \\
Non-Local Block $\to \mathbb{R}^{h \times w \times C''} $ & Residual Block $\to \mathbb{R}^{h \times w \times C''} $ \\
Residual Block  $\to \mathbb{R}^{h \times w \times C''} $ & $m \times $ $\{$ Residual Block, Upsample Block$\}$ $\to \mathbb{R}^{H \times W \times C'} $ \\
GroupNorm, Swish, Conv2D $\to \mathbb{R}^{h \times w \times n_z}$  &  GroupNorm, Swish, Conv2D $\to \mathbb{R}^{H \times W \times C}$\\
\bottomrule
\end{tabular}
\caption{High-level architecture of the encoder and decoder of our
  \emph{VQGAN}. The design of the networks follows the architecture presented
  in \cite{ho2020denoising} with no skip-connections. For the discriminator, we
  use a patch-based model as in \cite{isola2017image}. Note that $h =
  \frac{H}{2^m}$, $w = \frac{W}{2^m}$ and $f=2^m$.}
\label{tab:vqaearchitecture}
\end{table*}
}

\newcommand{\hyperparameters}{
\begin{table*}[tbp]
\centering
\begin{tabular}{c   c c c  c c c c c c}
  Experiment & $n_{layer}$ & \# params $[M]$ & $n_z$ & $\vert \codebook \vert$ & dropout & $\text{length}(s)$ & $n_e$ & $m$  \\
\toprule
  RIN & 12 & 85 & 64 & 768 & 0.0 & 512 & 1024 & 4  \\    %
  c-RIN & 18 & 128 & 64 & 768 & 0.0 & 257 & 768 & 4  \\  %
  D-RINv1 & 14 & 180 & 256 & 1024 & 0.0 & 512 & 768 & 4  \\   %
  D-RINv2 & 24 & 307 & 256 & 1024 & 0.0 & 512 & 1024 & 4\\
  IN & 24 & 307 & 256 & 1024 & 0.0 & 256 & 1024 & 4  \\
  c-IN & 24 & 307 & 256 & 1024 & 0.0 & 257 & 1024 & 4  \\
  c-IN (big) & 48 & 1400 & 256 & 16384 & 0.0 & 257 & 1536 & 4  \\
  IN-Edges & 24 & 307 & 256 & 1024 & 0.0 & 512 & 1024 & 3  \\
  IN-SR & 12 & 153 & 256 & 1024 & 0.0 & 512 & 1024 & 3  \\  %
  S-FLCKR, $f=4$ & 24 & 307 & 256 & 1024 & 0.0 & 512 & 1024 & 2 \\
  S-FLCKR, $f=16$ & 24 & 307 & 256 & 1024 & 0.0 & 512 & 1024 & 4 \\
  S-FLCKR, $f=32$ & 24 & 307 & 256 & 1024 & 0.0 & 512 & 1024 & 5 \\
  $( \text{FacesHQ}, \: f=1)^*$ & 24 & 307 & -- & 512 & 0.0 & 512 & 1024 & --  \\
  FacesHQ, $f=2$ & 24 & 307 & 256 & 1024 & 0.0 & 512 & 1024 & 1  \\
  FacesHQ, $f=4$ & 24 & 307 & 256 & 1024 & 0.0 & 512 & 1024 & 2 \\
  FacesHQ, $f=8$ & 24 & 307 & 256 & 1024 & 0.0 & 512 & 1024 & 3 \\
  $\text{FacesHQ}^{**}$, $f=16$ & 24 & 307 & 256 & 1024 & 0.0 & 512 & 1024 & 4  \\
  $\text{FFHQ}^{**}$, $f=16$ & 28 & 355 & 256 & 1024 & 0.0 & 256 & 1024 & 4  \\
  $\text{CelebA-HQ}^{**}$, $f=16$ & 28 & 355 & 256 & 1024 & 0.0 & 256 & 1024 & 4  \\
  FFHQ (big) & 24 & 801 & 256 & 1024 & 0.0 & 256 & 1664 & 4  \\
  CelebA-HQ (big) & 24 & 801 & 256 & 1024 & 0.0 & 256 & 1664 & 4  \\
  COCO-Stuff & 32 & 651 & 256 & 8192 & 0.0 & 512 & 1280 & 4  \\
  ADE20K & 28 & 405 & 256 & 4096 & 0.1 & 512 & 1024 & 4  \\
  DeepFashion & 18 & 129 & 256 & 1024 & 0.0 & 340 & 768 & 4  \\  %
  LSUN-CT & 24 & 307 & 256 & 1024 & 0.0 & 256 & 1024 & 4  \\
  CIFAR-10 & 24 & 307 & 256 & 1024 & 0.0 & 256 & 1024 & 1  \\  %
\bottomrule
\end{tabular}
\caption{Hyperparameters. For every experiment, we set the number of attention
  heads in the transformer to $n_h = 16$. $n_{layer}$ denotes the
  number of transformer blocks, \# params the number of transformer parameters,
  $n_z$ the dimensionality of codebook entries, $\vert \codebook \vert$
  the number of codebook entries, dropout the dropout rate for training the
  transformer, $\text{length}(s)$ the total length of the sequence,
  $n_e$ the embedding dimensionality and $m$ the number of downsampling
  steps in the \emph{VQGAN}.
  D-RINv1 is the experiment which compares to Pixel-SNAIL in Sec.~\ref{sec:psnailvstrafo}. Note that the experiment $( \text{FacesHQ}, \: f=1)^*$ does not use a learned \emph{VQGAN} but a fixed k-means clustering algorithm as in \cite{chen2020generative} with $K=512$ centroids. A prefix ``c'' refers to a class-conditional model. The models marked with a `$**$` are trained on the same \emph{VQGAN}.
  }
\label{tab:hyperparameters}
\end{table*}
}

%% file: ms.bbl
\begin{thebibliography}{10}\itemsep=-1pt

\bibitem{aneja2020ncpvae}
Jyoti Aneja, Alexander~G. Schwing, Jan Kautz, and Arash Vahdat.
\newblock {NCP-VAE:} variational autoencoders with noise contrastive priors.
\newblock {\em CoRR}, abs/2010.02917, 2020.

\bibitem{bahdanau2016neural}
Dzmitry Bahdanau, Kyunghyun Cho, and Yoshua Bengio.
\newblock Neural machine translation by jointly learning to align and
  translate, 2016.

\bibitem{bengio2013estimating}
Yoshua Bengio, Nicholas L{\'{e}}onard, and Aaron~C. Courville.
\newblock Estimating or propagating gradients through stochastic neurons for
  conditional computation.
\newblock {\em CoRR}, abs/1308.3432, 2013.

\bibitem{brock2018large}
Andrew Brock, Jeff Donahue, and Karen Simonyan.
\newblock {Large Scale GAN Training for High Fidelity Natural Image Synthesis}.
\newblock In {\em 7th International Conference on Learning Representations,
  {ICLR}}, 2019.

\bibitem{brown2020language}
Tom~B. Brown, Benjamin Mann, Nick Ryder, Melanie Subbiah, Jared Kaplan,
  Prafulla Dhariwal, Arvind Neelakantan, Pranav Shyam, Girish Sastry, Amanda
  Askell, Sandhini Agarwal, Ariel Herbert{-}Voss, Gretchen Krueger, Tom
  Henighan, Rewon Child, Aditya Ramesh, Daniel~M. Ziegler, Jeffrey Wu, Clemens
  Winter, Christopher Hesse, Mark Chen, Eric Sigler, Mateusz Litwin, Scott
  Gray, Benjamin Chess, Jack Clark, Christopher Berner, Sam McCandlish, Alec
  Radford, Ilya Sutskever, and Dario Amodei.
\newblock {Language Models are Few-Shot Learners}.
\newblock {\em arXiv preprint arXiv:2005.14165}, 2020.

\bibitem{caesar2018cvpr}
Holger Caesar, Jasper Uijlings, and Vittorio Ferrari.
\newblock {COCO-Stuff: Thing and stuff classes in context}.
\newblock In {\em Computer vision and pattern recognition (CVPR), 2018 IEEE
  conference on}. IEEE, 2018.

\bibitem{chen2017deeplab}
Liang-Chieh Chen, G. Papandreou, I. Kokkinos, Kevin Murphy, and A. Yuille.
\newblock {DeepLab: Semantic Image Segmentation with Deep Convolutional Nets,
  Atrous Convolution, and Fully Connected CRFs}.
\newblock {\em IEEE Transactions on Pattern Analysis and Machine Intelligence},
  2018.

\bibitem{chen2020generative}
Mark Chen, Alec Radford, Rewon Child, Jeff Wu, Heewoo Jun, Prafulla Dhariwal,
  David Luan, and Ilya Sutskever.
\newblock Generative pretraining from pixels.
\newblock 2020.

\bibitem{ChenK17}
Qifeng Chen and Vladlen Koltun.
\newblock Photographic image synthesis with cascaded refinement networks.
\newblock In {\em {IEEE} International Conference on Computer Vision, {ICCV}
  2017, Venice, Italy, October 22-29, 2017}, pages 1520--1529. {IEEE} Computer
  Society, 2017.

\bibitem{chen2018pixelsnail}
Xi Chen, Nikhil Mishra, Mostafa Rohaninejad, and Pieter Abbeel.
\newblock Pixelsnail: An improved autoregressive generative model.
\newblock In {\em {ICML}}, volume~80 of {\em Proceedings of Machine Learning
  Research}, pages 863--871. {PMLR}, 2018.

\bibitem{child2020vdvae}
Rewon Child.
\newblock Very deep vaes generalize autoregressive models and can outperform
  them on images.
\newblock {\em CoRR}, abs/2011.10650, 2020.

\bibitem{child2019generating}
Rewon Child, Scott Gray, Alec Radford, and Ilya Sutskever.
\newblock Generating long sequences with sparse transformers, 2019.

\bibitem{dai2019diagnosing}
Bin Dai and David~P. Wipf.
\newblock Diagnosing and enhancing {VAE} models.
\newblock In {\em 7th International Conference on Learning Representations,
  {ICLR}}, 2019.

\bibitem{deng2009imagenet}
Jia Deng, Wei Dong, Richard Socher, Li-Jia Li, Kai Li, and Li Fei-Fei.
\newblock Imagenet: {A} large-scale hierarchical image database.
\newblock In {\em 2009 {IEEE} Computer Society Conference on Computer Vision
  and Pattern Recognition {CVPR}}, 2009.

\bibitem{dhariwal2021diffusion}
Prafulla Dhariwal and Alex Nichol.
\newblock Diffusion models beat gans on image synthesis, 2021.

\bibitem{anonymous2021an}
Alexey Dosovitskiy, Lucas Beyer, Alexander Kolesnikov, Dirk Weissenborn,
  Xiaohua Zhai, Thomas Unterthiner, Mostafa Dehghani, Matthias Minderer, Georg
  Heigold, Sylvain Gelly, et~al.
\newblock An image is worth 16x16 words: Transformers for image recognition at
  scale.
\newblock 2020.

\bibitem{dosovitskiy2016generating}
Alexey Dosovitskiy and Thomas Brox.
\newblock {Generating Images with Perceptual Similarity Metrics based on Deep
  Networks}.
\newblock In {\em Advances in Neural Information Processing Systems 29: Annual
  Conference on Neural Information Processing Systems, NeurIPS}, 2016.

\bibitem{esser2020invertible}
Patrick Esser, Robin Rombach, and Bj{\"{o}}rn Ommer.
\newblock {A Disentangling Invertible Interpretation Network for Explaining
  Latent Representations}.
\newblock In {\em 2020 {IEEE/CVF} Conference on Computer Vision and Pattern
  Recognition, {CVPR}}, 2020.

\bibitem{msp}
Jeffrey~De Fauw, Sander Dieleman, and Karen Simonyan.
\newblock Hierarchical autoregressive image models with auxiliary decoders.
\newblock {\em CoRR}, abs/1903.04933, 2019.

\bibitem{gan}
Ian~J. Goodfellow, Jean Pouget{-}Abadie, Mehdi Mirza, Bing Xu, David
  Warde{-}Farley, Sherjil Ozair, Aaron~C. Courville, and Yoshua Bengio.
\newblock {Generative Adversarial Nets}.
\newblock In {\em Advances in Neural Information Processing Systems 27: Annual
  Conference on Neural Information Processing Systems, NeurIPS}, 2014.

\bibitem{han2020notsobiggan}
Seungwook Han, Akash Srivastava, Cole~L. Hurwitz, Prasanna Sattigeri, and
  David~D. Cox.
\newblock not-so-biggan: Generating high-fidelity images on a small compute
  budget.
\newblock {\em CoRR}, abs/2009.04433, 2020.

\bibitem{DBLP:journals/corr/HeZRS15}
Kaiming He, Xiangyu Zhang, Shaoqing Ren, and Jian Sun.
\newblock Deep residual learning for image recognition.
\newblock {\em CoRR}, abs/1512.03385, 2015.

\bibitem{heljakka2018pioneer}
Ari Heljakka, Arno Solin, and Juho Kannala.
\newblock Pioneer networks: Progressively growing generative autoencoder.
\newblock In C.~V. Jawahar, Hongdong Li, Greg Mori, and Konrad Schindler,
  editors, {\em Computer Vision - {ACCV} 2018 - 14th Asian Conference on
  Computer Vision, Perth, Australia, December 2-6, 2018, Revised Selected
  Papers, Part {I}}, 2018.

\bibitem{heljakka2020towards}
Ari Heljakka, Arno Solin, and Juho Kannala.
\newblock Towards photographic image manipulation with balanced growing of
  generative autoencoders.
\newblock In {\em {IEEE} Winter Conference on Applications of Computer Vision,
  WACV 2020, Snowmass Village, CO, USA, March 1-5, 2020}, pages 3109--3118.
  {IEEE}, 2020.

\bibitem{ho2020denoising}
Jonathan Ho, Ajay Jain, and Pieter Abbeel.
\newblock Denoising diffusion probabilistic models, 2020.

\bibitem{ho2019axial}
Jonathan Ho, Nal Kalchbrenner, Dirk Weissenborn, and Tim Salimans.
\newblock Axial attention in multidimensional transformers.
\newblock {\em CoRR}, abs/1912.12180, 2019.

\bibitem{DBLP:conf/iclr/HoltzmanBDFC20}
Ari Holtzman, Jan Buys, Li Du, Maxwell Forbes, and Yejin Choi.
\newblock The curious case of neural text degeneration.
\newblock In {\em {ICLR}}. OpenReview.net, 2020.

\bibitem{isola2017image}
Phillip Isola, Jun{-}Yan Zhu, Tinghui Zhou, and Alexei~A. Efros.
\newblock {Image-to-Image Translation with Conditional Adversarial Networks}.
\newblock In {\em 2017 {IEEE} Conference on Computer Vision and Pattern
  Recognition, {CVPR}}, 2017.

\bibitem{jang2016categorical}
Eric Jang, Shixiang Gu, and Ben Poole.
\newblock {Categorical reparameterization with gumbel-softmax}.
\newblock {\em arXiv preprint arXiv:1611.01144}, 2016.

\bibitem{johnson2016perceptual}
Justin Johnson, Alexandre Alahi, and Li Fei{-}Fei.
\newblock Perceptual losses for real-time style transfer and super-resolution.
\newblock In {\em {ECCV} {(2)}}, volume 9906 of {\em Lecture Notes in Computer
  Science}, pages 694--711. Springer, 2016.

\bibitem{karras2018progressive}
Tero Karras, Timo Aila, Samuli Laine, and Jaakko Lehtinen.
\newblock Progressive growing of gans for improved quality, stability, and
  variation.
\newblock {\em CoRR}, abs/1710.10196, 2017.

\bibitem{karras2020training}
Tero Karras, Miika Aittala, Janne Hellsten, Samuli Laine, Jaakko Lehtinen, and
  Timo Aila.
\newblock Training generative adversarial networks with limited data.
\newblock In Hugo Larochelle, Marc'Aurelio Ranzato, Raia Hadsell,
  Maria{-}Florina Balcan, and Hsuan{-}Tien Lin, editors, {\em Advances in
  Neural Information Processing Systems 33: Annual Conference on Neural
  Information Processing Systems 2020, NeurIPS 2020, December 6-12, 2020,
  virtual}, 2020.

\bibitem{karras2019stylebased}
Tero Karras, Samuli Laine, and Timo Aila.
\newblock A style-based generator architecture for generative adversarial
  networks.
\newblock In {\em {IEEE} Conference on Computer Vision and Pattern Recognition,
  (CVPR) 2019, Long Beach, CA, USA, June 16-20, 2019}, pages 4401--4410.
  Computer Vision Foundation / {IEEE}, 2019.

\bibitem{karras2020analyzing}
Tero Karras, Samuli Laine, Miika Aittala, Janne Hellsten, Jaakko Lehtinen, and
  Timo Aila.
\newblock Analyzing and improving the image quality of stylegan.
\newblock In {\em 2020 {IEEE/CVF} Conference on Computer Vision and Pattern
  Recognition, {CVPR} 2020, Seattle, WA, USA, June 13-19, 2020}, pages
  8107--8116. {IEEE}, 2020.

\bibitem{katiyar2021improving}
Prateek Katiyar and Anna Khoreva.
\newblock Improving augmentation and evaluation schemes for semantic image
  synthesis, 2021.

\bibitem{kim2017structured}
Yoon Kim, Carl Denton, Luong Hoang, and Alexander~M. Rush.
\newblock Structured attention networks, 2017.

\bibitem{kingma2018glow}
Diederik~P. Kingma and Prafulla Dhariwal.
\newblock {Glow: Generative Flow with Invertible 1x1 Convolutions}.
\newblock In {\em Advances in Neural Information Processing Systems 31: Annual
  Conference on Neural Information Processing Systems 2018, NeurIPS}, 2018.

\bibitem{vae}
Diederik~P. Kingma and Max Welling.
\newblock {Auto-Encoding Variational Bayes}.
\newblock In {\em 2nd International Conference on Learning Representations,
  {ICLR}}, 2014.

\bibitem{lamb2016discriminative}
Alex Lamb, Vincent Dumoulin, and Aaron~C. Courville.
\newblock Discriminative regularization for generative models.
\newblock {\em CoRR}, abs/1602.03220, 2016.

\bibitem{larsen2015autoencoding}
Anders Boesen~Lindbo Larsen, Søren~Kaae Sønderby, Hugo Larochelle, and Ole
  Winther.
\newblock {Autoencoding beyond pixels using a learned similarity metric}, 2015.

\bibitem{li2019neural}
Naihan Li, Shujie Liu, Yanqing Liu, Sheng Zhao, and Ming Liu.
\newblock Neural speech synthesis with transformer network.
\newblock In {\em {AAAI}}, pages 6706--6713. {AAAI} Press, 2019.

\bibitem{lin2019cocogan}
Chieh~Hubert Lin, Chia{-}Che Chang, Yu{-}Sheng Chen, Da{-}Cheng Juan, Wei Wei,
  and Hwann{-}Tzong Chen.
\newblock {COCO-GAN:} generation by parts via conditional coordinating.
\newblock In {\em {ICCV}}, pages 4511--4520. {IEEE}, 2019.

\bibitem{liu2019acceleration}
Jinlin Liu, Yuan Yao, and Jianqiang Ren.
\newblock An acceleration framework for high resolution image synthesis.
\newblock {\em CoRR}, abs/1909.03611, 2019.

\bibitem{liu2018generating}
Peter~J. Liu, Mohammad Saleh, Etienne Pot, Ben Goodrich, Ryan Sepassi, Lukasz
  Kaiser, and Noam Shazeer.
\newblock Generating wikipedia by summarizing long sequences.
\newblock In {\em {ICLR} (Poster)}. OpenReview.net, 2018.

\bibitem{liuLQWTcvpr16DeepFashion}
Ziwei Liu, Ping Luo, Shi Qiu, Xiaogang Wang, and Xiaoou Tang.
\newblock Deepfashion: Powering robust clothes recognition and retrieval with
  rich annotations.
\newblock In {\em Proceedings of IEEE Conference on Computer Vision and Pattern
  Recognition (CVPR)}, June 2016.

\bibitem{MenickK19}
Jacob Menick and Nal Kalchbrenner.
\newblock Generating high fidelity images with subscale pixel networks and
  multidimensional upscaling.
\newblock In {\em 7th International Conference on Learning Representations,
  {ICLR} 2019, New Orleans, LA, USA, May 6-9, 2019}. OpenReview.net, 2019.

\bibitem{mentzer2020highfidelity}
Fabian Mentzer, George Toderici, Michael Tschannen, and Eirikur Agustsson.
\newblock High-fidelity generative image compression, 2020.

\bibitem{nash2021generating}
Charlie Nash, Jacob Menick, Sander Dieleman, and Peter~W. Battaglia.
\newblock Generating images with sparse representations, 2021.

\bibitem{nichol2021improved}
Alex Nichol and Prafulla Dhariwal.
\newblock Improved denoising diffusion probabilistic models, 2021.

\bibitem{obukhov2020torchfidelity}
Anton Obukhov, Maximilian Seitzer, Po-Wei Wu, Semen Zhydenko, Jonathan Kyl, and
  Elvis Yu-Jing Lin.
\newblock High-fidelity performance metrics for generative models in pytorch,
  2020.
\newblock Version: 0.3.0, DOI: 10.5281/zenodo.4957738.

\bibitem{ommer2007learning}
B. {Ommer} and J.~M. {Buhmann}.
\newblock Learning the compositional nature of visual objects.
\newblock In {\em 2007 IEEE Conference on Computer Vision and Pattern
  Recognition}, pages 1--8, 2007.

\bibitem{parikh2016decomposable}
Ankur~P. Parikh, Oscar Täckström, Dipanjan Das, and Jakob Uszkoreit.
\newblock A decomposable attention model for natural language inference, 2016.

\bibitem{park2019semantic}
Taesung Park, Ming-Yu Liu, Ting-Chun Wang, and Jun-Yan Zhu.
\newblock {Semantic Image Synthesis with Spatially-Adaptive Normalization}.
\newblock In {\em Proceedings of the IEEE Conference on Computer Vision and
  Pattern Recognition, CVPR}, 2019.

\bibitem{parmar2020dual}
Gaurav Parmar, Dacheng Li, Kwonjoon Lee, and Zhuowen Tu.
\newblock Dual contradistinctive generative autoencoder, 2020.

\bibitem{parmar2018image}
Niki Parmar, Ashish Vaswani, Jakob Uszkoreit, Lukasz Kaiser, Noam Shazeer,
  Alexander Ku, and Dustin Tran.
\newblock Image transformer.
\newblock In {\em {ICML}}, volume~80 of {\em Proceedings of Machine Learning
  Research}, pages 4052--4061. {PMLR}, 2018.

\bibitem{PidhorskyiAD20}
Stanislav Pidhorskyi, Donald~A. Adjeroh, and Gianfranco Doretto.
\newblock Adversarial latent autoencoders.
\newblock In {\em 2020 {IEEE/CVF} Conference on Computer Vision and Pattern
  Recognition, {CVPR} 2020, Seattle, WA, USA, June 13-19, 2020}, pages
  14092--14101. {IEEE}, 2020.

\bibitem{Radford2018ImprovingLU}
A. Radford.
\newblock Improving language understanding by generative pre-training.
\newblock 2018.

\bibitem{Radford2019LanguageMA}
A. Radford, Jeffrey Wu, R. Child, David Luan, Dario Amodei, and Ilya Sutskever.
\newblock Language models are unsupervised multitask learners.
\newblock 2019.

\bibitem{ramesh2021zeroshot}
Aditya Ramesh, Mikhail Pavlov, Gabriel Goh, Scott Gray, Chelsea Voss, Alec
  Radford, Mark Chen, and Ilya Sutskever.
\newblock Zero-shot text-to-image generation, 2021.

\bibitem{Ranftl2020}
Ren\'{e} Ranftl, Katrin Lasinger, David Hafner, Konrad Schindler, and Vladlen
  Koltun.
\newblock Towards robust monocular depth estimation: Mixing datasets for
  zero-shot cross-dataset transfer.
\newblock {\em IEEE Transactions on Pattern Analysis and Machine Intelligence
  (TPAMI)}, 2020.

\bibitem{razavi2019generating}
Ali Razavi, Aaron van~den Oord, and Oriol Vinyals.
\newblock Generating diverse high-fidelity images with vq-vae-2, 2019.

\bibitem{vae2}
Danilo~Jimenez Rezende, Shakir Mohamed, and Daan Wierstra.
\newblock {Stochastic backpropagation and approximate inference in deep
  generative models}.
\newblock In {\em Proceedings of the 31st International Conference on
  International Conference on Machine Learning, ICML}, 2014.

\bibitem{rombach2020making}
Robin Rombach, Patrick Esser, and Bj{\"{o}}rn Ommer.
\newblock Making sense of cnns: Interpreting deep representations and their
  invariances with inns.
\newblock In Andrea Vedaldi, Horst Bischof, Thomas Brox, and Jan{-}Michael
  Frahm, editors, {\em Computer Vision - {ECCV} 2020 - 16th European
  Conference, Glasgow, UK, August 23-28, 2020, Proceedings, Part {XVII}},
  volume 12362 of {\em Lecture Notes in Computer Science}, pages 647--664.
  Springer, 2020.

\bibitem{rombach2020network}
Robin Rombach, Patrick Esser, and Bjorn Ommer.
\newblock Network-to-network translation with conditional invertible neural
  networks.
\newblock In H. Larochelle, M. Ranzato, R. Hadsell, M.~F. Balcan, and H. Lin,
  editors, {\em Advances in Neural Information Processing Systems}, volume~33,
  pages 2784--2797. Curran Associates, Inc., 2020.

\bibitem{santurkar2019computer}
Shibani Santurkar, Dimitris Tsipras, Brandon Tran, Andrew Ilyas, Logan
  Engstrom, and Aleksander Madry.
\newblock Computer vision with a single (robust) classifier.
\newblock In {\em ArXiv preprint arXiv:1906.09453}, 2019.

\bibitem{SchonfeldSK20}
Edgar Sch{\"{o}}nfeld, Bernt Schiele, and Anna Khoreva.
\newblock A u-net based discriminator for generative adversarial networks.
\newblock In {\em 2020 {IEEE/CVF} Conference on Computer Vision and Pattern
  Recognition, {CVPR} 2020, Seattle, WA, USA, June 13-19, 2020}, pages
  8204--8213. {IEEE}, 2020.

\bibitem{seonghyeon2020}
Kim Seonghyeon.
\newblock Implementation of generating diverse high-fidelity images with
  vq-vae-2 in pytorch, 2020.

\bibitem{Siarohin_2019_NeurIPS}
Aliaksandr Siarohin, Stéphane Lathuilière, Sergey Tulyakov, Elisa Ricci, and
  Nicu Sebe.
\newblock First order motion model for image animation.
\newblock In {\em Conference on Neural Information Processing Systems
  (NeurIPS)}, December 2019.

\bibitem{VahdatK20}
Arash Vahdat and Jan Kautz.
\newblock {NVAE:} {A} deep hierarchical variational autoencoder.
\newblock In Hugo Larochelle, Marc'Aurelio Ranzato, Raia Hadsell,
  Maria{-}Florina Balcan, and Hsuan{-}Tien Lin, editors, {\em Advances in
  Neural Information Processing Systems 33: Annual Conference on Neural
  Information Processing Systems 2020, NeurIPS 2020, December 6-12, 2020,
  virtual}, 2020.

\bibitem{oord2016pixel}
A{\"{a}}ron van~den Oord, Nal Kalchbrenner, and Koray Kavukcuoglu.
\newblock Pixel recurrent neural networks.
\newblock In {\em {ICML}}, volume~48 of {\em {JMLR} Workshop and Conference
  Proceedings}, pages 1747--1756. JMLR.org, 2016.

\bibitem{oord2016conditional}
Aaron van~den Oord, Nal Kalchbrenner, Oriol Vinyals, Lasse Espeholt, Alex
  Graves, and Koray Kavukcuoglu.
\newblock Conditional image generation with pixelcnn decoders, 2016.

\bibitem{oord2018neural}
Aaron van~den Oord, Oriol Vinyals, and Koray Kavukcuoglu.
\newblock Neural discrete representation learning, 2018.

\bibitem{hugo_van_kemenade_2021_4659051}
Hugo van Kemenade, wiredfool, Andrew Murray, Alex Clark, Alexander Karpinsky,
  Ondrej Baranovič, Christoph Gohlke, Jon Dufresne, Brian Crowell, David
  Schmidt, Konstantin Kopachev, Alastair Houghton, Sandro Mani, Steve Landey,
  vashek, Josh Ware, Jason Douglas, David Caro, Uriel Martinez, Steve Kossouho,
  Riley Lahd, Stanislau T., Antony Lee, Eric~W. Brown, Oliver Tonnhofer,
  Mickael Bonfill, Peter Rowlands, Fahad Al-Saidi, German Novikov, and Michał
  Górny.
\newblock python-pillow/pillow: 8.2.0, Apr. 2021.

\bibitem{vaswani2017attention}
Ashish Vaswani, Noam Shazeer, Niki Parmar, Jakob Uszkoreit, Llion Jones,
  Aidan~N. Gomez, Lukasz Kaiser, and Illia Polosukhin.
\newblock {Attention is All you Need}.
\newblock In {\em Advances in Neural Information Processing Systems 30: Annual
  Conference on Neural Information Processing Systems, NeurIPS}, 2017.

\bibitem{wang2018pix2pixHD}
Ting-Chun Wang, Ming-Yu Liu, Jun-Yan Zhu, Andrew Tao, Jan Kautz, and Bryan
  Catanzaro.
\newblock High-resolution image synthesis and semantic manipulation with
  conditional gans.
\newblock In {\em Proceedings of the IEEE Conference on Computer Vision and
  Pattern Recognition}, 2018.

\bibitem{weissenborn2020scaling}
Dirk Weissenborn, Oscar T{\"{a}}ckstr{\"{o}}m, and Jakob Uszkoreit.
\newblock Scaling autoregressive video models.
\newblock In {\em {ICLR}}. OpenReview.net, 2020.

\bibitem{xiao2021vaebm}
Zhisheng Xiao, Karsten Kreis, Jan Kautz, and Arash Vahdat.
\newblock Vaebm: A symbiosis between variational autoencoders and energy-based
  models, 2021.

\bibitem{xiao2019generative}
Zhisheng Xiao, Qing Yan, Yi{-}an Chen, and Yali Amit.
\newblock Generative latent flow: {A} framework for non-adversarial image
  generation.
\newblock {\em CoRR}, abs/1905.10485, 2019.

\bibitem{yu15lsun}
Fisher Yu, Yinda Zhang, Shuran Song, Ari Seff, and Jianxiong Xiao.
\newblock Lsun: Construction of a large-scale image dataset using deep learning
  with humans in the loop.
\newblock {\em arXiv preprint arXiv:1506.03365}, 2015.

\bibitem{zhang2020cross}
Pan Zhang, Bo Zhang, Dong Chen, Lu Yuan, and Fang Wen.
\newblock {Cross-Domain Correspondence Learning for Exemplar-Based Image
  Translation}.
\newblock In {\em 2020 {IEEE/CVF} Conference on Computer Vision and Pattern
  Recognition, {CVPR}}, 2020.

\bibitem{zhang2018perceptual}
Richard Zhang, Phillip Isola, Alexei~A Efros, Eli Shechtman, and Oliver Wang.
\newblock {The Unreasonable Effectiveness of Deep Features as a Perceptual
  Metric}.
\newblock In {\em CVPR}, 2018.

\bibitem{zhang2018unreasonable}
Richard Zhang, Phillip Isola, Alexei~A Efros, Eli Shechtman, and Oliver Wang.
\newblock {The unreasonable effectiveness of deep features as a perceptual
  metric}.
\newblock In {\em Proceedings of the IEEE Conference on Computer Vision and
  Pattern Recognition, CVPR}, 2018.

\bibitem{zhou2016semantic}
Bolei Zhou, Hang Zhao, Xavier Puig, Sanja Fidler, Adela Barriuso, and Antonio
  Torralba.
\newblock Semantic understanding of scenes through the ade20k dataset.
\newblock {\em arXiv preprint arXiv:1608.05442}, 2016.

\bibitem{zhou2017view}
Tinghui Zhou, Shubham Tulsiani, Weilun Sun, Jitendra Malik, and Alexei~A.
  Efros.
\newblock View synthesis by appearance flow, 2017.

\bibitem{zhu2019sean}
Peihao Zhu, Rameen Abdal, Yipeng Qin, and Peter Wonka.
\newblock Sean: Image synthesis with semantic region-adaptive normalization,
  2019.

\end{thebibliography}
